\documentclass{article}

\usepackage[utf8]{inputenc} 
\usepackage[T1]{fontenc}    
\usepackage{hyperref}       
\usepackage{url}            
\usepackage{booktabs}       
\usepackage{amsfonts}       
\usepackage{nicefrac}       
\usepackage{microtype}      
\usepackage{enumitem}
\usepackage{xcolor}

\usepackage[accepted]{icml2021}

\usepackage{wrapfig}

\usepackage{definitions}

\usepackage{cleveref}

\begin{document}

\twocolumn[
\icmltitle{Parameter-free Gradient Temporal Difference Learning}

\begin{icmlauthorlist}
  \icmlauthor{Andrew Jacobsen}{ualberta}
  \icmlauthor{Alan Chan}{montreal,mila}
\end{icmlauthorlist}

\icmlaffiliation{ualberta}{Department of Computing Science, University of Alberta, Edmonton, Canada}
\icmlaffiliation{mila}{Mila, Montréal, Canada}
\icmlaffiliation{montreal}{Département d'informatique et de recherche opérationnelle, Université of Montréal, Montréal, Canada}

\icmlcorrespondingauthor{Andrew Jacobsen}{ajjacobs@ualberta.ca}

\icmlkeywords{Reinforcement Learning, Parameter-free Learning, Coin betting, Coin-betting, Gradient Temporal Difference Learning, Saddle-point Optimization}

\vskip 0.3in
]
\printAffiliationsAndNotice{}

\begin{abstract}
  Reinforcement learning lies at the intersection of several challenges.
  Many applications of interest involve extremely large state spaces, requiring \textit{function approximation} to enable tractable computation.
  In addition, the learner has only a single stream of experience with which to evaluate a large number of possible courses of action, necessitating
  algorithms which can learn \textit{off-policy}.
  However, the combination of off-policy learning with function approximation
  leads to divergence of temporal difference methods.
  Recent work into gradient-based temporal difference methods has promised a path to stability, but at the cost of expensive hyperparameter tuning.
  In parallel, progress in online learning has provided parameter-free methods that achieve minimax optimal guarantees up to logarithmic terms, but their application in reinforcement learning has yet to be explored.
  In this work, we combine these two lines of attack, deriving parameter-free, gradient-based temporal difference algorithms. Our algorithms run in linear time and achieve high-probability convergence guarantees matching those of GTD2 up to $\log$ factors.
  Our experiments demonstrate that our methods maintain high prediction performance relative to fully-tuned baselines, with no tuning whatsoever.
\end{abstract}

\section{Introduction}
\vspace{-1ex}

A central problem in reinforcement learning (RL) is policy evaluation: given a course of action --- otherwise known as a \textit{policy} --- what is its expected long-term outcome?
Policy evaluation is an essential step in policy iteration schemes, facilitates
model-based planning \citep{grimm2020value},
and in recent years has even found applications in representation learning \citep{veeriah2019discovery,schlegel2021general}.

Temporal difference (TD) methods \cite{sutton1988learning} comprise the most popular class of policy-evaluation algorithms, leveraging
approximate dynamic programming approaches to perform credit assignment incrementally.
Nevertheless, a notable deficiency is that they require a step-size, which determines the pace of adapting value function estimates to the current data point. A step-size that is too high may impede convergence and stability, while a step-size that is too low can stall progress. Some TD algorithms that enjoy other desirable properties, such as Gradient-TD (GTD) methods \cite{liu2018proximal} that can be shown to converge where other TD methods do not \cite{baird1995residual}, but can be incredibly sensitive to the step-size. Generally, selection of the step-size and other parameters --- known as \textit{tuning} --- requires massive amounts of computation time, to the extent that this tuning has become an impediment to reliable scientific results \cite{henderson2018where,henderson2019deep}.

Various approaches have been proposed to alleviate these difficulties.
Step-size schedules leading to optimal convergence rates
and robust guarantees are well-known in the stochastic optimization literature
\citep{nemirovski2009robust,ghadimi2013stochastic},
but they often depend on problem-dependent quantities that are unavailable to the practitioner.
Meta-descent methods such as Stochatic Meta-descent \citep{schraudolph1999local}  and TIDBD \cite{kearney2018tidbd} instead \textit{learn}
a step-size according to some meta-objective, thereby obviating the need to tune one. Yet
these methods inevitably introduce \textit{new} parameters to be tuned, such as
a meta-stepsize, decay parameter parameter, etc, and can be quite sensitive to these new parameters in practice \citep{jacobsen2019meta}. Quasi second-order methods adopted from the deep learning community --- such as the
well-known Adam algorithm \citep{kingma2014adam} --- can be effective in practice but
have no theoretical guarantees when applied to TD methods. Further, while these quasi second-order methods
are often less sensitive to their parameters, they never fully manage to remove the need for parameter
tuning completely.

To address these deficiencies, we adapt recent work on parameter-free online learning to policy evaluation.
We leverage a series of black-box reductions to reduce policy evaluation to a problem that can be solved
using a powerful class of online linear optimization (OLO) algorithm based on betting strategies \citep{orabona2016coin,orabona2019modern,cutkosky2018black}.
The key characteristic of these approaches is that they require no parameter tuning, yet guarantee near-optimal regret bounds. We construct our algorithms in such a way that they
avoid requiring any \textit{a priori} information about problem-dependent constants, while still making high-probability guarantees and achieving similar performance to
fully-tuned baselines.

\section{Preliminaries}

Reinforcement learning (RL) is a framework in which a decision-maker (the \textit{agent}) learns via repeated trial-and-error
interaction with its \textit{environment}. The environment is formalized as a \textit{Markov decision process} $(\cS,\cA, P, R)$,
where $\cS$ and $\cA$ are sets of states and actions, $P(s'|s,a)$ is a transition kernel specifying the probability of transitioning to
state $s'\in\cS$ when taking action $a\in\cA$ in state $s\in\cS$, and $R:\cS\times\cA\rightarrow\R$ is a reward function mapping state-action
pairs to rewards. At each discrete time-step $t\in\N$, the agent observes the current state $S_{t}\in\cS$ and selects an
action $A_{t}$ according to policy $\pi:\cS\times\cA\rightarrow[0,1]$. The \textit{value function} $v^{\pi}$ is the
expected discounted reward when following policy $\pi$ from state $s$:
\[
  v^{\pi}(s) = \EE{\sum_{k=0}^{\infty}\gamma^{k}R_{t+k}|S_{0}=s},
\]
where $\gamma\in[0,1)$ is a constant discount factor and $R_{t}$ denotes $R(S_{t},A_{t})$.
The process of estimating $v^{\pi}$ is referred to as \textit{policy evaluation}.

When $\abs{\cS}$ is large, it is common to estimate $v^{\pi}$ using parametric
function approximation. A simple yet often effective strategy is to approximate $v^{\pi}$
with a linear approximation. We assume access to a \textit{feature map} $\phi:\cS\rightarrow\R^{d}$
which maps each state to a vector of features, and approximate $\hat v^{\pi}(s)=\theta^{\top}\phi(s)\approx v^{\pi}(s)$.
In vector form, we write $V^{\pi}\in\R^{\abs{S}}$ and $\hat V^{\pi}=\Phi\theta$, where $\Phi\in\R^{\abs{\cS}\times d}$ is the matrix with rows $\phi(s)$.

In practice, the learner may have several candidate policies to choose between.
Rather than evaluating each of them separately, it is often desirable to
evaluate them simultaneously, in parallel, while following some sufficiently exploratory
behavior policy $\pi_{b}$.
This is \textit{off-policy} evaluation. In particular,
given a target policy $\pi$, we would like to estimate
estimate $V^{\pi}$ using a dataset of $T\in\N$ transitions
$\cD\defeq\set{(s_{t}, a_{t}, r_{t}, s_{t}')}_{t=1}^{T}$, where
$s_{t}\sim \xi(\cdot)$ for some state distribution $\xi$, $a_{t}\sim\pi_{b}(\cdot|s_{t})$, $r_{t}=R(s_{t},a_{t})$, and
$s_{t}^{\prime}\sim P(\cdot|s_{t},a_{t})$. For a given transition $(s_{t},a_{t},r_{t},s_{t}')\in\cD$ we denote
the importance sampling ratio $\rho_{t}\defeq\frac{\pi(a_{t}|s_{t})}{\pi_{b}(a_{t}|s_{t})}$,
the shorthands $\phi_{t}\defeq\phi(s_{t})$ and $\phi_{t}'\defeq\phi(s_{t}')$,
the TD error $\delta_{t}\defeq r_{t}+\gamma\theta^{\top}\phi_{t}' - \theta^{\top}\phi_{t}$, and $\Xi \defeq \text{diag}(\xi)\in\R^{\abs{\cS}\times\abs{\cS}}$ the diagonal matrix induced by the state distribution.
Finally, in what follows we will make use of the following definitions
\begin{align}
  \hat A_{t}\defeq\rho_{t}\phi_{t}(\phi_{t}-\gamma\phi_{t}')^{\top},\ \   \hat b_{t}\defeq\rho_{t}r_{t}\phi_{t},\ \  \hat C_{t}\defeq\phi_{t}\phi_{t}^{\top} \label{eq:abc}
\end{align}
and their expectations $A\defeq\E[\hat A_{t}]$, $b\defeq\E[\hat b_{t}]$ $C\defeq\E[\hat C_{t}]$.

\textbf{Gradient-based Temporal Difference Learning.}
Under both function approximation and off-policy sampling, classic
TD algorithms can fail to converge for any scalar step-sizes \cite{baird1995residual}. Specifically, in the off-policy setting, $\xi$ is generally neither the stationary distribution of $\pi$ nor the discounted future state visitation distribution. In this case, $A$ can fail to be positive-definite, which could lead to non-convergence of the resulting dynamical system.

This deficiency has motivated attempts to reformulate the typical TD algorithms as
stochastic gradient methods. These gradient-based TD (GTD) methods seek to minimize
the \textit{Norm of the Expected Update} (NEU), or the \textit{Mean-square Projected Bellman Error} (MSPBE), either of which
can be described by the objective function
\begin{equation}
  J(\theta) =  \norm{\Phi^{\top}\Xi(\cB^{\pi}\hat V^{\pi}-\hat V^{\pi})}^{2}_{M^{\inv}}=\norm{b - A\theta}_{M^{\inv}},\label{eq:objective}
\end{equation}
where  $\cB^{\pi}$ is the Bellman operator
\citep{liu2018proximal}.
The NEU is captured by setting $M=I$, and the MSPBE via $M=\E[\phi_{t}\phi_{t}^{\top}]$.

In this paper, we leverage the saddle-point (SP) formulation in \citet{liu2018proximal}.
Minimization of $J(\theta)$ can be equivalently formulated as the following
SP problem:
\begin{align}
  \min_{\theta\in\Theta}\max_{y\in\cY} \set{L(\theta, y)\defeq \inner{b-A\theta,y}-\half\norm{y}^{2}_{M}},\label{eq:sp}
\end{align}
where $\Theta\subset\R^{d}$ and $\cY\subset\R^{d}$ are assumed convex and compact.
The quality of a candidate solution $(\theta,y)\in\Theta\times\cY$ is measured in terms of the \textit{duality gap}:
\begin{equation}
  \max_{y^{*}\in\cY}L(\theta,y^{*})-\min_{\theta^{*}\in\Theta}L(\theta^{*},y).\label{eq:duality-gap}
\end{equation}
It can be shown  that this quantity is non-negative, and any
point $(\theta,y)$ for which $\max_{y^{*}\in\cY}L(\theta,y^{*})-\min_{\theta^{*}\in\Theta}L(\theta^{*},y)=0$
is a saddle-point of the SP problem \ref{eq:sp} (see Appendix \ref{app:sp-to-olo}). Further, it is known that if $(\theta^{*},y^{*})$ is a
saddle-point of Equation \ref{eq:sp}, then $\theta^{*}$ is a minimizer of the MSPBE \cite{liu2018proximal}.

\textbf{Notations.}
In what follows, we assume for simplicity\footnote{We note that our results can be easily extended to arbitrary Hilbert spaces, at the expense of additional space and notation.} that $\Theta$ and $\cY$ are equipped with norms $\norm{\cdot}=\norm{\cdot}_{2}$.
For a PSD matrix $M$ we denote $\norm{x}_{M}^{2}\defeq \inner{x,Mx}$.
The decision sets $\Theta$ and $\cY$ are assumed compact and convex, and
we denote $D\defeq\max\set{\max_{\theta\in\Theta}\norm{\theta},\max_{y\in\cY}\norm{y}}$ and $D_{\infty}\defeq\max\set{\max_{\theta\in\Theta}\norm{\theta}_{\infty}, \max_{y\in\cY}\norm{y}_{\infty}}$.
Given a closed convex set $W$, we define the projection $\Pi_{W}(w) = \argmin_{\hat w\in W}\norm{\hat w-w}$.
The notation $O(\cdot)$ hides constant terms and $\widetilde{O}(\cdot)$ additionally hides log terms.
Throughout this work, we will denote $g_{t}^{\theta}=\E[\hat g_{t}^{\theta}|\theta_{t},y_{t}]\in \partial_{\theta}L(\theta_{t},y_{t})$ and
$g_{t}^{y}=\E[\hat g_{t}^{y}|\theta_{t},y_{t}]\in-\partial_{y}L(\theta_{t},y_{t})$, with
$\hat g_{t}^{\theta}$ and $\hat g_{t}^{y}$ denoting stochastic subgradients of \Cref{eq:sp}.
Concretely,
\begin{align}
    &\ghat_{t}^{\theta}= -\hat A_{t}^{\top}y_{t},\qquad
      \ghat_{t}^{y}= -\hat b_{t} + \hat A_{t}\theta_{t}+\hat M_{t}y_{t}\label{eq:noisy-subgrads}.
\end{align}
It will often be convenient to work directly with the joint space $\Theta\times\cY$, which we endow with norms
$\norm{(\theta,y)}\defeq\sqrt{\norm{\theta}^{2}+\norm{y}^{2}}$ and $\norm{(\theta,y)}_{\infty}=\max\set{\norm{\theta}_{\infty},\norm{y}_{\infty}}$. When clear from context, we
will denote $g_{t}\defeq(g_{t}^{\theta},g_{t}^{y})$, $\ghat_{t}\defeq(\ghat_{t}^{\theta},\ghat_{t}^{y})$,
$z_{t}\defeq(\theta_{t},y_{t})$ and $z^{*}\defeq(\theta^{*},y^{*})$.

\begin{minipage}{\columnwidth}
  \textbf{Assumptions.} Throughout this work we make the following standard assumptions
  \begin{enumerate}[label=\textbf{(A\arabic*)}]
    \item  \textit{(feasibile sets)} The feasible sets $\Theta$ and $\cY$ are assumed
      to be compact and convex, and the the solution $(\theta^{*},y^{*})$
      of SP problem \ref{eq:sp} is assumed to belong to $\Theta\times\cY$. \label{asm:1}
    \item \textit{(non-singularity)} The covariance matrix $C=\EE{\phi_{t}\phi_{t}^{\top}}=\Phi^{\top}D\Phi$
      and $A=\EE{\rho_{t}\phi_{t}(\phi_{t}-\gamma\phi_{t}')^{\top}}$ are assumed
      to be non-singular.\label{asm:2}
    \item \textit{(boundedness)} The features $(\phi_{t},\phi_{t}')$ have uniformly
      bounded second moments. Furthermore, we assume the upperbounds
      $\max_{t}\norm{\phi_{t}}_{\infty}\le L$, $\rho_{t}\le\rho_{\max}$, and
      $R(s_{t}, a_{t})\le R_{\max}$ exist and are finite for all $t\in[T]$. \label{asm:3}
  \end{enumerate}
\end{minipage}

\newcommand{\sigmaPlacement}{_}
\newcommand{\sigmaTheta}{\sigma\sigmaPlacement{\theta}}
\newcommand{\sigmaY}{\sigma\sigmaPlacement{y}}
Together assumptions A2 and A3 imply the existence of values $G_{\theta},G_{y},\sigma_{\theta},\sigma_{y}\in\R$ and such that
for $g_{t},\ghat_{t}\in\Theta\times\cY$ with $\EE{\ghat_{t}|\theta_{t},y_{t}}=g_{t}$, the following bounds hold
\begin{align}
  \EE{\norm{\ghat_{t}}^{2}}&\le G^{2}\defeq G_{\theta}^{2}+G_{y}^{2}\\
  \EE{\norm{\ghat_{t}-g_{t}}^{2}}&\le \sigma^{2}\defeq \sigma_{\theta}^{2}+\sigma_{y}^{2}
\end{align}

\section{Parameter-free Gradient Temporal Difference Learning}

In this section, we derive parameter-free algorithms for off-policy policy evaluation with linear function approximation.
Our approach proceeds in three steps. In \Cref{section:sp-to-olo}, we first reduce our policy evaluation problem to an online learning problem, and show that the performance guarantees
of the GTD2 algorithm can be matched using any online learning algorithm exhibiting a particular kind of \textit{adaptivity} --- this will enable us to make robust guarantees without appealing to any hyperparameters of the algorithm in question.
\Cref{section:olo-to-pfolo} then discusses a class of online learning algorithm which avoids tuning step-sizes while maintaining near-optimal guarantees, as well as an algorithmic component that will make their use more practical in
our problem setting. Finally, in \Cref{section:pfolo-to-policy-eval} we present our proposed algorithm, PFGTD+.

\subsection{From Saddle-point Optimization to OLO}\label{section:sp-to-olo}

Our first step is to reduce our saddle-point problem (\Cref{eq:sp}) to an OLO problem.
Luckily, the structure of \Cref{eq:sp} makes this easy: at each time $t$ we can simply compute the stochastic subgradients (as given in \Cref{eq:noisy-subgrads})
and pass them along to an OLO algorithm $\cA$, as shown in \Cref{alg:sp-to-olo}. As the following theorem then shows, we need not make any assumptions about the internals of $\cA$ to obtain a high-probability bound on the duality gap --- we only require that $\cA$ exhibits a particular form of \textit{adaptivity} to the sequence of subgradients.
\begin{Algorithm}
  \small
  \STATE  \textbf{Input:} Online Learning algorithm $\cA$, dataset $\cD=\set{\phi_{t},a_{t},r_{t},\phi_{t}', \rho_{t}}_{t=1}^{T}$, policies $\pi$ and $\pi_{b}$
  \For{$t=1:T$}{
    \STATE  Get $\theta_{t}$ and $y_{t}$ from $\cA$
    \STATE Set $\hat A_{t}=\rho_{t}\phi_{t}\left(\phi_{t}-\gamma\phi_{t}'\right)^{\top}$,\quad
    $\hat b_{t}=\rho_{t}r_{t}\phi_{t}$,\quad $\hat M_{t}=\phi_{t}\phi_{t}^{\top}$
    \STATE Set $\hat g_{t}^{\theta}=-\hat A_{t}^{\top}y_{t}$ and $\hat g_{t}^{y}= \hat A_{t}\theta_{t}-\hat b_{t} +\hat M_{t}y_{t}$

    \STATE Send $\hat g_{t}^{\theta}$ and
    $\hat g_{t}^{y}$ to $\cA$ as the $t^{\text{th}}$ subgradients
  }
  \STATE \textbf{Return:} average iterates $\thetabar_{T}=\frac{1}{T}\sum_{t=1}^{T}\theta_{t}$, $\ybar_{T}=\frac{1}{T}\sum_{t=1}^{T}y_{t}$
  \caption{Saddle-point to Online Learning Reduction}
  \label{alg:sp-to-olo}
  \vspace{-2ex}
\end{Algorithm}
\begin{restatable}{theorem}{AdaptiveBound}
  Suppose that for any $u\in\Theta\times\cY$, online learning algorithm $\cA$ guarantees regret
  \vspace{-2ex}
  \[
    R_{T}^{\cA}(u) = \sum_{t=1}^{T}\inner{g_{t}, z_{t}- u} \le A_{T}(u) + B_{T}(u)\sqrt{\sum_{t=1}^{T}\norm{g_{t}}^{2}}
  \]
  where
  $A_{T}$ and $B_{T}$ are arbitrary deterministic, non-negative functions.
  Then under Assumptions A1-A3,
  Algorithm \ref{alg:sp-to-olo} guarantees
  \begin{align*}
    &\EE{\max_{y^{*}\in\cY}L(\thetabar_{T}, y^{*}) - \min_{\theta^{*}\in\Theta}L(\theta^{*},\ybar_{T})}\\
    &\qquad\le\frac{A_T(z^{*})}{T}+\frac{B_{T}(z^{*})D\brac{2\norm{A} + \lambda+\frac{\sigma+\norm{b}}{D}}}{\sqrt{T}},
  \end{align*}
  where $\lambda=\lambda_{\max{}}(M)$.
  Further, if the light-tailed condition $\EE{\Exp{\frac{\norm{\ghat_{t}}^{2}}{G^{2}}}}\le\Exp{1}$ holds,
  then for any $\delta>0$, Algorithm \ref{alg:sp-to-olo} guarantees with probability at least
  $1-\delta$
  \begin{align*}
    &\max_{y^{*}\in\cY}L(\thetabar_{T}, y^{*}) - \min_{\theta^{*}\in\Theta}L(\theta^{*},\ybar_{T})
      \le\\
    &\frac{A_T(z^{*})}{T}+\frac{\brac{1+\Log{\frac{2}{\delta}}}B_{T}^{\delta}(z^{*})D\brac{2\norm{A} + \lambda+\frac{\sigma+\norm{b}}{D}}}{\sqrt{T}},
  \end{align*}
  where $B_{T}^{\delta}(z^{*}) = 2\max\Big\{B_{T}(z^{*}),\linebreak[1] \frac{4D}{\sqrt{1+\Log{2/\delta}}}\Big\}$.
  \label{thm:sp-to-olo}
\end{restatable}

Proof of this theorem can be found in Appendix \ref{app:highprob}.
\Cref{thm:sp-to-olo} tells us that the sample complexity $O\brac{\frac{\norm{A}+\lambda_{\max}(M)+\sigma}{\sqrt{T}}}$ of GTD2 can be matched up to
the coefficients $A_{T}$ and $B_{T}$ by any algorithm which achieves the so-called
\textit{second-order regret bound} $R_{T}\le O\brac{\sqrt{\sum_{t=1}^{T}\norm{g_{t}}^{2}}}$, and
that these guarantees will hold with high probability.
Notably, the result \textit{makes no appeal to step-sizes}. In fact, the theorem assumes nothing at all about the
internals of $\cA$, constrasting the high-probability bounds derived in \citet{liu2018proximal} which require a particular choice of step-size. This
would be problematic for our purposes, as our algorithms will not have a
step-size to tune in the first place!
Finally, we note that attaining high-probability bounds with parameter-free
algorithms is generally an open problem \citep{jun2019parameter},
and \Cref{thm:sp-to-olo} is the first that we are aware of. Generalizing this
result to a broader class of stochastic optimization problems is an exciting direction for future work.

\subsection{Algorithms for Parameter-free Learning}\label{section:olo-to-pfolo}

\begin{figure*}
  \centering
  \includegraphics[width=0.7\textwidth]{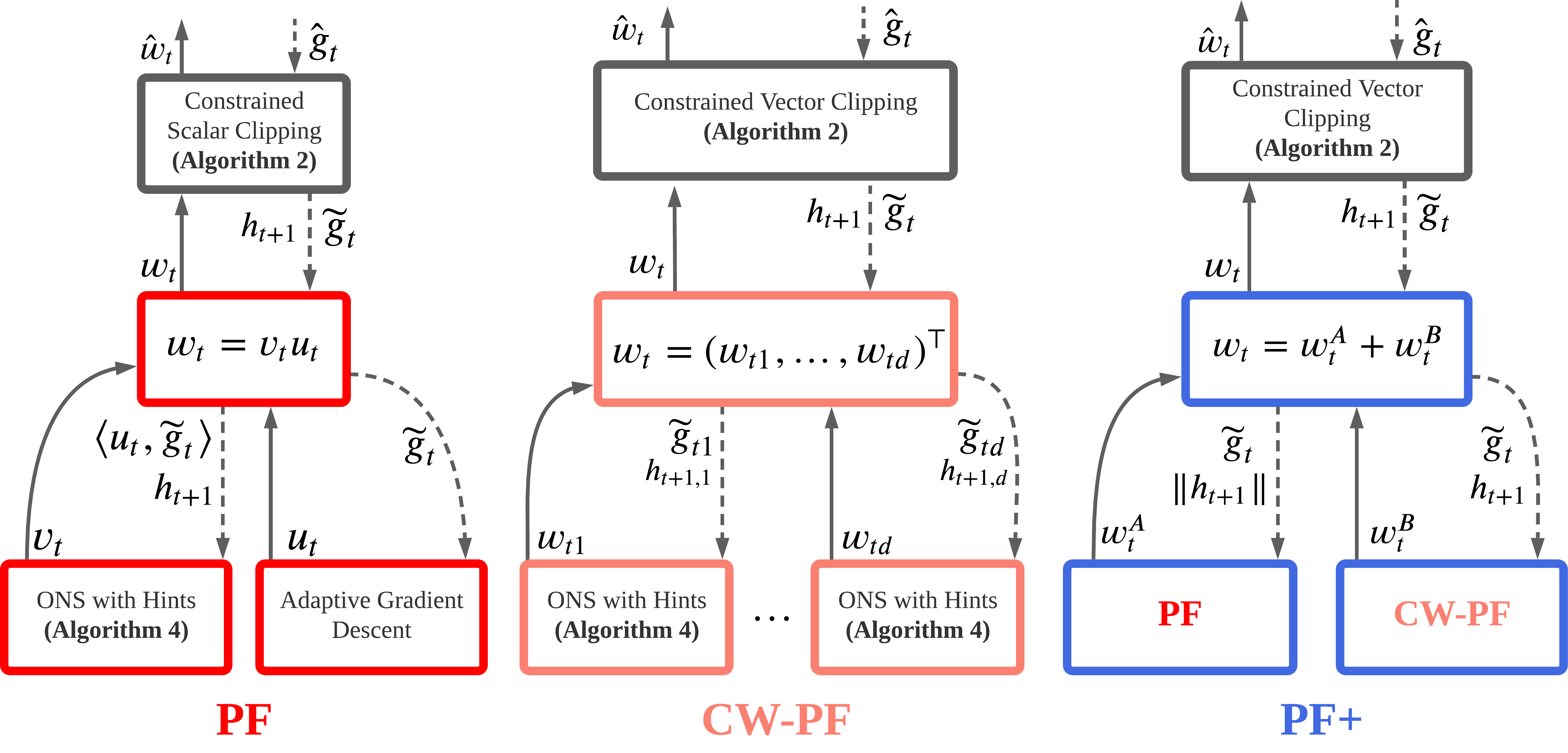}
  \caption{
    Diagrams of the three different parameter-free OLO procedures studied in this paper (best viewed in color).
    Each procedure consists of \Cref{alg:constrained-clipping},
    followed by one of three sub-procedures: PF, CW-PF, or PF+. Decisions
    made by each reduction in the hierarchy are propagated from bottom to top along the solid arrows, and feedback
    is propagated from the top down along the dashed arrows. Adaptive Gradient Descent in the PF routine
    refers to gradient descent over domain $\set{x\in\R^{d}: \norm{x}\le 1}$ with
    step-sizes
    $\eta_{t}=\sqrt{2}/2\sqrt{\sum_{\tau=1}^{t}\norm{\widetilde{g}_{\tau}}^{2}}$.
    Pseudocode is additionally provided in Appendix \ref{app:tuning-free-olo}.
  }
  \label{fig:olo-algs}

\end{figure*}

Given the reduction of the previous section, our policy evaluation problem can be solved by
any OLO algorithm which achieves an adaptive regret bound of the form $R_{T}(u)\le O\Big(\sqrt{\sum_{t=1}^{T}\norm{g_{t}}^{2}}\Big)$.
We additionally want to achieve such a bound without any hyperparameter tuning. Interestingly,
we can achieve both of these goals by instead setting our sights on a stronger bound of the form $R_{T}(u)\le O\Big(\norm{u}\sqrt{\sum_{t=1}^{T}\norm{g_{t}}^{2}}\Big)$ ---
the \textit{parameter-free} regret bound. Bounds of this form are what would be achieved if online (sub)gradient descent
could be applied with the fixed stepsize $\eta^{*}=\norm{u}/\sqrt{\sum_{t=1}^{T}\norm{g_{t}}^{2}}$, yielding the optimal
dependence on both the comparator $u$ and the sequence of subgradients $(g_{t})_{t=1}^{T}$ (see \citet[Theorem 2.13]{orabona2019modern}). Unfortunately, this step-size cannot be chosen as it involves quantities unknown to the learner in advance, and in fact it turns out that these bounds can not be
obtained without prior knowledge of $u$ \citep{orabona2018scale}.
However, several recent works make guarantees of this form \textit{up to $\log$ factors},
particularly those operating in the coin-betting framework \cite{cutkosky2018black,orabona2016coin}.

In the coin-betting framework, the objective of guaranteeing low regret $R_{T}=\sum_{t=1}^{T}\inner{g_{t},w_{t}-u}$ is re-cast as a betting problem in which the objective is to guarantee high wealth, $W_{T}\defeq W_{0}-\sum_{t=1}^{T}\inner{g_{t},w_{t}}$, starting from some initial endowment of $W_{0}$.
Concretely, in the 1-dimensional OLO case, at each time $t\in[T]$ the learner places a bet $w_{t}=\beta_{t}W_{t-1}$ --- corresponding to some (signed) fraction $\beta_{t}\in[-1,1]$ of their current wealth $W_{t-1}$ ---
on the outcome of a continuous-valued ``coin'' $-g_{t}$. The learner wins (or loses) the amount $-g_{t}w_{t}$, and their wealth becomes $W_{t}=W_{t-1}-g_{t}w_{t}=W_{t-1}(1-\beta_{t}g_{t})$.
The key observation making this a viable strategy is that any algorithm which guarantees $W_{T}\ge F(\sum_{t}-g_{t})$ for some convex $F$ also guarantees $R_{T}(u)\le F^{*}(u)+W_{0}$, where $F^{*}$
denotes the Fenchel dual of $F$ \citep{mcmahan2014unconstrained}.

To use these approaches in our problem setting, we must address some subtle issues. First,
on each round coin-betting algorithms make bets of the form $w_{t}=\beta_{t}W_{t}$ and seek to maximize the wealth $W_{T}$ ---
this makes coin-betting algorithms more naturally suited to unbounded domains, in which $\norm{w_{t}}$ can be arbitrarily large. This is
not always an appropriate assumption for our problems, in which numerical stability may be a concern in deployment. Second,
recall that in response to choosing betting fraction $\beta_{t}$, the wealth becomes $W_{t+1}=W_{t}(1-g_{t}\beta_{t})$ --- this procedure
only makes sense if the wealth is always non-negative, requiring that $\abs{\beta_{t}}\le1/\abs{g_{t}}$. Yet $\beta_{t}$ is chosen
before observing $g_{t}$, so ensuring non-negative wealth requires knowing a bound $G\ge \abs{g_{t}}$ in advance.
In practice, this is somewhat unsatisfying because such a bound is either completely unknown \textit{a priori} or is difficult to compute.

\begin{Algorithm}[t!]
  \small
  \STATE \textbf{Input:} Algorithm $\cA$ that takes hints, initial value $\gbound$, scaling function $\cM$
  \STATE \textbf{Initialize:}  initial hint $h_{1}=\gbound$
  \For{$t=1:T$}{
    \STATE Get $w_{t}$ from $\cA$
    \STATE \textbf{Play} $\hat w_{t} = \Pi_{\cW}(w_{t})$,\quad \textbf{Receive} $\hat g_{t}\in\partial\ell_{t}(\hat w_{t})$\STATE

    \STATE // Gradient Clipping (Algorithm \ref{alg:imperfect-hinting})
    \STATE Set $\gtrunc_{t}=\begin{cases} h_{t}\dfrac{\hat{g}_{t}}{\cM(\hat g_{t})} \qquad&\text{ if }\cM(\hat{g}_{t})>h_{t} \\ \hat{g}_{t}&\text{ otherwise}\end{cases}$
    \STATE \textbf{Update} $h_{t+1}=\max\brac{h_{t},\cM(\hat g_{t})}$\STATE

    \STATE // Constraint-set reduction (Algorithm \ref{alg:constraint-set-new})
    \STATE Set $S(w)=\norm{w-\Pi_{\cW}(w)}$, \quad $\widetilde w_{t}=\frac{w_{t}-\hat{w}_{t}}{\norm{w_{t}-\hat{w}_{t}}}$
    \STATE Define $\widetilde\ell_{t}(w)$ by
    \STATE$\begin{cases}
      \inner{\gtrunc_{t},w}&\text{if }\inner{\gtrunc_{t},w_{t}-\hat{w}_{t}}\ge0\\
      \inner{\gtrunc_{t},w}-\inner{\gtrunc_{t},\widetilde{w}_{t}}S(w)&\text{otherwise}
    \end{cases}$
    \STATE Compute $\widetilde g_{t}\in\partial\widetilde\ell_{t}(w_{t})$\STATE
    \STATE Send $\widetilde g_{t}$ and $h_{t+1}$ to $\cA$
  }
  \caption{Constrained Clipping}
  \label{alg:constrained-clipping}
\end{Algorithm}

In this paper, we circumvent these issues using \Cref{alg:constrained-clipping}, which is a composition of
the constraint-set reduction of \citet{cutkosky2020parameter} and the gradient clipping algorithm of \citet{cutkosky2019artificial}.\footnote{A detailed
  description of these reductions as well as the ideas sketched in the previous discussion can be found in \Cref{app:reductions}.}
Briefly, \Cref{alg:constrained-clipping} accepts as input an algorithm $\cA$ with unbounded domain. Whenever $\cA$ returns
a point outside our constraint set, $w_{t}\notin\cW$, we project $\hat{w}_{t}=\proj_{\cW}(w_{t})$ and an additional penalty is added to
the subgradient $\widetilde{g}_{t}$ sent to $\cA$ for violating the constraints. The penalty is designed in such a way that
the regret of $\cA$ on the penalized sequence is an upper bound the true regret. Next, to circumvent knowing a gradient bound,
we provide $\cA$ with ``hints'' $h_{t+1}$ specifying a bound on the next subgradient
$h_{t+1}\ge\norm{\widetilde{g}_{t+1}}$. Whenever a hint is incorrect,
we make it \textit{appear} correct to $\cA$ by passing a clipped subgradient.
\Cref{alg:constrained-clipping} controls the scaling of the gradient clipping procedure using a ``scaling function'' $\cM$.
When $\cM$ maps $\hat g_{t}\mapsto\norm{\hat g_{t}}$, we get the
regular clipping algorithm of \citet{cutkosky2019artificial}; we refer to this map as $\cM^{\text{sc}}$ and refer to the \Cref{alg:constrained-clipping} using $\cM^{\text{sc}}$ as
\textit{constrained scalar clipping}. It is straightforward to show that composition of constraint-set reduction and gradient clipping incurs no more than a constant
penalty of $G(D+\norm{\cmp})$ to the regret $R_{T}(\cmp)$ (\Cref{thm:constrained-scalar-clipping}, \Cref{app:tuning-free-olo}). In problems with sparse gradients, the hints $h_{t}\ge\max_{\tau<t}\norm{\hat g_{\tau}}$
can become quite large relative to the components $\hat g_{ti}$, and can prevent coordinate-wise,
``AdaGrad-style'' algorithms from taking full advantage of the sparsity.
In these situations, the map $M^{\text{vec}}$ which maps each $\hat g_{t}\mapsto(\abs{\hat g_{t1}},\ldots,\abs{\hat g_{td}})^{\top}$ will tend to
yield better performance in practice. We denote this map $M^{\text{vec}}$ and will refer to
\Cref{alg:constrained-clipping} with $M^{\text{vec}}$ as \textit{constrained vector clipping}.
When using the vector-valued map, the $h_{t}$ become hint \textit{vectors}, and operations on $h_{t}$ and $\cM(\hat g_{t})$ are understood
as broadcasting element-wise. The penalty incurred using this approach could be up to a factor of $d$ worse than
in the scalar clipping algorithm (see \Cref{thm:constrained-vector-clipping}, \Cref{app:tuning-free-olo}). However, as we will
see in \Cref{section:experiments}, the performance gains in problems with sparse gradients can be significant.

\subsection{From Parameter-free OLO to Policy Evaluation}\label{section:pfolo-to-policy-eval}

With the reductions of the previous sections, we can now construct practical parameter-free
algorithms for off-policy policy evaluation with linear function approximation. Each of these algorithms runs in
linear time, yet comes equipped with robust guarantees and
requires no parameter tuning to match the guarantees of GTD2 up to $\log$ factors.

To apply the SP to OLO reduction (Algorithm \ref{alg:sp-to-olo}), we need an OLO algorithm which takes
\begin{wrapfigure}{l}{0.4\columnwidth}
  \includegraphics[width=0.4\columnwidth]{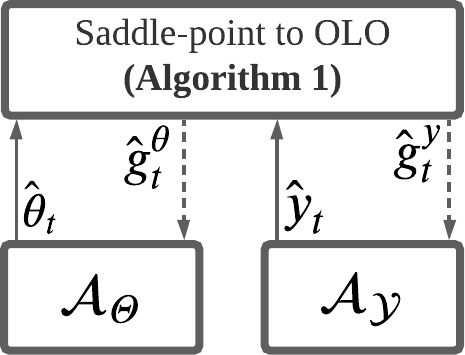}
\end{wrapfigure}
$\hat g_{t}^{\theta}\in \partial_{\theta}\hat L_{t}(\cdot, y_{t})$ and $\hat g_{t}^{y}\in-\partial_{y}\hat L_{t}(\theta_{t},\cdot)$ and returns parameters $\hat \theta_t\in\Theta$ and
$\hat y_{t}\in\cY$. We opt for a communication-free protocol in which the linear optimizations over $\Theta$ and $\cY$ are handled separately by algorithms $\cA_{\Theta}$ and $\cA_{\cY}$, as depicted inline.
Intuitively, this setup will tend to give us a better dependence on the primary quantities of interest because we are assuming the use of \textit{adaptive} algorithms. Namely,
via Cauchy-Schwarz,
\(
\norm{\theta^{*}}\sqrt{\sum\norm{\hat g^{\theta}_{t}}^{2}}+\norm{y^{*}}\sqrt{\sum\norm{\hat g_{t}^{y}}^{2}}\le \norm{z^{*}}\sqrt{\sum\norm{\hat g_{t}}^{2}}.
\)
With this setup, we construct OLO algorithms $\cA_{\cW}$ over an arbitrary compact convex domain $\cW$, which we can then apply separately as $\cA_{\Theta}$ and $\cA_{\cY}$.
Figure \ref{fig:olo-algs} illustrates three such subroutines studied in this paper.
Each of subroutine begins by applying \Cref{alg:constrained-clipping}, followed
by one of three OLO subroutines, labelled PF, CW-PF, and PF+.

The PF subroutine applies the dimension-free reduction of \citet{cutkosky2018black} (see \Cref{app:reductions}). The reduction works by decomposing the decisions $w_{t}$ into
components $v_{t}\in\R_{+}$ and $u_{t}\in\set{x\in\R^{d}:\norm{x}\le 1}$ --- representing its \textit{scale} and \textit{direction} respectively --- and playing $w_{t} = v_{t}u_{t}$.
The scale $v_t$ is handled by an instance of the ONS with Hints\footnote{Note that any
  Lipschitz-adaptive parameter-free algorithm can be
  substituted for ONS with Hints,
  such  as the scale-invariant algorithms of \citet{mhammedi2020lipschitz}.
  We focus on the ONS with Hints approach for simplicity and provide discussion
  of alternatives in \Cref{app:scale-free}}
algorithm of
\citet{cutkosky2019artificial} over domain $\R$,
and the direction $u_t$ is handled by an instance of gradient descent
with stepsizes $\eta_{t} = \frac{\sqrt{2}}{2\sqrt{\sum_{\tau=1}^{t}\norm{\widetilde{g}_{\tau}}^{2}}}$ (see Figure \ref{fig:olo-algs}, left). This reduction has the leads to a \textit{dimension-free}
regret bound of
$R_{T}(u)\le\widetilde{O}\brac{\norm{u}\sqrt{\sum_{t=1}^{T}\norm{\hat{g}_{t}}^{2}}}$,
making it well-suited to high-dimensional optimization problems.
We refer to the algorithm which uses the PF component for both $\cA_{\Theta}$ and $\cA_{\cY}$ as \pf{PFGTD}.
The following proposition shows that \pf{PFGTD} satisfies the adaptivity requirements of
\Cref{thm:sp-to-olo}, and thus matches the guarantees of GTD2 up to $\log$ terms, without
any hyperparameter tuning.
\begin{restatable}{proposition}{PFGTDRegret}\label{prop:pfgtd}
  Let $\text{PF}_{\Theta}$ and $\text{PF}_{\cY}$ be endowed with initial wealth $\half W_{0}>0$
  and $\gbound>0$. Then
  for any $z^{*}\in\Theta\times\cY$, the regret of PFGTD is bounded by
  \vspace{-2ex}
  \[
    R_{T}(z^{*})\le A_{T}(z^{*}) + B_{T}(z^{*})\sqrt{\sum_{t=1}^{T}\norm{\hat g_{t}}^{2}},
  \]
  where
  \(
  B_{T}(z^{*})\le O\Big(\norm{z^{*}}\sqrt{\Log{\frac{\norm{z^{*}} G T}{W_{0}}+1}}\Big)
  \)
  \(
  A_{T}(z^{*})\le
  O\Big(
  W_{0}+ GD+\norm{z^{*}}\Log{\frac{\norm{z^{*}} G T}{W_{0}}+1}\Big)\), and
  $G={\displaystyle\max_{t}}\norm{\hat g_{t}}$.
\end{restatable}
While PFGTD matches the guarantees of GTD2, it will fail to take advantage of possible sparsity in the gradients, which can be advantageous in the linear function approximation setting  \citep{liu2019utility}.
An alternative is the CW-PF subroutine where we play a parameter-free algorithm separately in each dimension, leading to a regret bound of the
form $R_{T}(u)\le\widetilde{O}\brac{\sum_{i=1}^{d}\abs{u_{i}}\sqrt{\sum_{t=1}^{T}\hat{g}_{ti}^{2}}}$. Coordinate-wise bounds of this form can be significantly smaller
than the dimension-free regret bound of the PFGTD algorithm when gradients are sparse. 
We refer to the algorithm using the CW-PF component for both $\cA_{\Theta}$ and $\cA_{\cY}$ as \cwpf{CW-PFGTD}.
The regret of \cwpf{CW-PFGTD} is characterized in the following proposition.
\begin{restatable}{proposition}{CWPFGTDRegret}\label{prop:cwpfgtd}
  Let $\text{CW-PF}_{\Theta}$ and $\text{CW-PF}_{\cY}$ be endowed with initial wealth $W_{0}>0$
  and $\gbound>0$.
  Then for any $z^{*}\in\Theta\times\cY$, the regret of CW-PFGTD is bounded as
  \vspace{-1ex}
  \begin{align*}
    R_{T}(z^{*})&\le
                  dA_{T} + \sum_{i=1}^{d}B_{T}^{(i)}\sqrt{\sum_{t=1}^{T}\abs{\hat g_{ti}}^{2}},
  \end{align*}
  where
  \(
  B_{T}^{(i)}\le O\brac{\norm{z^{*}}_{\infty}\sqrt{\frac{\norm{z^{*}}_{\infty}G_{\infty}T}{W_{0}}+1}}
  \),\\
  \(
  A_{T}\le O\Big(W_{0}+D_{\infty} G_{\infty}
  +\norm{z^{*}}_{\infty}\log\Big(\frac{\norm{z^{*}}_{\infty}G_{\infty}T}{W_{0}}+1\Big)\Big)
  \), and $G_{\infty}=\max_{t}\norm{\hat g_{t}}_{\infty}$.
\end{restatable}
As an immediate consequence, \cwpf{CW-PFGTD} satisfies the conditions of \Cref{thm:sp-to-olo}, but
the functions $A_{T}(z^{*})$ and $B_{T}(z^{*})$ could be up to an additional factor of $d$ larger than those of \pf{PFGTD} (\Cref{corr:cwpfgtd}, \Cref{app:tuning-free-gtd}), making this algorithm less suitable for high-dimensional problems unless one can be sure in advance that the gradients will be sufficiently sparse --- a condition which
may not be easy to quantify in practice.
Luckily,
we can guarantee the better of these two regret bounds up to constant factors by simply adding the iterates of the two algorithms together \citep{cutkosky2019combining}, as depicted in
Figure \ref{fig:olo-algs}. We refer to the algorithm which adds the iterates of PF and CW-PF routines
as PF+, and the algorithm that plays PF+ separately in $\Theta$ and $\cY$ as \pfc{PFGTD+}.
\begin{restatable}{proposition}{PFGTDPlusRegret}\label{prop:pfgtdplus}
  Let $\text{PF}$ and $\text{CW-PF}$ be endowed with initial wealth $\half W_{0}$ for
  some $W_{0}>0$. Then for any $z^{*}\in\Theta\times\cY$, the regret of PFGTD+ is bounded by
  \vspace{-1ex}
  \begin{align*}
    R_{T}(z^{*})&\le d\brac{W_{0}+2G_{\infty}(D_{\infty}+\norm{z^{*}}_{\infty})} \\
                &\qquad+ \min\set{R_{T}^{\text{PF}}(z^{*}),R_{T}^{\text{CW-PF}}(z^{*})}.
  \end{align*}
\end{restatable}

\begin{figure*}[t!]
  \centering
  \includegraphics[width=0.975\textwidth]{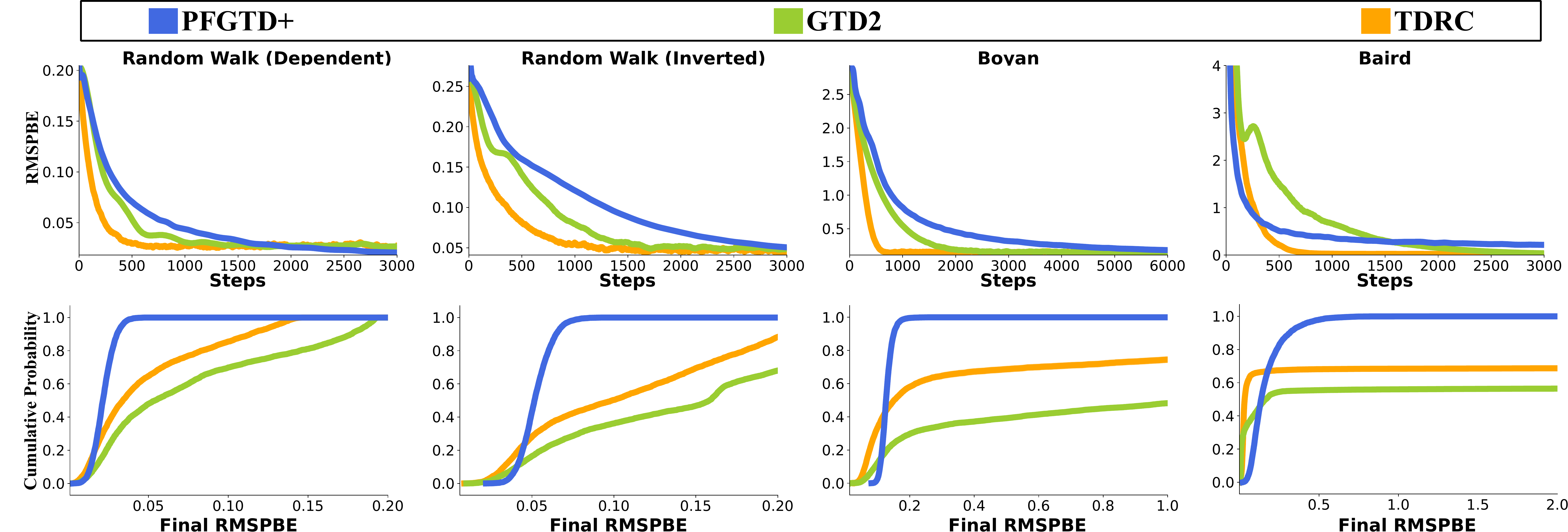}
  \caption{\textbf{Top row}: Learning curves in classic RL domains. The curves for TDC and GTD2 are with their respective best hyperparameter combinations after a grid search (see \Cref{app:experiment-details}). Each curve shows the average RMSPBE over 200 runs. Standard errors are plotted but are vanishingly small. \textbf{Bottom row}: CDF plots of the final RMSPBE for each algorithm. Uniform sampling to select hyperparameters of the baselines on each of 5,000 runs. In each plot a point $(x, y)$ means that $y$ proportion of the runs resulted in a final $\text{RMSPBE}\le x$. }
  \label{fig:classic-learning-curves-cdf}
  \vspace{-2ex}
\end{figure*}

\pfc{PFGTD+} takes advantage of sparsity when the coordinate-wise bound is the
better of the two, while still guaranteeing the dimension-free bound of the PF
subroutine. Notice that we still retain a dimension-dependent penalty as a
constant factor in either case, so
the bound is never \textit{truly} dimension-free, even when $R_{T}^{\text{PF}}\le R_{T}^{\text{CW-PF}}$.
We might instead think of the algorithm as ``almost dimension-free'', in the
sense that the undesireable dimension penalty is only on constant terms. 
Achieving both the dimension-free and coordinate-wise bounds
simultaneously has been recently achieved using a generic combiner algorithm
\citep{bhaskara2020online}, but requires solving a linear optimization
problem on each step. Instead, our approach trades this (potentially large)
computational overhead for a transient dimension penalty.
Our experiments suggest that this penalty
has little effect, with \pfc{PFGTD+} performing similarly to whichever algorithm happens to work best
in each problem.

\section{Experiments}\label{section:experiments}
We test the performance of our algorithms in two main settings: \textbf{(1)} classic MDPs used to study learning stability under function approximation and off-policy updates, and \textbf{(2)}
a prediction problem on data from a mobile robot, involving an immense number of sensory inputs, all of varying scales and levels of noise corruption.

As our parameter-free algorithms utilize the saddle-point formulation of GTD2 \citep{liu2018proximal}, GTD2 is our primary baseline.
A favourable comparison between our parameter-free methods and a well-tuned GTD2
baseline would indicate that our algorithms can be used as a drop-in replacement for GTD2, with
similar guarantees and performance in practice, yet without parameter tuning.
We additionally include TDRC \citep{ghiassian_gradient_2020}, a recently proposed extension of
TDC which is currently the state-of-the-art gradient TD method. 
Together with GTD2, these two baselines
captured the full range of performance of all other tested baselines, so
we defer additional
baselines to \Cref{app:experiment-results}.

Our aim in these experiments is not to show that our algorithms outperform the baselines --- this would be expecting too much,  as parameter-free methods achieve optimal rates only up to logarithmic terms. Our goal is rather to show that our methods can achieve performance \textit{reasonably comparable} to that of tuned baselines, with no tuning at all.
This result would indicate both a favourable performance-computation trade-off and robustness to unknown conditions.

\subsection{Classic RL Problems}

We test
our algorithms on three classic MDPs: Baird's counterexample \citep{baird1995residual}, Boyan's chain \citep{boyan2002technical}, and a five-state random walk MDP \citep{sutton2009fast}.
In Baird's counterexample, the combination of large importance sampling ratios and pathogenic generalization from the features leads TD to diverge. Boyan's chain poses similar generalization difficulties, but is an on-policy problem. The random walk MDP has been used extensively in prior benchmarks of gradient TD methods; we apply it with two
difficult feature sets: the \textit{inverted} and \textit{dependent} feature sets. Each of these problems are all well-known in our problem setting so we defer their details to \Cref{app:experiment-details}.

\Cref{fig:classic-rl} (top row) plots performance over time, measured in terms of root mean-square projected Bellman error (RMSPBE)\footnote{That is, the square root of \Cref{eq:objective} with $M = \Phi^\top D \Phi$.}, which
can be computed exactly in these problems.
From these plots we can glean a couple of trends. First, even in comparison to \textit{extensively tuned} baselines, our parameter-free algorithms still achieve comparable levels of error, without any hyperparameter tuning.
Second, we observe that \pfc{PFGTD+} initially converges more slowly than the baselines, consistent with the logarithmic penalty we pay in our regret bounds. Nevertheless, this penalty does not hamper asymptotic convergence, and indeed for all four plots, the RMSPBE of \pfc{PFGTD+} is equal to or comparable to the RMSPBEs of the baselines at the right end of the $x$-axis. 

To measure the performance in the absence of \textit{a priori} information about optimal hyperparameters, we use cumulative distribution function (CDF) plots.
Given an algorithm and a method of choosing hyperparameters on each run, CDF plots treat each run as a draw from a random variable that represents that algorithm's performance, plotting the resulting empirical CDF. For a given curve, a point $(x, y)$ means that $y$ proportion of the runs resulted in a final RMSPBE $\leq$ $x$. On each of 5,000 runs, we select the step-size of the baseline algorithms uniformly randomly, reflecting
a situation in which relevant information is missing prior to running the algorithm.
These plots also illustrate a work-performance trade off: algorithms which require substantial tuning to achieve good performance
are sensitive to the tuned parameter, and will thus have a higher proportion of poorly performing runs when the parameter is chosen randomly.

The CDF plots for each of the classic RL problems are shown in \Cref{fig:classic-rl} (bottom row). A larger proportion of the runs of our parameter-free methods achieve a low final RMSPBE compared to the baselines. For example, on Baird's counterexample, about 90\% of the runs of \pfc{PFGTD+} achieve an error $\leq$ 0.5, while only at most about 60\% of the runs of GTD2 and TDRC achieve an error $\leq$ 2. The reason for this is that achieving the optimal rates for the baselines depend upon hyperparameter search,
whose difficulty is captured by the area under the CDF curve. Indeed, error values $x$ such that $\int_{0}^{x}\text{CDF}(z)dz$ is small are values that were encountered infrequently, indicating that the parameters to achieve such error are
difficult to find. On the other hand, the CDF plots demonstrate that \pfc{PFGTD+} is less likely to achieve error as low as the very best runs of the baselines. Yet, the \pfc{PFGTD+} CDF curves
tend to rise sharply near $x=0$ before plateauing at $y=1$, indicating that \pfc{PFGTD+} reliably achieves performance not far off from the optimally-tuned baselines with no hyperparameter tuning whatsoever.

\begin{wrapfigure}{l}{0.45\columnwidth}
  \includegraphics[width=0.5\columnwidth]{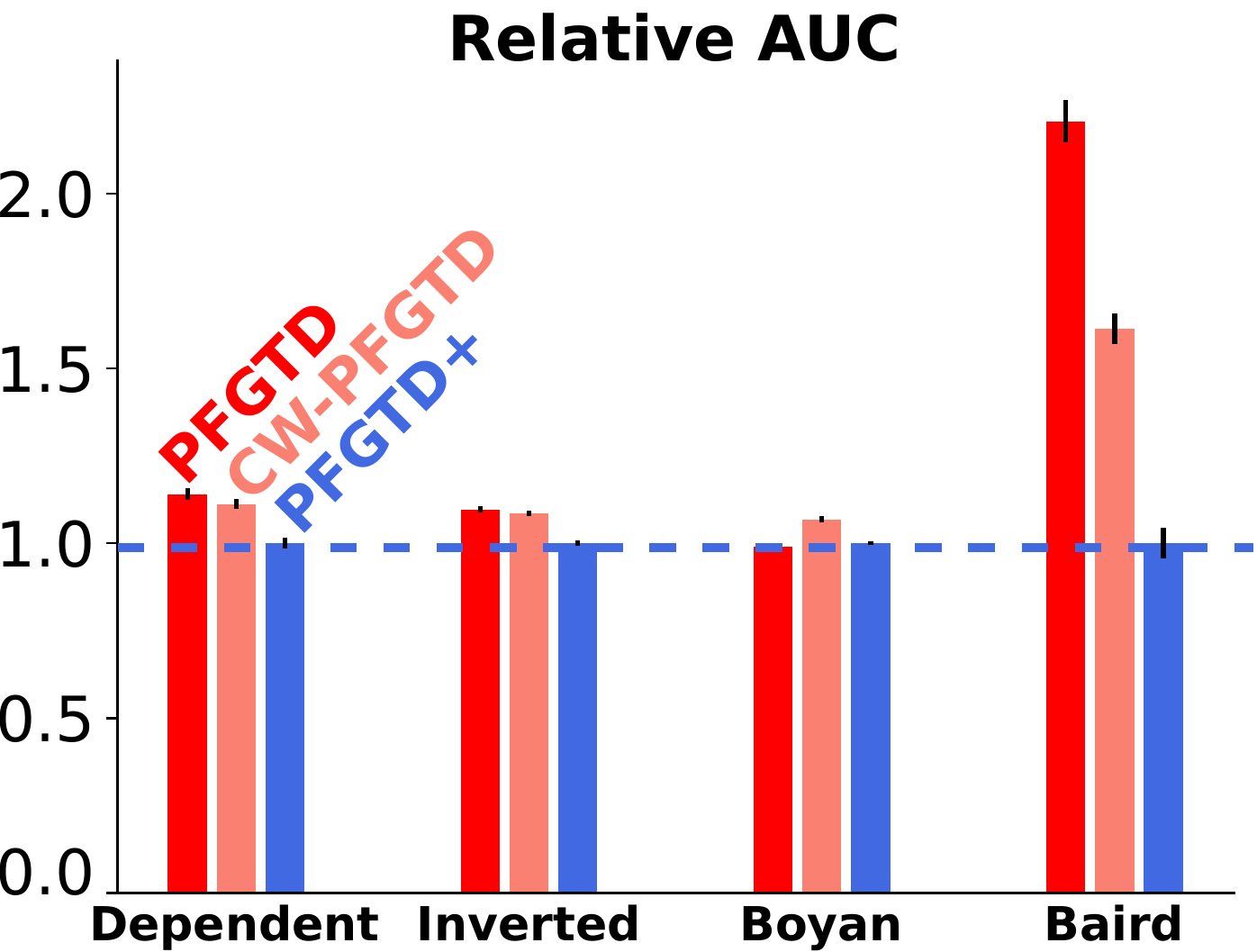}
  \vspace{-2ex}
  \label{fig:barplots-inline}
\end{wrapfigure}

Finally, the bar plot depicted in-line
shows the average area under the RMSPBE curve over 200 runs $\pm$ 1 standard error in each of the classic RL problems. The measurements are
normalized by the performance of \pfc{PFGTD+} to provide a similar scale across problems.
We observe that \pfc{PFGTD+} tends to perform at least as well
as the best of \pf{PFGTD} and \cwpf{CW-PFGTD}. 
Furthermore, with the exception of Boyan's chain, we actually see
a slight improvement over \textit{both} components,
suggesting that the iterate adding approach is itself beneficial.

\subsection{Large-scale Prediction}
In practical problems, agents can face a large array of signals, each of varying magnitudes, signal-to-noise ratios, and degrees of non-stationarity.
A learning rate that is optimal for one signal is thus unlikely to perform well for another signal, yet tuning learning rates for every single sensor can be prohibitively expensive, forcing us to compromise across all sensors.
We test performance under such conditions by recreating
the robot nexting experiment of \citet{modayil2014multi}, in which a mobile robot interacts with the environment according to a fixed behaviour policy making predictions
about future sensor readings along the way. We predict the future returns of each of the 53 sensor readings at a discount rate of $\gamma=0.9875$, corresponding to approximately 80 steps into the future.
The features are constructed from the robot's sensor readings using the same sparse representation as the original work.
Performance was measured in terms of the symmetric mean absolute percentage error (SMAPE). For a given sensor $s$, the SMAPE over the first $T$ samples is defined as $\text{SMAPE}(T, s) \defeq \frac{1}{T} \sum_{t = 1}^T  \frac{| \hat{v}^{(s)}(S_t) - G_t^{(s)}|}{|G_t^{(s)}| + |\hat{v}^{(s)}(S_t)|}.$ SMAPE is bounded in [0, 1] and scale-independent, allowing the comparison of errors across different sensors.

\begin{figure}[!htb]
  \centering
  \includegraphics[width=0.8\columnwidth]{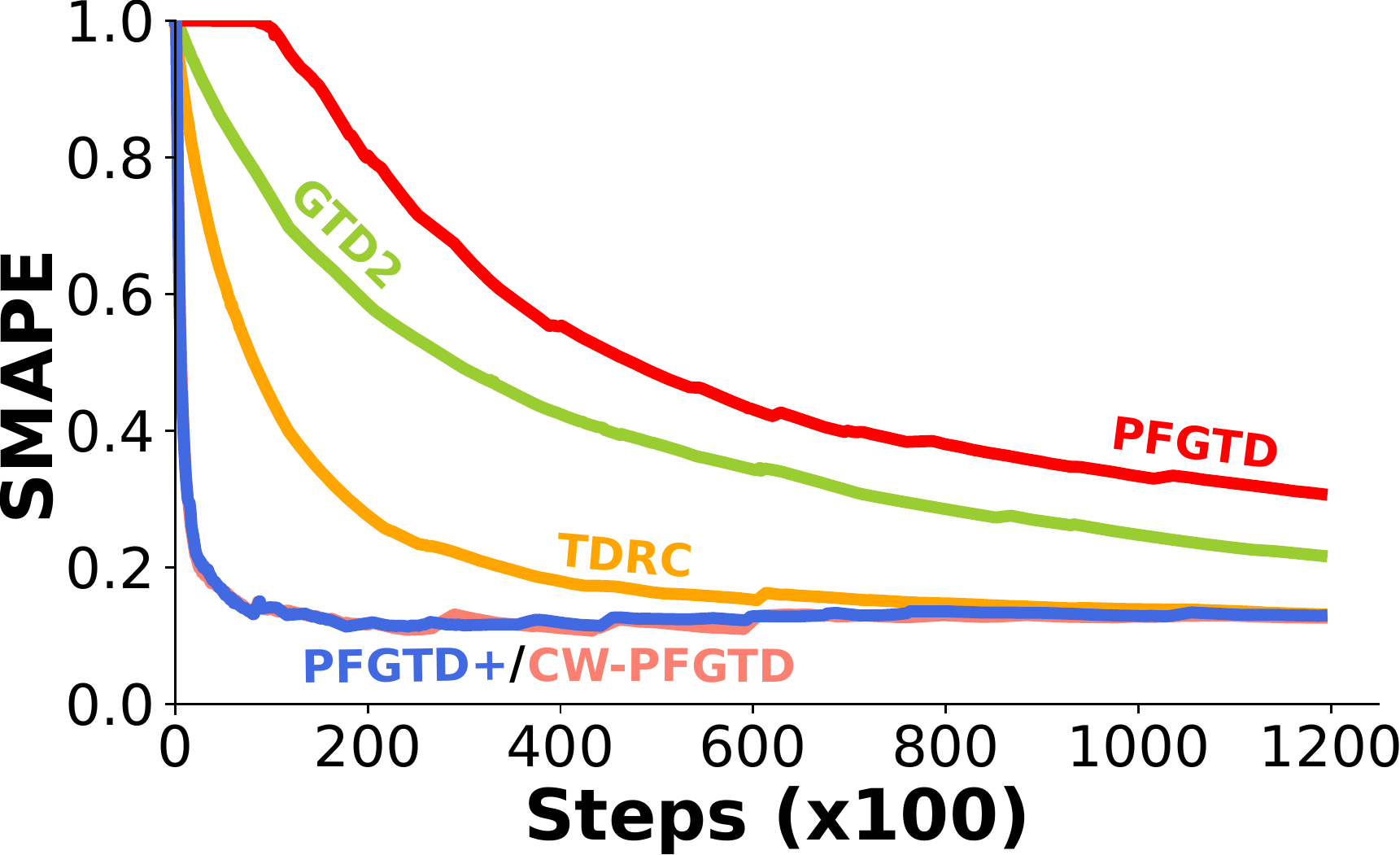}
  \caption{The median symmetric mean absolute percentage error (SMAPE) across all 53 sensors in the large-scale prediction problem. The SMAPE measures prediction accuracy compared with the ideal predictions (the true returns), computed offline using all future data. Baseline algorithms are tuned via grid search (see \Cref{app:experiment-details}).}
  \label{fig:nexting-compare-medians}
  \vspace{-2ex}
\end{figure}

The results on this task are shown in \Cref{fig:nexting-compare-medians}. As with the previous set of experiments, \pfc{PFGTD+} achieves comparable or better final error to the baselines.
Notably, despite tuning both GTD2 and TDRC, neither algorithm is able to match the convergence rate of \pfc{PFGTD+} and \cwpf{CW-PFGTD}.
In this high-dimensional prediction problem, no single step-size will perform well across all sensors due to the variation in scale and noise levels between the different sensors.
Our algorithms obviate such concerns by adapting to both the sequence of gradients and the unknown saddle-point, 
and doing so without requiring knowledge of problem-dependent constants.
The result also highlights the importance of the combined approach of \pfc{PFGTD+} --- 
the \pf{PFGTD} algorithm leaves considerable performance gains on the table by being unable to adapt to sparsity.

\vspace{-1ex}
\section{Conclusion}\label{section:conclusion}

In this paper, we bring to bear recent work in parameter-free online learning to policy evaluation. Using the Saddle-point formulation of GTD2 and the framework of parameter-free online learning through coin-betting, we derive algorithms with robust guarantees and performance
similar to that of tuned baselines, without \textit{any} parameter tuning.
While tuned baselines are capable of achieving lower error overall, 
the computational expense is large and may need to be repeated every time a
distributional shift occurs in the environment.

There are several interesting directions for future work.
First, two-timescale algorithms such as TDC and TDRC are well-known to be more performant than the algorithms derived from the Saddle-point perspective \citep{sutton2009fast,ghiassian_gradient_2020}.
Developing parameter-free variants of these algorithms is less straightforward, but could have even greater practical utility.
Second, as noted in \Cref{section:pfolo-to-policy-eval}, our proposed algorithm
never quite achieves both the dimension-free and AdaGrad regret bound simultaneously. 
Recent work has achieved such bounds with a generic combiner
algorithm \citep{bhaskara2020online}, but can require non-trivial additional computation on each step.
It remains to be seen whether removing the penalty in our approach is worth the extra computation.
Finally, future work will investigate extensions of our approach to the control setting.

{
  \bibliography{paper}
  \bibliographystyle{icml2021}

  \clearpage
  
\appendix

\section{Saddle-point to OLO Reduction}\label{app:sp-to-olo}

In this section we show how a convex-concave saddle-point problem
can be reduced to a constrained OLO problem. This reduction is included
for completeness, but similar arguments
can be seen in a variety of works which apply Mirror Descent
or other ``proximal''-type algorithms to saddle-point problems
\cite{nemirovski2009robust,mertikopoulos2018optimistic,juditsky2019algorithms}.

Let $\Theta\subset \R^{m}$ and $\cY\subset\R^{n}$ be compact convex sets
and let $L:\Theta\times\cY\rightarrow\R$ be a continuous
map such that $L(\cdot, y)$ is convex and $L(\theta,\cdot)$ concave.
We are interested in finding the saddle point of $L$, \ie, an $(\theta^*,y^*)\in\Theta\times\cY$ such that
\begin{equation}
L(\theta^*,y^*) = \inf_{\theta\in\Theta}\sup_{y\in\cY} L(\theta,y) = \sup_{y\in\cY}\inf_{\theta\in\Theta}L(\theta,y).\label{eq:minimax}
\end{equation}
With our assumptions on $\Theta$ and $\cY$ the existence of such a point $(x^{*},y^{*})\in\cZ$ follows from Sion's Minimax Theorem \cite{boyd2004convex}.
We measure the quality of a candidate solution
$z=(\theta,y)$ via the Duality gap:
\begin{align}
\max_{y^{*}\in\cY}L(\theta,y^{*}) - \min_{\theta^{*}\in\Theta}L(\theta^{*},y)\label{eq:gap}
\end{align}
It can be shown that this quantity is non-negative, and
any $(\theta,y)\in\Theta\times\cY$ for which
$\max_{y^{*}\in\cY}L(\theta,y^{*})-\min_{\theta^{*}\in\Theta}L(\theta^{*},y)=0$ must be a saddle-point
of SP problem \ref{eq:sp} (shown as an aside in Box \ref{box:no-gap-is-sp}).
Finally, it can be shown that the saddle point of our saddle point problem, Equation \ref{eq:sp},
corresponds to a minimizer of the MSPBE (or NEU) \cite{liu2018proximal}.

Next, we show that this duality gap can be bounded in terms of the regrets incurred by two separate online learning algorithms.
Consider
points $\thetabar_{T}=\frac{1}{T}\sum_{t=1}^{T}\theta_{t}$ and
$\ybar_{T}=\sum_{t=1}^{T}y_{t}$ for sequences with $\theta_{t}\in\Theta$ and $y_{t}\in\cY$ chosen by
online learning algorithms $\cA_{\Theta}$ and $\cA_{\cY}$ respectively.
Let $(\theta^{*},y^{*})\in\Theta\times\cY$.
Using the convexity of $L$ in its first argument and $-L$ in its second argument, we can write
\begin{align*}
  &L(\thetabar_{T},y^{*})-L(\theta^{*},\ybar_{T})\\
  &\quad\le
    \frac{1}{T}\sum\limits_{t=1}^{T}L(\theta_{t},y^{*}) - \frac{1}{T}\sum_{t=1}^{T}L(\theta^{*},y_{t})\\
  &\quad=
    \frac{\sum\limits_{t=1}^{T}L(\theta_{t},y^{*})-L(\theta^{*},y_{t})}{T}\\
  &\quad=
    \frac{\sum\limits_{t=1}^{T}L(\theta_{t},y^{*})-L(\theta_{t},y_{t})+L(\theta_{t},y_{t})-L(\theta^{*},y_{t})}{T}\\
  &\quad\overset{(a)}{\le}
    \frac{\sum\limits_{t=1}^{T}\inner{g_{t}^{y}, y_{t}-y^{*}}
    +\sum\limits_{t=1}^{T} \inner{g_{t}^{\theta},\theta_{t}-\theta^{*}}}{T}\\
\end{align*}
where
$(a)$ defined $g_{t}^{\theta}\in\partial_{\theta}L(\theta_{t},y_{t})$, $g_{t}^{y}\in-\partial_{y}L(\theta_{t},y_{t})$ and used of convexity of
$L(\cdot, y_{t})$ and $-L(\theta_{t},\cdot)$. Observe that the quantities in the last line can be seen as average regrets $\frac{1}{T}R_{T}^{\cA_{\Theta}}$ and $\frac{1}{T}R_{T}^{\cA_{\cY}}$ of two algorithms $\cA_{\Theta}$ and $\cA_{\cY}$ played against the loss sequences $\theta\mapsto\inner{g_{t}^{\theta},\theta}$ and $y\mapsto\inner{g_{t}^{y},y}$ respectively.
Thus we can ensure that the duality gap approaches zero by providing OLO algorithms $\cA_{\Theta}$ and $\cA_{\cY}$ which guarantee sublinear regret. Thus we have the following
``Folk Theorem'', commonly used to reduce min-max problems to separate OLO problems:
\begin{proposition}[Folk Theorem]
  Let $\cA_{\Theta}$ and $\cA_{\cY}$ be online learning algorithms with compact and convex domains $\Theta$ and $\cY$ respectively.
  Let $R_{T}^{\cA_{\Theta}}(\theta^{*})$ and $R_{T}^{\cA_{\cY}}(y^{*})$ denote the linear regret of algorithms
  $\cA_{\Theta}$ and $\cA_{\cY}$.
  Then for any $(\theta^{*},y^{*})\in\Theta\times\cY$,
  \[
    L\brac{\overline{\theta}_{T},y^{*}} - L\brac{\theta^{*}, \overline{y}_{T}}\le\frac{R_{T}^{\cA_{\Theta}}(\theta^{*})+R_{T}^{\cA_{\cY}}(y^{*})}{T}.
  \]
  \label{prop:sp-to-olo}
\end{proposition}
\begin{Boxed}{${\max_{y^{*}\in\cY}L(\theta,y^{*})-\min_{\theta^{*}\in\Theta}L(\theta^{*},y)}$}
Letting $(\theta,y)$ be arbitrary feasible points, the following holds via strong duality:
\begin{align*}
  \min_{\theta^{*}\in\Theta}L(\theta^{*},y)&\le\min_{\theta^{*}\in\Theta}\max_{y^{*}\in\cY}L(\theta^{*},y^{*})\\
  &=\max_{y^{*}\in\cY}\min_{\theta^{*}\in\Theta}L(\theta^{*},y^{*})\le\max_{y^{*}\in\cY}L(\theta,y^{*}),
\end{align*}
and so subtracting $\min_{\theta^{*}\in\Theta}L(\theta^{*},y)$ from both sides:
\[
  0 \le\max_{y^{*}\in\cY}L(\theta,y^{*}) - \min_{\theta^{*}\in\Theta}L(\theta^{*},y).
\]
Furthermore, any point $(\theta, y)$ such that $\max_{y^{*}\in\cY}L(\theta,y^{*})-\min_{\theta^{*}\in\Theta}L(\theta^{*},y)=0$
must be a saddle point. To see this, write
\[
  L(\theta, y) \le \max_{y^{*}\in\cY}L(\theta,y^{*})=\min_{\theta^{*}\in\Theta}L(\theta^{*},y)
\]
and
\[
  L(\theta,y)\ge\min_{\theta^{*}\in\Theta}L(\theta^{*},y)=\max_{y^{*}\in\cY}L(\theta, y^{*}),
\]
which together imply
\[
  \min_{\theta^{*}\in\Theta}L(\theta^{*},y)\le L(\theta,y)\le\max_{y^{*}\in\cY}L(\theta,y^{*}),
\]
showing that $(\theta,y)$ is a saddle-point by definition.\hfill$\blacksquare$
\labelbox{box:no-gap-is-sp}
\end{Boxed}
Note that we can equivalently state the above in terms of the regret of a single algorithm with domain $\cZ=\Theta\times\cY$. Namely, by letting
$g_{t}\defeq(g_{t}^{\theta},g_{t}^{y})$, $z_{t}\defeq(\theta_{t}, y_{t})$, and $z^{*}=(\theta^{*},y^{*})$, we can write
\begin{align*}
  L\brac{\overline{\theta}_{T},y^{*}} - L\brac{\theta^{*}, \overline{y}_{T}}&\le
  \frac{\sum\limits_{t=1}^{T}\inner{g_{t}, z_{t}-z^{*}}}{T}
= \frac{R_{T}^{\cA_{\cZ}}(z^{*})}{T}.
\end{align*}
This form will sometimes be a bit easier to work with due to requiring less notation.

The Folk Theorem extends straight-forwardly to the stochastic setting using a simple online-to-batch conversion \citep{cesa2004generalization,orabona2019modern}
argument.
\begin{proposition}
  Let $\cA_{\Theta}$ and $\cA_{\cY}$ be online learning algorithms with domains $\Theta$ and $\cY$ respectively.
  Suppose $(\ghat_{t}^{\theta})_{t=1}^{T}$ and $(\ghat_{t}^{y})_{t=1}^{T}$ are sequences satisfying $\EE{\hat g_{t}^{\theta}|\theta_{t},y_{t}}\in\partial_{\theta}L(\theta_{t},y_{t})$
  and $\EE{\hat g_{t}^{y}|\theta_{t},y_{t}} \in-\partial_{y}L(\theta_{t},y_{t})$ for all $t\in[T]$. Then
  for any $(\theta^{*},y^{*})\in\Theta\times\cY$,
  \[
    \EE{L\brac{\overline{\theta}_{T},y^{*}} - L\brac{\theta^{*}, \overline{y}_{T}}}\le
    \frac{\EE{R_{T}^{\cA_{\Theta}}(\theta^{*})+R_{T}^{\cA_{\cY}}(y^{*})}}{T},
  \]
  where $R_{T}^{\cA_{\Theta}}(\theta^{*})=\sum_{t=1}^{T}\inner{\hat g_{t}^{\theta},\theta_{t}-\theta^{*}}$ and $R_{T}^{\cA_{\cY}}(y^{*})=\sum_{t=1}^{T}\inner{\hat g_{t}^{y},y_{t}-y^{*}}$.
  \label{prop:stochastic-sp-to-olo}
\end{proposition}
The proof follows easily by using convexity followed by the tower rule:
\begin{align*}
  &\EE{L(\thetabar_{T},y^{*})-L(\theta^{*},\ybar_{T})}\\
  &\qquad\le \frac{\EE{\sum_{t=1}^{T}\inner{g_{t}^{\theta},\theta_{t}-\theta^{*}}+ \sum_{t=1}^{T}\inner{g_{t}^{y}, y_{t}-y^{*}}}}{T}\\
  &\qquad= \frac{\EE{\sum_{t=1}^{T}\inner{\hat g_{t}^{\theta},\theta_{t}-\theta^{*}}+ \sum_{t=1}^{T}\inner{\hat g_{t}^{y}, y_{t}-y^{*}}}}{T}\\
  &\qquad= \EE{\frac{R_{T}^{\cA_{\Theta}}(\theta^{*})+R_{T}^{\cA_{\cY}}(y^{*})}{T}}
\end{align*}

Using this simple reduction, we can see that our
problem of minimizing the MSPBE (or NEU) can be reduced to two separate OLO problems; if we can find algorithms
which guarantee sublinear regret on the stochastic linear losses $\ghat_{t}^{\theta}$ and $\ghat_{t}^{y}$ given by Equations \ref{eq:noisy-subgrads}, then
the duality gap (and thus MSPBE) will approach zero in expectation $T\rightarrow\infty$. In the following sections
we can thus focus our attention on constructing OLO algorithms with various desirable properties.

\section{Proof of Theorem \ref{thm:sp-to-olo}}\label{app:highprob}

The theorem is repeated below for for convenience.

\AdaptiveBound*

\textit{Proof:}
Let $\hatg_{t}^{\theta}$ and $\hatg_{t}^{y}$ be such that $\EE{\hatg_{t}^{\theta}|\theta_{t},y_{t}} = g_{t}^{\theta}\in\partial_{\theta}L(\theta_{t},y_{t})$ and
$\EE{\hatg_{t}^{y}|\theta_{t},y_{t}}=g_{t}^{y}\in\partial_{y}L(\theta_{t},y_{t})$ respectively.
Further, endow $\Theta\times\cY$ with norm $\norm{z}=\norm{(\theta,y)}=\sqrt{\norm{\theta}^{2}+\norm{y}^{2}}$ and define $z_{t}\defeq(\theta_{t},y_{t})$ and $\hatg_{t}\defeq (\hatg_{t}^{\theta}, -\hatg_{t}^{y})$.
Following Appendix \ref{app:sp-to-olo} we have that for saddle-point $z^{*}\defeq(\theta^{*},y^{*})\in\Theta\times\cY$,
\begin{align}
  &\EE{L(\thetabar_{T},y^{*})-L(\theta^{*},\ybar_{T})}
      \le
      \EE{\frac{R_{T}^{\cA}(z^{*})}{T}}\nonumber\\
    &\qquad\overset{(a)}{\le}\EE{\frac{A_T(z^{*})+B_{T}(z^{*})\sqrt{\sum_{t=1}^{T}\norm{\hatg_{t}}^{2}}}{T}}\nonumber\\
  &\qquad\overset{(b)}{\le}\frac{A_T(z^{*})+B_{T}(z^{*})\sqrt{\sum_{t=1}^{T}\EE{\norm{\hatg_{t}}^{2}}}}{T}\label{ineq:Ebound1}
\end{align}
\begin{Boxed}{$\EE{\norm{\ghat_{t}}^{2}}\le G^{2}$}
  Here we derive an upper bound for $\EE{\norm{\ghat_{t}}^{2}}$.
  Following similar steps to \citet{liu2018proximal}, we have that
    \begin{align*}
      &\EE{\norm{\hatg_{t}}^{2}}
      =
        \EE{\norm{\ghat_{t}-g_{t}+g_{t}}^{2}}\\
        &\quad= \EE{\norm{\ghat_{t}-g_{t}}^{2}} + \norm{g_{t}}^{2}
      \le
        \sigma^{2} + \norm{g_{t}^{\theta}}^{2} + \norm{g_{t}^{y}}^{2}\\
      &\quad=
        \sigma^{2} + \norm{A^{\top}y_{t}}^{2} + \norm{A\theta_{t}-b+My_{t}}^{2}\\
      &\quad\le
        \sigma^{2} + \norm{A}^{2}D^{2} + \brac{\norm{A}D+\norm{b}+\lambda_{\max}(M)D}^{2}\\
      &\quad\le
        \underbrace{D^{2}\brac{2\norm{A} + \lambda_{\max}(M) + \frac{\sigma + \norm{b}}{D}}^{2}}_{G^{2}},
    \end{align*}
    where $\sigma^{2}\defeq\sigma_{\theta}^{2}+\sigma_{y}^{2}$ and the last line uses the elementary inequality $a^{2}+b^{2}+c^{2}\le(a+b+c)^{2}$.
    \labelbox{box:Enormgt}
\end{Boxed}
where $(a)$ used the regret guarantee of $\cA$ applied to the
loss sequence $z_t\mapsto\inner{\ghat_t,z_t}$, and $(b)$ applied Jensen's inequality \wrt the function $x\mapsto\sqrt{x}$.
As shown in Box \ref{box:Enormgt}, the terms $\EE{\norm{\ghat_{t}}^{2}}$ are bounded above by $G^{2}\defeq D^{2}\brac{2\norm{A} + \lambda_{\max}(M)+\frac{\sigma+\norm{b}}{D}}^{2}$, so (\ref{ineq:Ebound1}) becomes
\begin{align*}
  &\EE{\frac{L(\thetabar_{T},y^{*})-L(\theta^{*},\ybar_{T})}{T}}\\
  &\qquad\le\frac{A_T(z^{*})+B_{T}(z^{*})\sqrt{TG^{2}}}{T}\\
  &\qquad=\frac{A_T(z^{*})}{T}+\frac{B_{T}(z^{*})D\brac{2\norm{A} + \lambda+\frac{\sigma+\norm{b}}{D}}}{\sqrt{T}},
\end{align*}
with $\lambda=\lambda_{\max}(M)$.

For the high probability statement, define $G^{2}$ as before and assume the
following stronger assumption\footnote{This stronger assumption is standard when moving to
  high-probability bounds, and is used in \citet{liu2018proximal} and the work their
  result is based on, \citet{nemirovski2009robust}} holds:
\begin{align}
  \EE{\Exp{\frac{\norm{\hatg_{t}}^{2}}{G^{2}}}}\le\Exp{1}.\label{ineq:light-tailed}
\end{align}
Similarly to the expectation bound, we begin by bounding the duality gap by the
regret of $\cA$:
\begin{align*}
  &L(\thetabar_{T},y^{*})-L(\theta^{*},\ybar_{T})\le\frac{\sum_{t=1}^{T}\inner{g_{t},z_{t}-z^{*}}}{T}\\
  &\qquad=\frac{\sum_{t=1}^{T}\inner{\hatg_{t},z_{t}-z^{*}} + \inner{g_{t}-\ghat_{t},z_{t}-z^{*}}}{T}\\
  &\qquad=\underbrace{\frac{\sum_{t=1}^{T}\inner{\hatg_{t},z_{t}-z^{*}}}{T}}_{\encircle{1}} + \underbrace{\frac{\sum_{t=1}^{T}\inner{g_{t}-\ghat_{t},z_{t}-z^{*}}}{T}}_{\encircle{2}}\\
\end{align*}

\begin{Boxed*}{$\Exp{-\eta\eps}\EE{\Exp{\eta \sqrt{\sum_{t=1}^{T}\norm{\ghat_{t}}^{2}}}}\le\Exp{\frac{-\eps}{\sqrt{T}G}+2}$}
  Let $\eta=\frac{1}{\sqrt{T}G}$, then
  \begin{align*}
    \EE{\Exp{\eta \sqrt{\sum_{t=1}^{T}\norm{\ghat_{t}}^{2}}}}
      &= \EE{\Exp{\sqrt{\frac{\sum_{t=1}^{T}\norm{\ghat_{t}}^{2}}{TG^{2}}}}}
      \le \EE{\Exp{\sqrt{1+\frac{\sum_{t=1}^{T}\norm{\ghat_{t}}^{2}}{TG^{2}}}}}\\
      &\le \EE{\Exp{1+\frac{\sum_{t=1}^{T}\norm{\ghat_{t}}^{2}}{TG^{2}}}}
    =
      \Exp{1}\EE{\Exp{\frac{\sum_{t=1}^{T}\norm{\ghat_{t}}^{2}}{TG^{2}}}}\\
    &\overset{(a)}{\le}\Exp{1}\EE{\frac{1}{T}\sum_{t=1}^{T}\Exp{\frac{\norm{\ghat_{t}}^{2}}{G^{2}}}}
    \overset{(b)}{\le}\Exp{1}\frac{1}{T}\sum_{t=1}^{T}\Exp{1} = \Exp{2},
  \end{align*}
  where $(a)$ applied Jensen's inequality \wrt $x\mapsto\Exp{x}$, and $(b)$ applied the
  light-tailed assumption $T$ times. Combining this with $\Exp{-\eps\eta}=\Exp{\frac{-\eps}{\sqrt{T}G}}$, we have
  \begin{align*}
    \Exp{-\eta\eps}\EE{\Exp{\eta \sqrt{\sum_{t=1}^{T}\norm{\ghat_{t}}^{2}}}}
    \le \Exp{\frac{-\eps}{\sqrt{T}G}+2}
  \end{align*}
  \labelbox{box:light-tailed-mgf}
\end{Boxed*}

\textbf{Bounding \encircle{1}} Using the regret guarantee of $\cA$, we again have
\begin{align*}
  \frac{\sum_{t=1}^{T}\inner{\hatg_{t},z_{t}-z^{*}}}{T} \le\frac{A_{T}(z^{*})+B_{T}(z^{*})\sqrt{\sum_{t=1}^{T}\norm{\hatg_{t}}^{2}_{*}}}{T},
\end{align*}
where $A_{T}(\cdot)$ and $B_{T}(\cdot)$ are deterministic non-negative functions.
To bound this term we can use a Cramer-Chernoff-style argument.
Let $\eps>0$, and consider $\Prob{\sqrt{\sum_{t=1}^{T}\norm{\ghat_{t}}^{2}}>\eps}$.
Letting $\eta>0$ be a variable to be sorted out later, we have
\begin{align*}
  &\Prob{\sqrt{\sum_{t=1}^{T}\norm{\ghat_{t}}^{2}}\ge\eps}\\
  &=
    \Prob{\Exp{\eta \sqrt{\sum_{t=1}^{T}\norm{\ghat_{t}}^{2}}} \ge \Exp{\eta\eps}}\\
  &\le
    \Exp{-\eta\eps}\EE{\Exp{\eta \sqrt{\sum_{t=1}^{T}\norm{\ghat_{t}}^{2}}}},
\end{align*}
with the last line resulting from
Markov's inequality. Next, we notice that the second term bears some resemblance to the terms in
the light-tailed assumption (\ref{ineq:light-tailed}) which are bounded above by $\Exp{1}$, so a natural approach would be
to try to set $\eta$ to expose terms of this form. Indeed, as shown in Box \ref{box:light-tailed-mgf},
if we set $\eta = \frac{1}{\sqrt{T}G}$, then
\begin{align*}
  &\Exp{-\eta\eps}\EE{\Exp{\eta \sqrt{\sum_{t=1}^{T}\norm{\ghat_{t}}^{2}}}}\\
  &\scratch{\qquad\le\Exp{\frac{-\eps}{\sqrt{T}G}}\Exp{1}}\\
  &\qquad\le\Exp{\frac{-\eps}{\sqrt{T}G}+2},
\end{align*}
and so the deviation bound for $\encircle{1}$ is
\begin{align}
  &\Prob{\frac{A_{T}(z^{*})+B_{T}(z^{*})\sqrt{\sum_{t=1}^{T}\norm{\ghat_{t}}^{2}}}{T}\ge \frac{A_{T}(z^{*})+B_{T}^{*}(z^{*})\eps}{T}}\nonumber\\
  &\qquad\le
    \Exp{\frac{-\eps}{\sqrt{T}G}+2}
\end{align}

\newcommand{\epsp}{\eps}

\textbf{Bounding \encircle{2}} This term requires some additional trickery but the basic idea is
still the same. As before, introduce a variable $\eta>0$ to be set later and
let $\epsp>0$. Then
\begin{align}
  &\Prob{\sum_{t=1}^{T}\inner{g_{t}-\ghat_{t},z_{t}-z^{*}}\ge\epsp}\nonumber\\
  &=
    \Prob{\Exp{\eta \sum_{t=1}^{T}\inner{g_{t}-\ghat_{t},z_{t}-z^{*}}} \ge \Exp{\eta\epsp}}\nonumber\\
  &\le
    \Exp{-\eta\epsp}\EE{\Exp{\eta\sum_{t=1}^{T}\inner{g_{t}-\ghat_{t},z_{t}-z^{*}}}}\label{ineq:circle2prob}.
\end{align}
Next, define the following short-hand notations
\begin{align*}
\xi_{t}&\defeq\inner{g_{t}-\ghat_{t},z_{t}-z^{*}}\\
\cF_{t}&\defeq(s_{\tau},a_{\tau},r_{\tau},s_{\tau}',\theta_{\tau},y_{\tau})_{\tau=1}^{t}
\end{align*}
\begin{Boxed*}{$\EE{\Exp{\eta\xi_{t}}\Big|\cF_{t-1}}\le\Exp{4\eta^{2}G^{2}D^{2}}$}
  The trick here is to find ways to introduce terms of the form $\frac{\norm{g_{t}-\ghat_{t}}^{2}}{4G^{2}}$, which we could bound as
  $\frac{\norm{g_{t}-\ghat}^{2}}{4G^{2}}\le\frac{2G^{2}+2\norm{\ghat_{t}}^{2}}{4G^{2}}$. Then using Jensen's inequality and the light-tailed assumption, we'd have
  \begin{align*}
    \EE{\Exp{\frac{\norm{g_{t}-\ghat_{t}}^{2}}{4G^{2}}}}&\le\EE{\Exp{\frac{2G^{2}+2\norm{\ghat_{t}}^{2}}{4G^{2}}}}=\EE{\Exp{\half\brac{1+\frac{\norm{\ghat_{t}}^{2}}{G^{2}}}}}\le\EE{\Exp{1}\Exp{\frac{\norm{\ghat_{t}}^{2}}{G^{2}}}}^{\half}\\
    &\le\Exp{\half}\Exp{\half}=\Exp{1}.
  \end{align*}
  With this in mind, consider again $\EE{\Exp{\eta\xi_{t}}\Big|\cF_{t-1}}$.
  First, if we assume that $4\eta^{2}G^{2}D^{2}\le 1$, then
  \begin{align*}
    \EE{\Exp{\eta\xi_{t}}\Big|\cF_{t-1}}
    &\overset{(a)}{\le}
    \EE{\eta\xi_{t}\Big|\cF_{t-1}}+\EE{\eta\Exp{\eta^{2}\xi_{t}^{2}}\Big|\cF_{t-1}}
    \overset{(b)}{=}
    \EE{\Exp{\eta^{2}\xi_{t}^{2}}\Big|\cF_{t-1}}\\
    &\le\EE{\Exp{\eta^{2}D^{2}\norm{g_{t}-\ghat_{t}}^{2}}\Big|\cF_{t-1}}
      =\EE{\Exp{4\eta^{2}D^{2}G^{2}\frac{\norm{g_{t}-\ghat_{t}}^{2}}{4G^{2}}}\Big|\cF_{t-1}}\\
    &=\EE{\Exp{\frac{\norm{g_{t}-\ghat_{t}}^{2}}{4G^{2}}}^{4\eta^{2}D^{2}G^{2}}\Big|\cF_{t-1}}
      \overset{(c)}{\le}\EE{\Exp{\frac{\norm{g_{t}-\ghat_{t}}^{2}}{4G^{2}}}\Big|\cF_{t-1}}^{4\eta^{2}D^{2}G^{2}}\\
    &\le\Exp{1}^{4\eta^{2}D^{2}G^{2}} = \Exp{4\eta^{2}D^{2}G^{2}}
  \end{align*}
  where $(a)$ used the elementary inequality $\Exp{x}\le x + \Exp{x^{2}}$, $(b)$ used that $\EE{\xi_{t}\Big|\cF_{t-1}}=0$,
  and $(c)$ used the assumption that $4\eta^{2}G^{2}D^{2}\le 1$ to apply Jensen's inequality \wrt $x\mapsto x^{4\eta^{2}G^{2}D^{2}}$.
  Otherwise, if $4\eta^{2}G^{2}D^{2}>1$, we can get the same upper bound as follows:
  \begin{align*}
    \EE{\Exp{\eta\xi_{t}}\Big|\cF_{t-1}}
    &=
      \EE{\Exp{\inner{\eta\brac{g_{t}-\ghat_{t}},z_{t}-z^{*}}\Big|\cF_{t-1}}}
      \overset{(d)}{\le}
      \EE{\Exp{\frac{\eta^{2}\norm{g_{t}-\ghat_{t}}^{2}}{2\rho}+\frac{\rho D^{2}}{2}}\Big|\cF_{t-1}}\\
      &\overset{(e)}{=}\EE{\Exp{\frac{\norm{g_{t}-\ghat_{t}}^{2}}{4G^{2}}+4\eta^{2}G^{2}D^{2}}^{\half}\Big|\cF_{t-1}}
      \overset{(f)}{=}\sqrt{\Exp{4\eta^{2}G^{2}D^{2}}\EE{\Exp{\frac{\norm{g_{t}-\ghat_{t}}^{2}}{4G^{2}}}\Big|\cF_{t-1}}}\\
    &\overset{(g)}{\le}\sqrt{\Exp{2\cdot4\eta^{2}G^{2}D^{2}}}=\Exp{4\eta^{2}G^{2}D^{2}},
  \end{align*}
  where $(d)$ applied the Fenchel-Young inequality with $\rho>0$, $(e)$ factored out $\half$ to give $\Exp{\cdot}^{\half}$ and chose $\rho=4\eta^{2}G^{2}$, and $(f)$
  applied Jensen's inequality \wrt $x\mapsto\sqrt{x}$, and $(g)$ used that $1<4\eta^{2}G^{2}D^{2}$.
  Thus, in either case we have an upper bound of $\Exp{4\eta^{2}G^{2}D^{2}}$.
  \labelbox{box:mgf-bound}
\end{Boxed*}
and observe that $\xi_{t}$ is a deterministic function of $\cF_{t}$, and that $\EE{\xi_{t}|\cF_{t-1}}=0$. Thus, we can write
\begin{align*}
  &\EE{\Exp{\eta\sum_{t=1}^{T}\inner{g_{t}-\ghat_{t},z_{t}-z^{*}}}} \\
  &\qquad=
    \EE{\Exp{\eta\sum_{t=1}^{T}\xi_{t}}}\\
  &\qquad=
    \EE{\EE{\Exp{\eta\sum_{t=1}^{T}\xi_{t}}\Big|\cF_{T-1}}}\\
  &\qquad=
    \EE{\Exp{\eta\sum_{t=1}^{T-1}\xi_{t}}\EE{\Exp{\eta\xi_{T}}\Big|\cF_{T-1}}}.
\end{align*}
As we show in Box \ref{box:mgf-bound}, for each $t$ we have that $\EE{\Exp{\eta\xi_{t}}\Big|\cF_{t-1}}\le\Exp{4\eta^{2}G^{2}D^{2}}$,
so
\begin{align*}
  &\EE{\Exp{\eta\sum_{t=1}^{T-1}\xi_{t}}\EE{\Exp{\eta\xi_{T}}\Big|\cF_{T-1}}}\\
  &\le
    \EE{\Exp{\eta\sum_{t=1}^{T-1}\xi_{t}}\Exp{4\eta^{2}G^{2}D^{2}}}\\
  &\le\ldots\le\Exp{4\eta^{2}G^{2}D^{2}T}.
\end{align*}
Plugging this back into (\ref{ineq:circle2prob}) gives us
\begin{align*}
  &\Prob{\sum_{t=1}^{T}\inner{g_{t}-\ghat_{t},z_{t}-z^{*}}\ge\epsp}\\
  &\qquad\le
    \Exp{-\eta\epsp + 4\eta^{2}G^{2}D^{2}T}
  =
    \Exp{\frac{-\epsp^{2}}{16TG^{2}D^{2}}},
\end{align*}
where in the final equality we set $\eta=\frac{\eps}{8G^{2}D^{2}T}$ to balance the terms.
Thus for $\encircle{2}$ we have a deviation bound of
\begin{align*}
  \Prob{\frac{\sum_{t=1}^{T}\inner{g_{t}-\ghat_{t},z_{t}-z^{*}}}{T}>\frac{\eps}{T}}\le\Exp{\frac{-\epsp^{2}}{16TG^{2}D^{2}}}.
\end{align*}

With bounds for $\encircle{1}$ and $\encircle{2}$ in hand, we have that
for any $\eps_{1},\eps_{2}>0$,
\begin{align*}
  &\Prob{\encircle{1}+\encircle{2}\ge \frac{A_{T}(z^{*})+B_{T}(z^{*})\eps_{1}+\eps_{2}}{T}}\\
  &\qquad\le\Exp{\frac{-\eps_{1}}{\sqrt{T}G}+2}+\Exp{\frac{-\eps_{2}^{2}}{16TG^{2}D^{2}}}\\
  &\qquad\le2\Exp{\frac{-\eps_{2}^{2}}{16TG^{2}D^{2}}+1}
\end{align*}
where in the last line we used $\Exp{\frac{-\eps_{2}^{2}}{16TG^{2}D^{2}}}\le\Exp{\frac{-\eps_{2}^{2}}{16TG^{2}D^{2}}+1}$ and then chose $\eps_{1}= \sqrt{T}G\brac{1+\frac{\eps_{2}^{2}}{16TG^{2}D^{2}}}$.
Equivalently, letting $\delta\defeq2\Exp{\frac{-\eps_{2}^{2}}{16TG^{2}D^{2}}+1}$,
we solve for $\eps_{2}$ to get $\eps_{2}= 4D\sqrt{1+\Log{\frac{2}{\delta}}}\sqrt{T}G$, and
conclude that with probability at least $1-\delta$,
\begin{align*}
  &\max_{y^{*}\in\cY}L(\thetabar_{T},y^{*})-\min_{\theta^{*}\in\Theta}L(\theta^{*},\ybar_{T})\\
  &\qquad\le \frac{A_{T}(z^{*})+B_{T}(z^{*})\sqrt{\sum_{t=1}^{T}\norm{\ghat_{t}}^{2}} + \sum_{t=1}^{T}\xi_{t}}{T}\\
  &\qquad\le \frac{A_{T}(z^{*})}{T} + \frac{B_{T}(z^{*})\brac{1+\Log{\frac{2}{\delta}}}G}{\sqrt{T}}\\
  &\qquad\qquad\qquad+\frac{4D\sqrt{1+\Log{\frac{2}{\delta}}}G }{\sqrt{T}}\\
\end{align*}
The proof is then completed by using $a+b\le 2\max\set{a,b}$ and substituting the lipschitz bound $G$ as before.

\section{Coin-betting and Black-box Reductions}\label{app:reductions}

In this section we give a brief overview of the key ideas and algorithms from
the online learning literature used in this paper.

Our primary interest in this paper is avoiding having to tune a step-size parameter.
Interestingly, it turns out that achieving this goal is related to
a particular type of adaptivity --- adaptivity to the norm of
any given comparator point $u$ --- manifesting in regret bounds of the
form $O\brac{\norm{u}\sqrt{\sum_{t=1}^{T}\norm{g_{t}}^{2}}}$.
\footnote{
  Here we're assuming for brevity the initial point is $x_{1}=\zeros$. Otherwise one can simply replace $\norm{u}$ with $\norm{x_{1}-u}$ throughout this section.
}
To see this, one can imagine
running online gradient descent (OGD) with a fixed step-size $\eta$, and ask what the optimal choice of fixed choice $\eta$ would be in hindsight --- if we could choose or approximate
this optimal $\eta$ in advance, no tuning would be necessary. Following
standard analysis \citep[Theorem 2.13]{orabona2019modern}, one can derive that the regret of OGD with a fixed step-size is bounded as
$R_{T}(u)\le\frac{\norm{u}^{2}}{2\eta}+\frac{\eta}{2}\sum_{t=1}^{T}\norm{g_{t}}^{2}$, from which the optimal bound
$R_{T}(u)\le\norm{u}\sqrt{\sum_{t=1}^{T}\norm{g_{t}}^{2}}$ can be derived by choosing $\eta^{*}=\frac{\norm{u}}{\sqrt{\sum_{t=1}^{T}\norm{g_{t}}^{2}}}$.
Of course, we could never actually select this $\eta^{*}$ in practice as it
requires knowing all the future subgradients $g_{t}$ as well as the norm of the unknown comparator point $u$. In
fact one can show that it's actually impossible to achieve this $O\brac{\norm{u}\sqrt{\sum_{t=1}^{T}\norm{g_{t}}^{2}}}$ bound without knowing $u$
in advance \citep[Chapter 5]{orabona2019modern}. However, several recent works can make guarantees of this form \textit{up to log factors},
particularly those operating in the coin-betting framework \cite{cutkosky2018black,orabona2016coin}.

In the coin-betting framework, the objective of guaranteeing low regret $R_{T}=\sum_{t=1}^{T}\inner{g_{t},x_{t}-u}$ is re-cast as a betting problem in which the objective is to guarantee high wealth, $W_{T}\defeq W_{0}-\sum_{t=1}^{T}\inner{g_{t},x_{t}}$, starting from some initial endowment of $W_{0}$.
Concretely, in the 1-dimensional OLO case, at each time $t\in[T]$ the learner places a bet $x_{t}=\beta_{t}W_{t-1}$ --- corresponding to some (signed) fraction $\beta_{t}\in[-1,1]$ of their current wealth $W_{t-1}$ ---
on the outcome of a continuous-valued ``coin'' $-g_{t}\in[-1,1]$. The learner either wins (or loses) the amount $-g_{t}x_{t}$, and their wealth becomes $W_{t}=W_{t-1}-g_{t}x_{t}=W_{t-1}(1-\beta_{t}g_{t})$.
The key idea is that guaranteeing a lower-bound on wealth corresponds to guaranteeing an upper-bound on regret: letting $F$ be a convex potential function and $F^{*}$ its Fenchel conjugate, it holds that \citep{mcmahan2014unconstrained}
\begin{align*}
  W_{T} \ge F\brac{\sum_{t=1}^{T}-g_{t}}-W_{0} \longeq R_{T}(u)\le F^{*}(u) + W_{0}.
\end{align*}
The key observation here is that designing a regret-minimizing algorithm
which is adaptive \wrt the comparator $u$ --- the key property of parameter-free algorithms ---
is equivalent to designing a betting strategy which is adaptive \wrt $\sum_{t=1}^{T}-g_t$.
Notably, it is much easier to design algorithms which are adaptive to this latter quantity
because it can be directly observed and measured by the learner.

In this paper, we employ a betting strategy based on the Online Newton Step Coin-betting algorithm of \citet[Algorithm 1]{cutkosky2018black}. The basic idea is as follows: at any time $t$ we can express
our accumulated wealth as a function of our chosen betting fractions, $W_{t}=W_{t-1}\brac{1-g_{t}\beta_{t}}=W_{0}\prod_{s=1}^{t-1}\brac{1-g_{s}\beta_{s}}$,
and we can consider how much more wealth \textit{could have} been attained if we had just used the best fixed betting fraction in hindsight:
\begin{align*}
  \frac{\max_{\beta^{*}\in[-1,1]}\prod_{s=1}^{t-1}\brac{1-g_{t}\beta^{*}}}{\prod_{s=1}\brac{1-g_{t}\beta_{t}}}.
\end{align*}
Equivalently, by taking the logarithm of this quantity we're interested in the difference
$\max_{\beta\in[-1,1]}\sum_{s=1}^{t}\Log{1-g_{t}\beta^{*}}-\sum_{s=1}^{t}\Log{1-g_{t}\beta_{t}}$, which we observe is the \textit{regret} of our choices $\beta_{t}$ on the sequence of loss functions $\ell_{t}(\beta)\defeq-\Log{1-g_{t}\beta}$. We can therefore
try to learn about the optimal betting fraction $\beta^{*}$ by applying an online learning algorithm to this loss sequence.
Luckily, these loss functions happen to have nice curvature properties which enable us to use the efficient Online Newton Step (ONS)  algorithm \citep{hazan2007logarithmic} --- this
means that we'll be able to converge to good betting fractions very quickly and obtain high wealth. In fact, this strategy will give us exactly the
regret we're looking for, $R_{T}(u) = \widetilde{O}\brac{\abs{u}\sqrt{\sum_{t=1}^{T}\abs{g_{t}}^{2}}}$, and the algorithm can be extended
to higher dimensions to achieve a bound of the same form
using either the black-box reduction of \citet[Algorithm 2]{cutkosky2018black}, or by applying the betting strategy separately in each dimension, resulting in a coordinate-wise,
AdaGrad-style algorithm.

However, for our purposes the approach described above is a bit unsatisfying as it
requires knowing a Lipschitz bound on the gradients \textit{a priori}.
Recall that in the coin-betting framework, the learner bets $x_{t+1}=\beta_{t+1}W_{t}$, and as a result the
wealth becomes $W_{t+1}= W_{t}(1-\beta_{t+1}g_{t+1})$. This strategy is only well-defined if the wealth is always non-negative, and thus for all $t$ we need $\abs{\beta_{t}}\le1/\abs{g_{t}}$.
Yet because the learner chooses $\beta_{t}$ prior to observing $g_{t}$, this condition can only be guaranteed if one knows a Lipschitz bound $G$ such that $\abs{g_{t}}\le G$.
In practice, this is somewhat unsatisfying because such a bound is either completely unknown \textit{a priori} or is difficult to compute tight bounds for.
In our problem setting, it just-so-happens that our assumptions enable us to bound subgradients $\norm{\hatg_{t}}$ almost surely, but these almost-sure bounds will typically be extremely pessimistic
  and lead to poor performance in practice if used to normalize the subgradients.

\begin{Algorithm}[!t]
  \STATE \textbf{Input:} Algorithm $\cA$ that takes hints, initial guess of gradient bound $\gbound$
  \STATE \textbf{Initialize:} $h_{1}=\gbound$
  \For{$t=1:T$}{
    \STATE Get $w_{t}$ from $\cA$, play $w_{t}$
    \STATE Receive subgradient $\widetilde{g}_{t}$
    \IfElse{$\abs{\widetilde{g}_t}>h_t$}{
      \STATE $\gtrunc_{t}=h_{t}\frac{\widetilde{g}_{t}}{\norm{\widetilde{g}_{t}}_{*}}$
    }{
      \STATE $\gtrunc_{t}=\widetilde{g}_{t}$
    }
    \STATE Set $h_{t+1}=\max\brac{h_{t},\norm{\widetilde{g}_{t}}_{*}}$
    \STATE Send $\gtrunc_{t}$ and $h_{t+1}$ to $\cA$
  }
  \caption{Gradient Clipping\cite{cutkosky2019artificial}}\label{alg:imperfect-hinting}
\end{Algorithm}

To address this situation, we employ the gradient clipping approach of \citet{cutkosky2019artificial}. To get some intuition for how this
works, suppose we had access to a sequence of ``hints'' $h_{t}$ which would inform us, ahead of time, that the next subgradient will satisfy $\norm{g_{t}}\le h_{t}$. Given access to such a sequence of hints, we could easily ensure that $\beta_{t}\le\frac{1}{\abs{g_{t}}}$ by clipping $\beta_{t}$ to be in the
range $[-1/\abs{h_{t}},1/\abs{h_{t}}]$. In practice we do not have access to such a sequence of hints, but we can approximate it using $h_{t+1} = \max_{\tau<t}\norm{g_{\tau}}$.
If our hint turns out to be incorrect, we can still make it \textit{appear} to be correct to the algorithm $\cA$ by passing it a truncated subgradient $\gtrunc_{t+1}=\frac{h_{t+1}}{\norm{g_{t+1}}}g_{t+1}$.
One can show that in the worst-case, such a procedure only incurs an additional regret penalty of the form $G\brac{\norm{u}+\max_{t}\norm{w_{t}}}$ \citep[Theorem 2]{cutkosky2019artificial}, shown in \Cref{thm:imperfect-hinting}.
Notably, in our problem setting this quantity is bounded almost-surely and can be considered a term in the function $A_{T}(z^{*})$ of \Cref{thm:sp-to-olo};
this penalty thus vanishes at a fast rate of $O(1/T)$, so we sacrifice very little in exchange for a great deal of freedom from prior-knowledge of problem-specific quantities.
The ONS coin-betting algorithm discussed above is then modified to make use of the hints $h_{t+1}$ by clipping the betting fractions to be within $[-\frac{1}{2h_{t+1}},\frac{1}{2h_{t+1}}]$.
The result is \Cref{alg:ons-with-hints}, and its regret guarantee is repeated here for completeness in \Cref{thm:ons-with-hints}

\begin{theorem}[\citet{cutkosky2019artificial}, Theorem 2]
  Suppose $\cA$ obtains $R_{T}(u,h_{T})$ given hints $h_{1}\le\ldots\le h_{T}$. Then Algorithm \ref{alg:imperfect-hinting} obtains
  \[
    R_{T}(u)\le R_{T}\Big(u, \max\set{\gbound,G}\Big) + G\max_{t\in[T]}\norm{w_{t}}+G\norm{u}
  \]
  where $G=\max_{t\in[T]}\norm{g_{t}}_{*}$.
  \label{thm:imperfect-hinting}
\end{theorem}

\begin{Algorithm}[!t]
  \STATE \textbf{Input:} Initial wealth $W_{0}>0$
  \STATE \textbf{Initialize:} betting fraction $\beta_{1}=0$, initial hint $h_{1}$
  \For{$t=1:T$}{
    \STATE Bet $z_{t}=\beta_{t}W_{t-1}$
    \STATE Receive $g_{t}\le h_{t}$
    \STATE Receive $h_{t+1}\ge h_{t}$
    \STATE Update wealth $W_{t}=W_{t-1}-g_{t}z_{t}$\STATE
    \skipline
    \tcc{Compute new betting fraction $\beta_{t}\in[-\half h_{t+1},\half h_{t+1}]$ using Online Newton Step on the loss function $\ell_{t}(\beta) = -\Log{1-\beta g_{t}}$}
    \STATE Set $v_{t}=\frac{d}{d\beta_{t}}\sbrac{-\Log{1-\beta g_{t}}}= \frac{g_{t}}{1-\beta_{t}g_{t}}$
    \STATE Set $A_{t}= 1 + \sum_{s=1}^{t}v_{s}^{2}$
    \STATE Set $\beta_{t+1}=\max\brac{\min\brac{\beta_{t}-\frac{2v_{t}}{(2-\Log{3})A_{t}}, \frac{1}{2h_{t+1}}}, \frac{-1}{2h_{t+1}}}$
  }
  \caption{Coin-betting ONS with Hints \cite{cutkosky2019artificial}}\label{alg:ons-with-hints}
\end{Algorithm}

\begin{theorem}[\citet{cutkosky2019artificial}, Theorem 1]
  The regret of Algorithm \ref{alg:ons-with-hints} is bounded by
  \begin{align*}
    &R_{T}(u,h_{T})\le\\
    &\begin{aligned} W_{0}+\abs{u}\max&\left\{8h_{T}\Log{16\abs{u}h_{T}e^{\frac{1}{4h_{T}^{2}}}\brac{1+g_{1:T}^{2}}^{4.5}},\right.\\
      &\left.2\sqrt{g_{1:T}^{2}\Log{\frac{4\brac{g_{1:T}^{2}}^{10}e^{\frac{1}{2h_{T}^{2}}}\abs{u}^{2}}{W_{0}^{2}}+1}}\right\}\end{aligned}
  \end{align*}
  where $g_{1:T}^{2}\defeq\sum_{t=1}^{T}\norm{g_{t}}^{2}$.
  \label{thm:ons-with-hints}
\end{theorem}

Next, notice that \Cref{alg:ons-with-hints} is only a procedure for a 1-dimensional OLO problem.
The most obvious way to extend this to higher dimensions is to simply run an instance of
it separately in each dimension, leading to a coordinate-wise algorithm in the flavor of AdaGrad.
This is precisely what the $\text{CW-PF}$ (\Cref{alg:cwpf}) subprocedure is doing.
However, as can be seen in \Cref{section:pfolo-to-policy-eval,app:tuning-free-olo,app:tuning-free-gtd},
this coordinate-wise decomposition can lead to additional dimension dependencies, and cause poor
performance when the subgradients happen to be high-dimensional and dense. Instead, one
can extend \Cref{alg:ons-with-hints} to higher dimensions using the dimension-free
reduction of \citet[Algorithm 2]{cutkosky2018black}, given here in \Cref{alg:1d-redux}.
The basic idea behind this reduction is to decompose the decisions $w_{t}\in\cW$  into
components $v_{t}\in\R_{+}$ and $u_{t}\in\set{x\in\R^{d}:\norm{x}\le 1}$ --- representing its \textit{scale} and \textit{direction} respectively --- and playing $w_{t} = v_{t}u_{t}$.
The 1-dimensional scale component
can then be handled by the 1-dimensional ONS with hints algorithm, and the direction component
can be handled with $O\Big(\sqrt{\sum_{t=1}^{T}\norm{g_{t}}}\Big)$ regret without thing any hyperparameters
using online gradient descent with step-sizes $\frac{\sqrt{2}}{2\sqrt{\sum_{\tau=1}^{t-1}\norm{g_{\tau}}^{2}}}$
\citep[Theorem 4.14]{orabona2019modern}. \Cref{thm:1d-redux} gives the regret associated with
\Cref{alg:1d-redux}, which paired with the ONS with hints guarantee (\Cref{thm:ons-with-hints}) and
the OGD guarantee gives the dimension-free regret bound of the form $\widetilde{O}\Big(\norm{\cmp}\sqrt{\sum_{t=1}^{T}\norm{g_{t}}^{2}}\Big)$.

\begin{Algorithm}[!t]
  \STATE \textbf{Require:} 1D coin-betting algorithm $\cA_{1D}$, Hilbert space $H$ and online learning algorithm $\cA_{S}$ defined on the unit-ball $S\subset H$
  \For{t=1:T}{
    \STATE Get point $x_{t}\in\R$ from $\cA_{1D}$
    \STATE Get point $y_{t}\in S$ from $\cA_{S}$
    \STATE Play $w_{t}=x_{t}y_{t}\in H$, receive gradient $g_{t}$
    \STATE Set $s_{t}=\inner{g_{t},y_{t}}$
    \STATE Send $s_{t}$ as the $t^{\text{th}}$ subgradient to $\cA_{1D}$
    \STATE Send $g_{t}$ as the $t^{\text{th}}$ subgradient to $\cA_{S}$
  }
  \caption{Dimension-free Reduction \cite{cutkosky2018black}}\label{alg:1d-redux}
\end{Algorithm}
\vspace{3em}
\begin{theorem}[\citet{cutkosky2018black}, Theorem 2]
  Suppose $\cA_{S}$ obtains regret $R_{t}^{\cA_{S}}(y^{*})$ for any $y^{*}$ in the unit ball and $\cA_{1D}$ obtains $R_{T}^{1D}(x^{*})$ for any $x^{*}\in\R$. Then
  Algorithm \ref{alg:1d-redux} guarantees regret
  \[
    R_{T}(z^{*})\le R_{T}^{1D}(\norm{z^{*}}) + \norm{z^{*}}R_{T}^{\cA_{S}}\brac{\frac{z^{*}}{\norm{z^{*}}}}.
  \]
  \label{thm:1d-redux}
\end{theorem}

Finally, a subtle detail that's been swept under the rug in the preceeding discussion is that coin-betting algorithms
are typically defined over \textit{unbounded} domains, which is not always appropriate in our
problem setting, and makes it difficult to reason about the variance of the stochastic subgradients. Luckily, this issue is easily overcome using the constraint-set reduction of \citet{cutkosky2020parameter}, shown in  \Cref{alg:constraint-set-new}.
Simply put, given a constraint set $\cW$
and algorithm $\cA_{\infty}$ with unbounded domain, we can generate decisions $x_{t}$ using $\cA_{\infty}$ and project them back into $\cW$ whenever necessary. Each time $\cA_{\infty}$ violates our constraints, we add a particular penalty function to
the loss that penalizes deviation from the constraint set $\cW$. These penalty functions are designed in such a way that
the regret suffered by $\cA_{\infty}$ on this penalized loss sequence upperbounds the regret in our original problem. In particular, \citet[Theorem 2]{cutkosky2020parameter} (repeated here in \Cref{thm:constraint-set-new}) shows that we can incorporate constraints this way without
any significant harm to our regret bounds.

\begin{Algorithm}[t!]
  \STATE \textbf{Input:} Domain $W\subset \R^{d}$, online learning algorithm $\cA$
  \For{t=1:T}{
    \STATE Get $w_{t}$ from $\cA$
    \STATE Play $\hat w_{t} \defeq \Pi_{W}(w_{t})$, receive $\hat g_{t}\in\partial\ell_{t}(\hat w_{t})$
    \STATE Let $\widetilde{w}_{t}\defeq\frac{w_{t}-\hat w_{t}}{\norm{w_{t}-\hat w_{t}}}$
    \STATE Define $S_{\cW}(w)\defeq\norm{w-\Pi_{\cW}(w)}$
    \STATE Define $\widetilde \ell_{t}(w)$ to be
    \STATE $\begin{aligned}
      &\quad\inner{\hat g_{t},w}\qquad&\text{if }\inner{\hat g_{t},w_{t}}\ge\inner{\hat g_{t},\hat w_{t}}\\
      &\quad\inner{\hat g_{t},w}-\inner{\hat g_{t},\widetilde{w}_{t}}S_{W}(w)&\text{otherwise}
    \end{aligned}$
    \STATE Compute $\widetilde g_{t}\in\partial\widetilde\ell_{t}(w_{t})$
    \STATE Send $\widetilde g_{t}$ to $\cA$ as the $t^{\text{th}}$ subgradient
  }
  \caption{Constraint Set Reduction \citep[Algorithm 1]{cutkosky2020parameter}}\label{alg:constraint-set-new}
\end{Algorithm}

\begin{theorem}[\citet{cutkosky2020parameter}, Theorem 2]
  The functions $\hat \ell_{t}$ defined in Algorithm \ref{alg:constraint-set-new} are convex functions defined on all
  of domain $W$ and the gradients sent to $\cA$ by Algorithm \ref{alg:constraint-set-new} satisfy $\norm{\widetilde g_{t}}\le\norm{\hat g_{t}}$.
  Also, for all $t$ and all $u\in W$, we have
  \[
    \inner{\hat g_{t},\hat w_{t}-u}\le\widetilde\ell_{t}(w_{t})-\widetilde\ell_{t}(u)\le\inner{\widetilde g_{t},w_{t}-u}
  \]
  \label{thm:constraint-set-new}
\end{theorem}

\section{Algorithms for Parameter-free OLO}\label{app:tuning-free-olo}

In this section we show how to derive the regret guarantees associated with the parameter-free OLO
subroutines used by PFGTD, CW-PFGTD, and PFGTD+, using a series of blackbox reductions. This
equips us with a toolbox of modular components which can be chained together to easily
derive the regret bounds for our policy evaluation algorithms in the next section.

\begin{lemma}[\citet{cutkosky2019artificial,cutkosky2020parameter}]
  \label{thm:constrained-scalar-clipping}
  Suppose $\cA$ obtains $R_{T}(\cmp,h_{T})$ given scalar hints $h_{1}\le \ldots\le h_{T}$. Then
  with scaling function $\cM^{\text{sc}}:g\mapsto\norm{g}_{*}$, the subgradients sent to $\cA$ satisfy
  $\norm{\widetilde{g}_{t}}_{*}\le\norm{\ghat_{t}}_{*}$ and \Cref{alg:constrained-clipping} obtains
  \begin{align*}
    R_{T}(\cmp)\le R_{T}(\cmp, \max\set{\gbound, G}) + (\norm{\cmp}+D)G,
  \end{align*}
  where $G=\max_{t\in[T]}\norm{\hat g_{t}}$ and  $D=\max_{w\in\cW}\norm{w}$.
\end{lemma}

\textit{Proof:} The claim then follows from Theorem 2 of \citet{cutkosky2019artificial} followed by Theorem 2 of \citet{cutkosky2020parameter} (provided in \Cref{app:reductions}, \Cref{thm:imperfect-hinting,thm:constraint-set-new}). To see this, consider applying \Cref{alg:constrained-clipping} to the linear loss sequence $\hat g_{1},\ldots,\hat g_{T}$;  we have for any $\cmp\in\cW$ that
\begin{align*}
  R_{T}(\cmp) &= \sum_{t=1}^{T}\inner{\hat g_{t}, \hat{w}_{t}-\cmp}\\
  &\overset{(a)}{\le} \sum_{t=1}^{T}\inner{\gtrunc_{t},\hat{w}_{t}-\cmp}+G(\norm{\cmp}+\max_{t}\norm{\hat{w}_{t}})\\
  &\overset{(b)}{\le}\sum_{t=1}^{T}\inner{\widetilde{g}_{t},w_{t} - \cmp} + G(\norm{\cmp}+D)\\
  &=R_{T}(\cmp, h_{T}) + G(\norm{\cmp}+D)\\
  &\overset{(c)}{=}R_{T}(\cmp, \max\set{\gbound,G}) + G(\norm{\cmp}+D)
\end{align*}
where $(a)$ applied \Cref{thm:imperfect-hinting}, $(b)$ applies \Cref{thm:constraint-set-new} to
the first term and applied $\hat w_{t}\in\cW\implies \norm{\hat w_{t}}\le D$ by the boundedness of $\cW$ to the second term, and
$(c)$ uses the fact that $h_{T}=\max\set{\gbound, \max_{t}\norm{g_{t}}_{*}}=\max\set{\gbound,G}$ by definition.
\Cref{thm:constraint-set-new} also tells us that $\norm{\widetilde{g}_{t}}_{*}\le\norm{\gtrunc_{t}}_{*}$,
and since $\norm{\gtrunc_{t}}_{*}\le\norm{\hat g_{t}}_{*}$ by definition, we have $\norm{\widetilde{g}_{t}}_{*}\le\norm{\hat g_{t}}_{*}$
as well.

\hfill$\blacksquare$

\begin{lemma}
  \label{thm:constrained-vector-clipping}
  Suppose $\cA$ obtains $R_{T}(\cmp,h_{T})$ given vector hints satisfying $h_{1i}\le \ldots\le h_{Ti}$ for all $i$. Then
  with scaling function $\cM^{\text{vec}}:g\mapsto(\abs{g_{1}},\ldots,\abs{g_{d}})^{\top}$, the subgradients sent to $\cA$ satisfy
  $\norm{\widetilde{g}_{t}}_{*}\le\norm{\ghat_{t}}_{*}$ and \Cref{alg:constrained-clipping} obtains
  \begin{align*}
    R_{T}(\cmp)\le R_{T}(\cmp, \Gbound) + d(\norm{\cmp}_{\infty}+D_{\infty})G_{\infty},
  \end{align*}
  where $\Gbound_{i}=\max\set{\gbound, G_{i}}$ for $G_{i}=\max_{t}\abs{\hat g_{ti}}$ and $G_{\infty}=\max_{i}G_{i}$.
\end{lemma}

\textit{Proof:} The result follows using similar arguments to \Cref{thm:constrained-scalar-clipping}, with
some minor modifications to handle the coordinate-wise clipping. Let $G_{ti}\defeq\max_{\tau\le t}\abs{\hat g_{ti}}$
\begin{align*}
  R_{T}(\cmp) &= \sum_{t=1}^{T}\inner{\hat g_{t}, \hat{w}_{t}-\cmp}\\
              &= \sum_{t=1}^{T}\inner{\gtrunc_{t},\hat{w_{t}}-\cmp} + \sum_{t=1}^{T}\inner{\hat{g}_{t}-\gtrunc_{t}, \hat{w}_{t} -\cmp}\\
              &\overset{(d)}{\le} \sum_{t=1}^{T}\inner{\widetilde{g}_{t},w_{t}-\cmp}
  + \sum_{t,i}\brac{\hat g_{ti}-\gtrunc_{ti}}\brac{\hat{w}_{ti}-\cmp_{i}}\\
              &\overset{(e)}{\le} R_{T}(\cmp, h_{T})
                +\brac{D_{\infty}+\norm{\cmp}_{\infty}} \sum_{t,i}\abs{\hat g_{ti}-\gtrunc_{ti}}
\end{align*}
where $(d)$ applies \Cref{thm:constraint-set-new} of the constraint-set reduction and $(e)$ applies the regret guarantee of $\cA$ given vector hints $h_{1},\ldots,h_{T}$, where $h_{T}$ is the vector of hints with
$h_{Ti}=\max\set{\gbound, G_{i}}$.
The double sum bounded as follows:
\begin{align*}
  \sum_{t=1}^{T}\sum_{i=1}^{d}\abs{\hat g_{ti}-\gtrunc_{ti}}
  &\le\sum_{i=1}^{d}\sum_{t:G_{ti}>h_{ti}}G_{ti}-h_{ti}\\
  &\overset{(f)}{\le}\sum_{i=1}^{d}\sum_{t:G_{ti}>h_{ti}}G_{ti}-G_{t-1,i}\\
  &=\sum_{i=1}^{d}G_{Ti} \le dG_{\infty}
\end{align*}
where $(f)$ used that $h_{ti} = \max\set{\gbound, G_{ti}}\ge G_{t-1,i}$.
Thus, we have that
\begin{align*}
  R_{T}(\cmp)
  &\le R_{T}(\cmp, \Gbound)
    +d\brac{D_{\infty}+\norm{\cmp}_{\infty}}G_{\infty}.
\end{align*}
Further, $\norm{\widetilde{g}_{t}}_{*}\le\norm{\hat g_{t}}_{*}$ follows from
the facts that $\norm{\widetilde{g}_{t}}_{*}\le\norm{\gtrunc_{t}}$ via \Cref{thm:constraint-set-new}
and
that for every $i$, either $\abs{\gtrunc_{ti}}\le \abs{\hat g_{ti}}$,
so the norm of $\gtrunc_{t}$ can be no larger than that of $\hat g_{t}$.

\hfill$\blacksquare$

Using the preceding Lemma, the subroutines used by our policy evaluation algorithms are
constructed by composing \Cref{alg:constrained-clipping} with a parameter-free OLO algorithm (see \Cref{app:tuning-free-gtd}).
The following lemmas show how to characterize the regret of the PF, CW-PF, and PF+ OLO subroutines, and
pseudocode for these subroutines is given by \Cref{alg:pf,alg:cwpf,alg:pfcombined}.

\begin{minipage}{\columnwidth}
  \begin{lemma}\label{lemma:pf}
    Given hints $h_{1}\le\ldots\le h_{T}\eqdef\Gbound$,
    regret of \Cref{alg:pf} is bounded as
    \begin{align*}
      R_{T}^{\cA_{\cW}}(\cmp, \Gbound)
      &\le
        W_{0}+\sqrt{2}\norm{\cmp}\sqrt{\sum_{t=1}^{T}\norm{\widetilde{g}_{t}}_{*}^{2}}\\
      &+\norm{\cmp}\max\set{S_{1}(\Gbound),  2\sqrt{S_{2}(\Gbound)\sum_{t=1}^{T}\norm{\widetilde{g}_{t}}_{*}^{2}}}\\
      &\le
        \widetilde{O}\Big(\norm{\cmp}\sqrt{\sum_{t=1}^{T}\norm{\widetilde{g}_{t}}_{*}^{2}}\Big),
    \end{align*}
    where both $S_{1}(\Gbound)$ and $S_{2}(\Gbound)$
    are bounded as $O\brac{\Log{\frac{\norm{\cmp}\Gbound T}{W_{0}}+1}}$.
  \end{lemma}
\end{minipage}

\begin{Algorithm}[t!]
  \small
  \STATE \textbf{Input:} Initial wealth $W_{0}> 0$
  \hfill(\textit{Default:} $W_{0}=1$)
  \STATE \textbf{Initialize:} betting fraction $\beta_{1}=0$, initial point $u_{1}\in\R^{d}$
  \STATE \textbf{Receive} Initial hint $h_{1}$
  \For{$t=1:T$}{
    \STATE Set $v_{t}=\beta_{t}W_{t-1}$

    \STATE\textbf{Play} $w_{t}=v_{t}u_{t}$
    \STATE\textbf{Receive} $\widetilde{g}_{t}$ and $h_{t+1}$ satisfying $\norm{\widetilde{g}_{t}}_{*}\le h_{t}\le h_{\tpp}$\STATE

    \STATE Set $s_{t} = \inner{\widetilde{g}_{t},u_{t}}$
    \STATE \textbf{Update} $W_{t}=W_{t-1}-s_{t}v_{t}$\STATE

    \STATE // ONS with Hints Update (Algorithm \ref{alg:ons-with-hints})
    \STATE Set $m_{t}= \frac{d}{d\beta}\Log{\frac{1}{1-\beta s_{t}}}=\frac{s_{t}}{1-\beta_{t}s_{t}}$
    \STATE \quad\ \  $\hat\beta_{{t+1}} =\beta_{t}-\frac{2}{(2-\Log{3})}\frac{m_{t}}{1+\sum_{\tau=1}^{t}m_{\tau}^{2}}$
    \STATE \textbf{Update} $\beta_{t+1}=\max\brac{\min\brac{\hat \beta_{t+1},\frac{1}{2h_{t+1}}}, \frac{-1}{2h_{t+1}}}$\STATE

    \tcc{Adaptive Gradient Descent on $S=\set{x:\norm{x}\le 1}$}
    \STATE \textbf{Update} $u_{t+1}=\Pi_{S}\brac{u_{t}-\frac{\sqrt{2}}{2\sqrt{\sum_{\tau=1}^t\norm{\widetilde{g}_{\tau}}^2}}\widetilde{g}_{t}}$ \STATE
  }
  \caption{\pf{PF}}
  \label{alg:pf}
\end{Algorithm}

\textit{Proof:} \Cref{alg:pf} plays points $w_{t}=v_{t}u_{t}$ with $u_{t}\in\R^{d}$ chosen by an algorithm $\cA_{S}$ with domain
$S=\set{x\in\R^{d}:\norm{x}\le 1}$ and $v_{t}$ chosen by an unconstrained OLO algorithm $\cA_{1D}$ with domain $\R$, so
using \Cref{thm:1d-redux} of the dimension-free reduction (Algorithm \ref{alg:1d-redux}) we have
\begin{align*}
  &\sum_{t=1}^{T}\inner{\widetilde{g}_{t},w_{t}-\cmp}
  \le R_{T}^{\cA_{1D}}(\norm{\cmp}) + \norm{\cmp}R_{T}^{\cA_{S}}\brac{\frac{\cmp}{\norm{\cmp}}}.
\end{align*}
If we let $\cA_{S}$ be online gradient descent with step-sizes $\eta_{t} = \dfrac{\sqrt{2}}{2\sqrt{\sum_{\tau=1}^{t}\norm{\widetilde{g}_{\tau}}_{*}^{2}}}$,
we can guarantee that \citep[Theorem 4.14]{orabona2019modern}
\[
R_{T}^{\cA_{S}}\brac{\frac{\cmp}{\norm{\cmp}}}\le \sqrt{2}\sqrt{\sum_{t=1}^{T}\norm{\widetilde{g}_{t}}^{2}_{*}},
\]
and by letting $\cA_{1D}$ be the ONS coin-betting with hints algorithm (Algorithm \ref{alg:ons-with-hints}, Appendix \ref{app:reductions}), we
use Theorem \ref{thm:ons-with-hints} and the fact that for any $u\in S$, $\inner{\widetilde{g},u}\le\max_{u:\norm{u}\le 1}\inner{\widetilde{g},u} = \norm{\widetilde{g}_{t}}_{*}$ to
get
\begin{align*}
  R_{T}^{1D}&(\norm{\cmp})\le W_{0}+\norm{\cmp}\max\set{S_{1},  2\sqrt{S_{2}\sum_{t=1}^{T}\norm{\widetilde{g}_{t}}_{*}^{2}}},
\end{align*}
where
\begin{align}
  S_{1}&\le 8h_{T}\Log{\frac{16\norm{\cmp}\Gbound e^{\frac{1}{4\Gbound^{2}}}\brac{1+TG}^{4.5}}{W_{0}}}\label{eq:S_1}\\
  S_{2}&\le\Log{\frac{4\norm{\cmp}^{2}e^{\frac{1}{2\Gbound^{2}}}\brac{TG}^{10} }{W_{0}^{2}}+1}\label{eq:S_2}.
\end{align}
Taken together, we have that
\begin{align*}
  R_{T}(\cmp)
  &\le
    \sum_{t=1}^{T}\inner{\widetilde{g}_{t},w_{t}-\cmp}\\
  &\le
    R_{T}^{\cA_{1D}}(\norm{\cmp}) + \norm{\cmp}R_{T}^{\cA_{S}}\brac{\frac{\cmp}{\norm{\cmp}}}\\
  &
    \begin{aligned}\le &W_{0}+\sqrt{2}\norm{\cmp}\sqrt{\sum_{t=1}^{T}\norm{\widetilde{g}_{t}}_{*}^{2}}\\
      &+\norm{\cmp}\max\set{S_{1},  2\sqrt{S_{2}\sum_{t=1}^{T}\norm{\widetilde{g}_{t}}_{*}^{2}}}\end{aligned}\\
  &\le \widetilde{O}\Big(\norm{\cmp}\sqrt{\sum_{t=1}^{T}\norm{\widetilde{g}_{t}}_{*}^{2}}\Big).
\end{align*}
Finally, that the terms $S_{1}$ and $S_{2}$ can be bounded as $O\brac{\Log{\frac{\norm{\cmp}\Gbound T}{W_{0}}+1}}$
follows from algebraic manipulations and properties of logarithms\footnote{The computation is rather tedius, but the basic idea is to simply use an upperbound which raises the quantities inside the logarithm to the same power, giving
  $\Log{x^{a}}=a\Log{x}\le O(\Log{x})$}.\hfill$\blacksquare$

The coordinate-wise component, CW-PF, simply plays the ONS with hints algorithm coordinate-wise, leading immediately to the
following lemma:

\begin{minipage}{\columnwidth}
  \begin{lemma}\label{lemma:cwpf}
    Given hint vectors such that $h_{1i}\le\ldots\le h_{Ti}\eqdef\Gbound_{i}$ for all $i\in[d]$,
    The regret of \Cref{alg:cwpf} is bounded as
    \begin{align*}
      R_{T}(\cmp,\Gbound)&\le dW_{0}
      +\sum_{i=1}^{d}\abs{\cmp_{i}}\max\set{S_{1i},  2\sqrt{S_{2i}\sum_{t=1}^{T}\abs{\widetilde g_{ti}}^{2}}}\\
                 &\le\widetilde{O}\Big(\sum_{i=1}^{d}\abs{\cmp_{i}}\sqrt{\sum_{t=1}^{T}\abs{\widetilde g_{ti}}^{2}}  \Big)
    \end{align*}
    where both $S_{1i}(\Gbound_{i})$ and $S_{2i}(\Gbound_{i})$
    are bounded as $O\brac{\Log{\frac{\abs{\cmp_{i}}\Gbound_{i}T}{W_{0}}+1}}$.
  \end{lemma}
\end{minipage}

\textit{Proof:} \Cref{alg:cwpf} runs a separate instance of the ONS with hints algorithm in each dimension, giving the
regret decomposition
\begin{align*}
  \sum_{t=1}^{T}\sum_{i=1}^{d}\widetilde{g}_{ti}(w_{ti}-\cmp_{i}) = \sum_{i=1}^{d}\underbrace{\sum_{t=1}^{T}\widetilde{g}_{ti}(w_{ti}-\cmp_{i})}_{R_{T}^{(i)}(\cmp_{i})}.
\end{align*}
The claim then follows by applying Theorem 1 of \citet{cutkosky2019artificial} in each dimension (see \Cref{thm:ons-with-hints}, \Cref{app:reductions}). \hfill$\blacksquare$

\begin{Algorithm}[t!]
  \small
  \STATE \textbf{Input:} Initial wealth $W_{0}\in\R^{d}_{+}$\hfill(default: $W_{0}=\ones$)
  \STATE \textbf{Initialize:} Betting fractions $\beta_{1}=\zeros$
  \STATE \textbf{Receive} Initial hint $h_{1}$
  \For{$t=1:T$}{
    \STATE \textbf{Play} $w_{t}=\beta_{t}W_{\tmm}$
    \STATE \textbf{Receive} $\widetilde{g}_{t}$ and $h_{\tpp}$ such that $\forall i,\ \abs{\widetilde{g}_{ti}}\le h_{t,i}\le h_{\tpp,i}$\STATE

    \For{$i=1:d$}{
      \STATE \textbf{Update} $W_{ti}=W_{t-1,i}-\widetilde{g}_{ti}\beta_{ti}$\STATE

      \tcc{ONS with Hints Update (Algorithm \ref{alg:ons-with-hints})}
      \STATE Set $m_{ti}= \frac{d}{d\beta}\Log{\frac{1}{1-\beta \widetilde{g}_{ti}}}=\dfrac{\widetilde{g}_{ti}}{1-\beta_{ti}\widetilde{g}_{ti}}$
      \STATE \quad\ \  $\hat\beta_{{t+1}} =\beta_{ti}-\frac{2}{(2-\Log{3})}\dfrac{m_{ti}}{1+\sum_{\tau=1}^{t}m_{\tau i}^{2}}$
      \STATE \textbf{Update} $\beta_{t+1,i}=\max\brac{\min\brac{\hat \beta_{t+1},\frac{1}{2h_{t+1,i}}}, \frac{-1}{2h_{t+1,i}}}$
    }
  }
  \caption{\cwpf{CW-PF}}
  \label{alg:cwpf}
\end{Algorithm}

\subsection{Combining Guarantees}\label{app:combining}
As noted by \citet{cutkosky2019combining}, the bounds of the form
\begin{align*}
  R_{T}(\cmp)\le\widetilde{O}\Big(\norm{\cmp}\sqrt{\sum_{t=1}^{T}\norm{g_{t}}^{2}}\Big)
\end{align*}
and those of the form
\begin{align*}
  R_{T}(\cmp)\le\widetilde{O}\Big(\sum_{i=1}^{d}\abs{\cmp_{i}}\sqrt{\sum_{t=1}^{T}\abs{g_{ti}}^{2}}\Big)
\end{align*}
are generally not comparable --- the better bound largely depends on the particular sequence of gradients received. This makes the choice between the preceeding algorithms
less clear in practice, as we may not know the properties of the gradients will be in advance. Remarkably, \citet{cutkosky2019combining} shows that adding the iterates of
two parameter-free algorithms enables us to guarantee the better regret bound of the two up to a constant factor.
To see this, let $w_{t}^{\cA}$ and $w_{t}^{\cB}$ be the iterates of parameter-free algorithms $\cA$ and $\cB$, set $w_{t}=w_{t}^{\cA}+w_{t}^{\cB}$, and decompose the regret as
\begin{align*}
  R_{T}(\cmp)&\le\sum_{t=1}^{T}\inner{g_{t},w_{t}-\cmp} = \sum_{t=1}^{T}\inner{g_{t},w_{t}^{\cA}+w_{t}^{\cB}-\cmp}\\
  &\overset{x+y=\cmp}{=} R_{T}^{\cA}(x) + R_{T}^{\cB}(y).
\end{align*}

\begin{Algorithm}[t!]
  \small
  \STATE \textbf{Input:} Instances of $\text{PF}$ and $\text{CW-PF}$ algorithms with initial wealth $W_{0}^{\text{PF}}=\frac{d}{2}W_{0}$ and $W_{0}^{\text{CW-PF}}=\half W_{0}$ for $W_{0}>0$
  \STATE \textbf{Receive} initial hint $h_{1}$
  \STATE \textbf{Send} $\norm{h_{1}}/\sqrt{d}$ to $\text{PF}$ as the initial hint
  \STATE \textbf{Send} $h_{1}$ to $\text{CW-PF}$ as the initial hint
  \For{$t=1:T$}{
    \STATE Get $w_{t}^{\text{PF}}$ from $\text{PF}$ and  $w_{t+1}^{\text{CW-PF}}$ from $\text{CW-PF}$\STATE

    \STATE \textbf{Play} $w_{t} = w_{t}^{\text{PF}}+w_{t}^{\text{CW-PF}}$
    \STATE \textbf{Receive} $\widetilde{g}_{t}$ and $h_{t+1}$\STATE

    \STATE Send $\widetilde{g}_{t}$ and $\norm{h_{t+1}}$ to $\text{PF}$
    \STATE Send $\widetilde{g}_{t}$ and $h_{t+1}$ to $\text{CW-PF}$
  }
  \caption{\pfc{PF+}}
  \label{alg:pfcombined}
\end{Algorithm}
This holds for \textit{any} such $(x,y)$ such that $x+y=\cmp$, and
in particular, it holds for both $(\cmp,\zeros)$ and $(\zeros,\cmp)$ \textit{simultaneously}, so the regret will be bounded by the lower of the two:
\begin{align*}
  R_{T}(\cmp)&\le \min\set{R_{T}^{\cA}(\cmp)+R_{T}^{\cB}(\zeros), R_{T}^{\cA}(\zeros)+R_{T}^{\cB}(\cmp)}.
\end{align*}
If there is a $\eps$ such that $R_{T}^{\cA}(\zeros)\le\eps$ and $R_{T}^{\cB}(\zeros)\le\eps$, we get
\begin{align*}
  R_{T}(\cmp)\le\eps+\min\set{R_{T}^{\cA}(\cmp),R_{T}^{\cB}(\cmp)}.
\end{align*}
This is quite convenient when $\cA$ and $\cB$ are parameter-free algorithms; such
an $\eps$ can be easily found due to the fact that all horizon-dependent terms in the upperbound also
have a multiplicative dependence on $\norm{\cmp}$,
and thus disappear when $\cmp=\zeros$.
Typically this results in an additional constant $\eps=W_{0}$ in the regret bound.
Supposing then that PF+ initializes initializes PF with $W_{0}^{\text{PF}}=dW_{0}$ and CW-PF with $W_{0}^{\text{CW-PF}}=W_{0}$ for
some $W_{0}>0$, it's easy to see that both algorithms would satisfy $R_{T}(\zeros)\le dW_{0}$. The following lemma is
then immediate:

\begin{minipage}{\columnwidth}
  \begin{lemma}\label{lemma:pfcombined}
    Assume hints $h_{1},\ldots,h_{T}$ satisfy $h_{1i}\le\ldots\le h_{Ti}\eqdef\Gbound_{i}$ for all $i\in[d]$.
    Then regret of \Cref{alg:pfcombined} is bounded as
    \begin{align*}
      R_{T}(\cmp)\le dW_{0}+\min\set{R_{T}^{\text{PF}}(\cmp,\norm{\Gbound}),R_{T}^{\text{CW-PF}}(\cmp,\Gbound)}
    \end{align*}
  \end{lemma}
\end{minipage}

There are a few subtleties to be mentioned. Notice that $\text{PF}$ is given hints $\norm{h_{t}}$,
where each $h_{ti}=\max_{\tau<t}\abs{\hat g_{\tau i}}$, rather than the regular scalar hints $h_{t}^{\text{sc}}\defeq\max_{\tau<t}\norm{\hat g_{\tau}}$.
These $\norm{h_{t}}$ hints will still work since
\begin{align*}
  h_{t}^{\text{sc}}=\max_{\tau<t}\norm{\hat g_{t}}\le\sqrt{\sum_{i=1}^{d}\max_{\tau<t}\hat g_{ti}^{2}}=\norm{h_{t}},
\end{align*}
and so the $\gtrunc_{t}$ provided to $\cA$ respect the hints $\norm{h_{t}}$: $\norm{\gtrunc_{t}}\le h_{t}^{\text{sc}}\le\norm{h_{t}}$.
However this also implies a worse dimension dependence when using the hints $\norm{h_{t}}$ --- recall from \Cref{lemma:pf} that
the maximal hint $\Gbound$ ends up in the log factors $O\brac{\Log{\frac{\norm{\cmp}\Gbound T}{W_{0}}+1}}$, and when using hints $\norm{h_{t}}$
this
$\Gbound$ could be as large as $\sqrt{d}G$, thus adding a horizon-dependent dimension dependence to the bound we'd get
using the regular scalar hints. We avoid incurring this horizon-dependent dimension penalty by initializing the
PF component with $dW_{0}$, so that $\Log{\frac{\norm{\cmp}\Gbound T}{dW_{0}}+1}\le\Log{\frac{\norm{\cmp}\Gbound^{\text{sc}} T}{W_{0}}+1}$. It's easy
to see that as a result, we'll have $R_{T}^{\text{PF}}(\cmp, \norm{h_{T}})\le (d-1)W_{0} + R_{T}^{\text{PF}}(\cmp, h_{T}^{\text{sc}})$,
so the resulting algorithm is never quite dimension-free.
To get the best of both worlds from the $\text{PF}$ and $\text{CW-PF}$ algorithms, we'd ideally be able to
use the regular $h_{t}^{\text{sc}}=\max_{\tau<t}\norm{\hat{g}_{t}}$ hints for $\text{PF}$ and the vector hints with $h_{ti}=\max_{\tau<t}\abs{\hat{g}_{ti}}$ for $\text{CW-PF}$, but this would involve
clipping the stochastic subgradients $\hat{g}_{t}$ in a different way for each of the algorithms. This is not a problem in an
unconstrained setting --- given a $\hat g_{t}$, we could simply send it off to both algorithms and
add the returned iterates together, regardless of what additional individual processing the algorithms did to
$\hat g_{t}$.

Things are a bit more difficult in the constrained setting, however.
First, observe that the regret penalty incurred for the
applying the gradient clipping algorithm is of the form $G\brac{D+\norm{\cmp}}$. This
holds because the $w_{t}$ returned to the clipping procedure are first constrained to $\cW$ via the
constraint-set reduction. More generally, this penalty is of the form $G\brac{\max_{t}\norm{w_{t}}+\norm{\cmp}}$, which could be arbitrarily large when $\cW$ is
unbounded so
the clipping should be applied \textit{before} the constraint-set reduction.
Yet we also need to perform the constraint-set reduction
\textit{after} having added $w_{t} = w_{t}^{\cA}+w_{t}^{\cB}$, since generally $w_{t}$
may not be in $\cW$ even if both $w_{t}^{\cA}$ and $w_{t}^{\cB}$ are. Thus, the constraint-set
reduction acts as a sort of bottleneck, preventing $\cA$ and $\cB$ from simultaneously using
the clipping algorithm that best suits their individual strengths.

The ideal scenario would be to achieve
the dimension-free regret bound of $\text{PF}$ in the worst-case, while still being able to
reap the benefits of the coordinate-wise algorithm automatically when the gradients happen
to be sparse. Such a result was recently achieved using a generic combiner
algorithm, but has the additional expense of solving a linear optimization problem
at each step in order to track the regret both the dimension-free and
AdaGrad-style algorithms \citep{bhaskara2020online}.
Instead, our algorithm is ``almost dimension-free'', in the sense that
the dimension dependence shows up only in constant terms rather than with any
horizon-dependent terms, but avoids this additional expensive computation. Our experimental results (\Cref{section:experiments})
suggest that in practice the dimension-dependent constant term has a negligable effect; the PF+ algorithm performs nearly
the same as either PF or CW-PF --- whichever happens to be better in a given
problem.

\subsection{Scale-invariant Updates}\label{app:scale-free}

As can be inferred from the feedback diagrams of \Cref{fig:olo-algs} in
\Cref{section:pfolo-to-policy-eval}, the ONS with Hints components
of the PF, CW-PF, and PF+ procedures could just-as-easily be replaced with any
other Lipschitz-adaptive parameter-free OLO algorithm.
Of particular note are the
Lipschitz-adaptive and scale-invariant algorithms of
\citet{mhammedi2020lipschitz}: FreeGrad and FreeRange. FreeGrad
is a parameter-free algorithm with similar guarantees to those derived
using ONS with Hints, but has the desirable feature of yielding
scale-invariant updates --- that is, rescaling all gradients
by some constant $c\in\R$ has no effect on the iterates
chosen by the algorithm.

Similar to the ONS-based approach,
FreeGrad's regret bound depends on a term $G/h_{1}$ --- the ratio
between the Lipschitz bound $G$ and our initial guess $h_{1}$, which
could be arbitrarily large depending on how poorly we chose our $h_{1}$.
This issue is referred to as the \textit{range-ratio} problem.
The FreeRange algorithm adds a simple wrapper around the FreeGrad
algorithm which avoids the range-ratio algorithm using
an extension of a restart scheme proposed in \citet{mhammedi2019lipschitz}, at the
expense of an additional penalty of at most
$h_{T}\brac{16\norm{\cmp}\log_{+}\brac{2\norm{\cmp}(T+1)^{3}}+2\norm{\cmp}+3}$,
where $\log_{+}(\cdot)=\max\set{\Log{\cdot}, 0}$.

In the case where $\norm{\cdot}=\norm{\cdot}_{2}$, FreeRange also has the
benefit of having a $d$-dimensional implementation \textit{without}
relying on the dimension-free reduction of \citet{cutkosky2018black} (see \Cref{alg:1d-redux}).
The resulting algorithm is considerably simpler
implementation-wise, and is included in \Cref{alg:free-range} for completeness.
However, this implementation is only valid under
$\norm{\cdot}=\norm{\cdot}_{2}$; for general dual-norm pairs, FreeRange/FreeGrad
also require the dimension-free reduction, just like
ONS with Hints --- the difference is that now \textit{both}
algorithms in the dimension-free reduction must be scale-invariant
if the resulting algorithm is to maintain the scale-invariant property.
This is could be remedied by instead applying a scale-free algorithm such as
SOLO-FTRL or SOLO-MD \citep{orabona2018scale} instead of adaptive gradient
descent.

Variants of PF, CW-PF, and PF+ which use FreeRange as the
parameter-free subroutine were also tested in the experiments of
\Cref{section:experiments}. Interestingly, the results were nearly identical
to the results when using ONS with Hints, even in the large-scale
prediction experiment where one would expect scale-invariance to have a significant
impact on the outcome. We thus focus on the ONS with Hints approach in the main
text for ease of exposition, and leave further investigation of the
scale-invariant versions of these algorithms to future work.

\begin{Algorithm}
  \caption{Free-Range \citep{mhammedi2020lipschitz}}
  \label{alg:free-range}
  \STATE \textbf{Require:} $\norm{\cdot}=\norm{\cdot}_{2}$
  \STATE Receive initial hint $h_{1}$
  \STATE Initialize $\mathbf{G}=\zeros\in\R^{d}$, $V = h_{1}^{2}$, $R=2$
  \For{$t=1:T$}{
    \If{$h_{t}/h_{1}>R$}{
      \STATE // Reset the algorithm\\
      \STATE $h_{1}\gets h_{t}$\\
      \STATE $V\gets h_{1}^{2}$\\
      \STATE $\mathbf{G}\gets \zeros$\\
      \STATE $R\gets 2$
    }
    \STATE
    \STATE \textbf{Play} $w_{t} = -\mathbf{G}\frac{(2V+h_{t}\norm{\mathbf{G}})h_{1}^{2}}{2(V+h_{t}\norm{\mathbf{G}})^{2}\sqrt{V}}\Exp{\frac{\norm{\mathbf{G}}^{2}}{2V+2h_{t}\norm{\mathbf{G}}}}$\\
    \STATE \textbf{Receive} $g_{t}$ and hint $h_{\tpp}$
    \STATE
    \STATE \textbf{Update} $\mathbf{G}\gets \mathbf{G}+g_{t}$,\quad
    $V\gets V+\norm{g_{t}}^{2}$,\quad
    \STATE \qquad\quad\  $R\gets R+\frac{\norm{g_{t}}}{h_{t}}$
  }
\end{Algorithm}

\section{Algorithms for Parameter-free Policy Evaluation}\label{app:tuning-free-gtd}

With the saddle-point to OLO reduction of \Cref{app:sp-to-olo} and the parameter-free subroutines of
 \Cref{app:tuning-free-olo}, we can construct parameter-free gradient temporal difference algorithms
by appropriately chaining together a sequence of reductions. In this section we assume $\norm{\cdot}$ is the
euclidean norm for simplicity, though similar results can be derived for arbitrary Hilbert spaces.

We begin with the dimension-free PFGTD algorithm.
As motivated in \Cref{section:pfolo-to-policy-eval},
PFGTD is constructed using the saddle-point reduction of \Cref{alg:sp-to-olo} along with
two algorithms $\cA_{\Theta}$ and $\cA_{\cY}$, which are applied to the stochastic subgradients
$(\hat g_{t}^{\theta})_{t=1}^{T}$ and $(\hat g_{t}^{y})_{t=1}^{T}$ respectively.
The algorithms $\cA_{\Theta}$ and $\cA_{\cY}$ are constructed by composing
constrained scalar clipping (\Cref{alg:constrained-clipping} with $M^{\text{sc}}:g\mapsto\norm{g}_{*}$) with the PF subroutine (\Cref{alg:pf}).

\PFGTDRegret*

\textit{Proof:}
Let $\hat g_{t}^{\theta}$ and $\hat g_{t}^{y}$ denote the stochastic subgradients defined by Equations \ref{eq:noisy-subgrads} and let $(\theta^{*},y^{*})\in\Theta\times\cY$ be a saddle point of Equation \ref{eq:sp}. PFGTD plays OLO algorithms $\cA_{\Theta}$ and $\cA_{\cY}$ against the sequences
$(\hat{g}_{t}^{\theta})_{t=1}^{T}$ and $(\hat{g}_{t}^{y})_{t=1}^{T}$ respectively, yielding the
regret decomposition
\begin{align}
R_{T}(z^{*})\le R_{T}^{\cA_{\Theta}}(\theta^{*})+R_{T}^{\cA_{\cY}}(y^{*}).\label{eq:pf-tmp}
\end{align}
PFGTD composes \Cref{alg:constrained-clipping,alg:pf}, so applying \Cref{thm:constrained-vector-clipping} followed by \Cref{lemma:pf} and regrouping terms with $\norm{\theta^{*}}$, we can write
\begin{align*}
  R_{T}^{\cA_{\Theta}}(\theta^{*})
  &\le
  A_{T}^{\theta}(\theta^{*},\Gbound_{\theta})+B_{T}^{\theta}(\theta^{*},\Gbound_{\theta})\sqrt{\sum_{t=1}^{T}\norm{\hat g_{t}^{\theta}}^{2}},\\
\end{align*}
where $\Gbound_{\theta}\defeq\max\set{\gbound, \hat{G}_{\theta}}$ for $\hat{G}_{\theta} = \max_{t}\norm{\hat g_{t}^{\theta}}$, and
non-negative deterministic functions $A_{T}$ and $B_{T}$ such that
\begin{align*}
  &A_{T}^{\theta}(\theta^{*},\Gbound_{\theta})\le\\ &\quad O\Bigg(
                                                      W_{0}+\hat G_{\theta}D
                                                      +
                                               \norm{\theta^{*}}\sbrac{\hat G_{\theta}+\Log{\frac{2\norm{\theta^{*}}\Gbound_{\theta} T}{W_{0}}+1}}
                                               \Bigg)
  \\
  &B_{T}^{\theta}(\theta^{*},\Gbound_{\theta})\le O\brac{ \norm{\theta^{*}}\sqrt{\Log{\frac{2\norm{\theta^{*}}\Gbound_{\theta} T}{W_{0}}+1}}}.
\end{align*}
and
Likewise, we have
\begin{align*}
  R_{T}^{\cA_{\cY}}(y^{*})&\le
   A_{T}^{y}(y^{*},\Gbound_{y})+B_{T}^{y}(y^{*},\Gbound_{y})\sqrt{\sum_{t=1}^{T}\norm{\hat g_{t}^{y}}^{2}}\\
\end{align*}
with $A_{T}^{y}(y^{*})$ and $B_{T}^{y}(y^{*})$ bounded analogously.
Letting $z^{*} = (\theta^{*},y^{*})$, $\hat g_{t} = (\hat g_{t}^{\theta},\hat g_{t}^{y})$,
and defining
\begin{align}
  A_{T}(z^{*})&\defeq A_{T}^{y}(y^{*},\Gbound_{y})+A_{T}^{\theta}(\theta^{*},\Gbound_{\theta})\label{eq:pf-At}\\
  B_{T}(z^{*})&\defeq \sqrt{B_{T}^{\theta}{(\theta^{*},\Gbound_{\theta})}^{2}+B_{T}^{y}{(y^{*},\Gbound_{y})}^{2}}\label{eq:pf-Bt},
\end{align}
we have via Cauchy-Schwarz inequality and \Cref{eq:pf-tmp} that
\begin{align*}
  R_{T}(z^{*})&\le
  A_{T}(z^{*})+B_{T}(z^{*})\sqrt{\sum_{t=1}^{T}\norm{\hat g_{t}}^{2}}.
\end{align*}
Thus, PFGTD satisfies the conditions of \Cref{thm:sp-to-olo} with $A_{T}(z)$ and $B_{T}(z)$ of \Cref{eq:pf-At,eq:pf-Bt}.
We can further upperbound these $A_{T}(z^{*})$ and $B_{T}(z^{*})$ to put them in terms
of $\norm{z^{*}}$, $\Gbound = \sqrt{\Gbound_{\theta}^{2}+\Gbound_{y}^{2}}$, and $\hat{G}=\sqrt{\hat{G}_{\theta}^{2}+\hat{G}_{y}^2}$.
In particular, we can get using the AM-GM inequality and Cauchy-Schwarz inequality that
\begin{align*}
  &A_{T}(z^{*})\le\\
    &\quad O\Bigg(W_{0}+\hat GD
  +\norm{z^{*}}\sbrac{\hat{G}+\Log{\frac{\norm{z^{*}}\Gbound T}{W_{0}}+1}}\Bigg)\\
  &B_{T}(z^{*})
  \le O\brac{\norm{z^{*}}\sqrt{\Log{\frac{\norm{z^{*}}\Gbound T}{W_{0}}+1}}}.
\end{align*}
To ease notational burden in the main text, the bound in the proposition
further bounds $\Gbound\le O(\hat G)$ and $\norm{z^{*}}\hat G\le O(\hat GD)$ to clean up notation.

\hfill$\blacksquare$

Observe that in contrast with \Cref{app:tuning-free-olo}, the bound of PFGTD (and the algorithms to follow)
involves dependencies on $\hat{G}=\max_{t}\norm{\hat{g}_{t}}$, the maximum \textit{stochastic} gradient.
The presence of the \(\max_{t}\) can make this term something of a nuisance to bound. Since these terms
only show up in the horizon-independent constant and in the logarithmic terms, we opt for the simple
solution of treating these terms as constants, noting that they can be bounded almost-surely (shown as an aside in Box \ref{box:stochastic-G-bounds}).

Continuing on, CW-PFGTD is constructed similarly to PFGTD, with algorithms $\cA_{\Theta}$ and $\cA_{\cY}$ separately applied to the stochastic subgradients $(\hat g_{t}^{\theta})_{t=1}^{T}$ and $(\hat g_{t}^{y})_{t=1}^{T}$ respectively.
The algorithms $\cA_{\Theta}$ and $\cA_{\cY}$ are constructed by composing constrained vector clipping (\Cref{alg:constrained-clipping} with $M^{\text{vec}}:g\mapsto(\abs{g_{1}},\ldots,\abs{g_{d}})^{\top}$) with $\text{CW-PF}$ (\Cref{alg:cwpf}).
The next proposition characterizes the regret of the coordinate-wise algorithm, CW-PFGTD.

\CWPFGTDRegret*

\textit{Proof:} Similar to \Cref{prop:pfgtd}, decompose the regret as
\(
R_{T}(z^{*}) \le R_{T}^{\cA_{\Theta}}(\theta^{*}) + R_{T}^{\cY}(y^{*})
\)
and apply \Cref{thm:constrained-vector-clipping} with $M^{\text{vec}}:\hat g_{t}\mapsto (\abs{\hat g_{t1}},\ldots,\abs{\hat g_{td}})^{\top}$ followed by \Cref{lemma:cwpf}. For $\cA_{\Theta}$ this gives us
\begin{Boxed}{$\norm{\hat{g}_{t}^{\theta}}^{2}\le \hat{G}_{\theta}^{2}\text{ and }\norm{\hat{g}_{t}^{y}}^{2}\le \hat{G}_{y}^{2}$}
Dy definition of $g_{t}^{\theta}$,
\begin{align*}
   \norm{\hatg_{t}^{\theta}} &=  \norm{\hat A_{t}y_{t}}
                  \le  \norm{\rho_{t}(\phi_{t}-\gamma\phi_{t}')\phi_{t}^{\top}y_{t}}\\
                    &\le  \rho_{t}\abs{\inner{\phi_{t},y_{t}}}\norm{\phi_{t}-\gamma\phi_{t}'}\\
                  &\multiscratch{&\le \rho_{t}\norm{\phi_{t}}\norm{y_{t}}\norm{\phi_{t}-\gamma\phi_{t}'}
                    \le \rho_{\max}\sqrt{d}\norm{\phi_{t}}_{\infty}D\norm{\phi_{t}-\gamma\phi_{t}'}\\
                  &\le \rho_{\max}\sqrt{d}L\brac{\norm{\phi_{t}}+\gamma\norm{\phi_{t}'}}D}\\
                                           &\le \rho_{\max}dL^{2}(1+\gamma)D\le  \underbrace{D(1+\gamma)\rho_{\max}dL^{2}}_{\hat{G}_{\theta}},
\end{align*}
where we used that $\abs{\inner{\phi_{t},y_{t}}}\le \norm{\phi_{t}}\norm{y_{t}}\le \sqrt{d}\norm{\phi_{t}}_{\infty}D$, and
$\norm{\phi_{t}-\gamma\phi_{t}'}\le(1+\gamma)(\norm{\phi_{t}}^{2}+\norm{\phi_{t}'}^{2})\le(1+\gamma)\sqrt{d}L$. Using
similar arguments we have that
\begin{align*}
  &\norm{\hat{g}_{t}^{y}}^{2}=\norm{-\hat b_{t}+\hat A_{t}\theta_{t}+\hat M_{t}y_{t}}\\
  &\quad\le \norm{\hat b_{t}} + \norm{\hat A_{t}}\norm{\theta_{t}}+\norm{\hat M_{t}y_{t}}\\
               &\quad= \norm{\rho_{t}r_{t}\phi_{t}} + \norm{\rho_{t}\phi_{t}(\phi_{t}-\gamma\phi_{t})^{\top}}\norm{\theta_{t}}+\norm{\phi_{t}\phi_{t}^{\top}y_{t}}\\
               &\multiscratch{&\quad\le \rho_{t}\abs{r_{t}}\sqrt{d}\norm{\phi_{t}}_{\infty} + \norm{\rho_{t}\phi_{t}(\phi_{t}-\gamma\phi_{t})^{\top}}\norm{\theta_{t}}+\norm{\phi_{t}\phi_{t}^{\top}y_{t}}\\
                        &\quad\le \rho_{\max}R_{\max}\sqrt{d}L + (1+\gamma)\rho_{\max}dL^{2}D+ \norm{\phi_{t}}\abs{\inner{\phi_{t},y_{t}}}\\
                        &\quad\le \rho_{\max}R_{\max}\sqrt{d}L + (1+\gamma)\rho_{\max}dL^{2}D+ \sqrt{d}L\norm{\phi_{t}}\norm{y_{t}}\\
                        &\quad\le \rho_{\max}R_{\max}\sqrt{d}L + (1+\gamma)\rho_{\max}dL^{2}D+ dL^{2}D\\
              }\\
              &\quad\le \rho_{\max}R_{\max}\sqrt{d}L + (1+\gamma)\rho_{\max}dL^{2}D+ dL^{2}D\\
               &\quad= \rho_{\max}\sqrt{d}L\brac{R_{\max}+D(1+\gamma)\sqrt{d}L} + dL^{2}D\\
                 &\quad= \underbrace{\rho_{\max}R_{\max}\sqrt{d}L + dL^{2}D\sbrac{(1+\gamma)\rho_{\max}+1}}_{\hat{G}_{y}}
\end{align*}
\labelbox{box:stochastic-G-bounds}
\end{Boxed}
\begin{align*}
  R_{T}^{\cA_{\Theta}}(z^{*})&\le
                               \sum_{i=1}^{d}A_{T}^{\theta_{i}}(\theta_{i}^{*},\Gbound_{\theta_i}) + B_{T}^{\theta_{i}}(\theta_{i}^{*},\Gbound_{\theta_i})\sqrt{\sum_{t=1}^{T}\abs{\hat g^{\theta}_{ti}}^{2}}
\end{align*}
where $\Gbound_{\theta_{i}}=\max\set{\gbound, G_{\theta_{i}}}$, $\hat G_{\theta_i}=\max_{t}\abs{\hat g_{ti}}$ and $A_{T}^{\theta_{i}},\  B_{T}^{\theta_i}$ are non-negative deterministic functions with
\begin{align*}
  A_{T}^{\theta_{i}}(\theta^{*}_{i},\Gbound_{\theta_{i}})&\le
                                      O\Bigg(W_{0}+\hat{G}_{\theta_i}\brac{D_{\infty,\theta}+\norm{\theta^{*}}_{\infty}} \\
  &\qquad\qquad
  \abs{\theta_{i}^{*}}\Log{\frac{\abs{\theta_{i}^{*}}\Gbound_{\theta_i}T}{W_{0}}+1}\Bigg)\\
B_{T}^{\theta_{i}}(\theta_{i}^{*},\Gbound_{\theta_{i}})&\le O\brac{\abs{\theta_{i}^{*}}\sqrt{\Log{\frac{\abs{\theta_{i}^{*}}\Gbound_{\theta_i}T}{W_{0}}+1}}},
\end{align*}
where $\Gbound_{\theta_{i}}\defeq \max\set{\gbound, \max_{t}\abs{\hat g_{ti}^{\theta}}^{2}}$ and $D_{\infty,\theta}=\max_{\theta\in\Theta}\norm{\theta}_{\infty}$.
A bound of the same form holds for $R_{T}^{\cA_{\cY}}(y^{*})$,
with $A_{T}^{y_{i}}$ and $B_{T}^{y_{i}}$ defined analogously.
Then by letting
\begin{align}
  A_{T}^{(i)}(z_{i}^{*})&\defeq A_{T}^{\theta_{i}}(\theta^{*},\Gbound_{\theta_{i}})+A_{T}^{y_{i}}(y^{*},\Gbound_{y_{i}})\label{eq:cw_At}\\
  B_{T}^{(i)}(z_{i}^{*})&\defeq \sqrt{B_{T}^{\theta_{i}}(\theta_{i}^{*},\Gbound_{\theta_{i}})^{2}+B_{T}^{y_{i}}(y_{i}^{*},\Gbound_{y_{i}})^{2}},\label{eq:cw_Bt}
\end{align}
we can write
\begin{align*}
  R_{T}(z^{*})\le \sum_{i=1}^{d}A_{t}^{(i)}(z^{*}_{i}) + B_{T}(z^{*}_{i})\sqrt{\sum_{t=1}^{T}\abs{\hat g_{ti}}^{2}}.
\end{align*}
To bound the functions $A_{T}^{(i)}$ and $B_{T}^{(i)}$, it will be useful to let $\norm{z}_{\infty}=\max\set{\norm{\theta}_{\infty},\norm{y}_{\infty}}$ for
$z=(\theta,y)\in\Theta\times\cY$.
Then use \holder's inequality and AM-GM inequality to write
\begin{align*}
  A_{T}^{(i)}(z_{i}^{*})&\le O\Bigg(2W_{0}+(D_{\infty}+\norm{z^{*}}_{\infty})2\avg{G}_{i} \\
  &\qquad+ 2\norm{z^{*}}_{\infty}\Log{\frac{\norm{z^{*}}_{\infty}\avg{\Gbound}_{i}T}{W_{0}}+1}\Bigg)
\end{align*}
where we've defined $\hat G_{i}= \frac{\hat G_{\theta_{i}}+\hat G_{y_{i}}}{2}$ and $\Gbound_{i}=\frac{\Gbound_{\theta_{i}}+\Gbound_{y_{i}}}{2}$.
It's easy to see that $B_{T}^{(i)}(z^{*}_{i})$ can be bounded similarly:
\begin{align*}
  B_{T}^{*}(z^{*}_{i})\le O\brac{2\norm{z^{*}}_{\infty}\sqrt{\frac{\norm{z^{*}}_{\infty}\avg{\Gbound}_{i}T}{W_{0}}+1}}
\end{align*}
Finally, bound $\avg{\Gbound}\le O(\avg{G})\le O(G_{\infty})$ and $\norm{z^{*}}_{\infty}G_{\infty}\le D_{\infty}G_{\infty}$ for $G_{\infty}=\max_{t}\norm{\hat g_{t}}_{\infty}$ to clean up notation.

\hfill$\blacksquare$

The next Corollary tells us that the bound for CW-PFGTD will also satisfy  the conditions of
\Cref{thm:sp-to-olo}, and thus this algorithm will also match the rate of GTD2 up to log terms --- though
unlike PFGTD, this algorithm has an unfavorable dependence on the dimension $d$ in the worst-case.

\begin{corollary}\label{corr:cwpfgtd}
  CW-PFGTD satisfies the conditions of \Cref{thm:sp-to-olo} with
    \begin{align*}
    A_{T}(z^{*})
    &\le O\Bigg(d\brac{W_{0}+D_{\infty}G_{\infty}}+\\
    &\quad+d\norm{z^{*}}_{\infty}\sbrac{G_{\infty}+\Log{\frac{\norm{z^{*}}_{\infty}G_{\infty}T}{ W_{0}}+1}}\Bigg)\\
      B_{T}(z^{*})&\le O\brac{\sqrt{d}\norm{z^{*}}_{\infty}\sqrt{\Log{\frac{\norm{z^{*}}_{\infty}G_{\infty}T}{W_{0}}+1}}},
    \end{align*}
    where $G_{\infty}=\max_{t}\norm{\hat g_{t}}_{\infty}$.
\end{corollary}

\textit{Proof:} The bound for $A_{T}(z^{*})$ follows immediately from adding up the $A_{T}^{(i)}$
and applying AM-GM inequality to bring together the log terms, yielding
\begin{align*}
  A_{T}(z^{*})
  &\le O\Bigg(2d\brac{W_{0}+D_{\infty}\avg{G}}+\\
  &\quad+d\norm{z^{*}}_{\infty}\sbrac{\avg{G}+2\Log{\frac{\norm{z^{*}}_{\infty}\avg{\Gbound}T}{ W_{0}}+1}}\Bigg).
\end{align*}
Next, let $B_{T}(z^{*})=\sqrt{\sum_{i=1}^{d}B_{T}^{(i)}(z_{i}^{*})^{2}}$ and use Cauchy-Schwarz inequality to write
\begin{align*}
  \sum_{i=1}^{d}B_{T}^{(i)}(z_{i}^{*})\sqrt{\sum_{t=1}^{T}\abs{\hat g_{ti}}}\le B_{T}(z^{*})\sqrt{\sum_{t=1}^{T}\norm{\hat g_{t}}^{2}},
\end{align*}
and use AM-GM inequality again to get
\begin{align*}
  B_{T}(z^{*})\le\sqrt{2d}\norm{z^{*}}_{\infty}\sqrt{\Log{\frac{\norm{z^{*}}_{\infty}\avg{\Gbound}T}{W_{0}}+1}}.
\end{align*}
Finally, again bound $\avg{\Gbound}\le O(\avg{G})\le O(G_{\infty})$ for $G_{\infty}=\max_{t}\norm{\hat g_{t}}_{\infty}$ to clean up notation.

\hfill$\blacksquare$

Finally, following \citet{cutkosky2019combining}, we can guarantee the best of both of these bounds by simply
adding the iterates of the PF and CW-PF procedures (See \Cref{app:combining}).
As we'll see, it turns out that the dimension dependence can largely be avoided giving us an ``almost dimension-free'' algorithm in the sense that
in the worst case, we get the dimension-free rate of PFGTD plus a dimension-dependent \textit{constant} penalty, but otherwise may still be able to take advantage of
faster rates when the gradients are sufficiently sparse.

\PFGTDPlusRegret*

\textit{Proof:} PFGTD+ composes the constrained vector clipping (\Cref{alg:constrained-clipping} with $\cM:g\mapsto(\abs{g_{1}},\ldots,\abs{g}_{d})^{\top}$) with PF+ for both $\cA_{\Theta}$ and $\cA_{\cY}$.
Thus, for $\cA_{\Theta}$ we have from \Cref{thm:constrained-vector-clipping} followed by \Cref{lemma:pfcombined} that
\begin{align*}
  R_{T}^{\cA_{\Theta}}(\theta^{*})&\le \frac{d}{2}W_{0} + d\avg{G_{\theta}}\brac{D_{\infty}+\norm{\theta^{*}}_{\infty}} \\
                                  &\qquad+ \min\set{R_{T}^{\text{PF}}(\theta^{*},\norm{\Gbound_{\theta}}), R_{T}^{\text{CW-PF}}(\theta^{*},\Gbound_{\theta})},
\end{align*}
where $\overline{G}_{\theta}=\frac{\sum_{i=1}^{d}G_{\theta_{i}}}{d}$ with $G_{\theta_{i}}=\max_{i}\abs{\hat g_{ti}^{\theta}}$, and likewise
we have a bound of the same form for $\cA^{\cY}$. Adding the two together gives us
\begin{align*}
  R_{T}(z^{*})&\le R_{T}^{\cA_{\Theta}}(\theta^{*}) + R_{T}^{\cA_{\cY}}(y^{*})\\
  &\le dW_{0} + 2d\avg{G}\brac{D_{\infty} + \norm{z^{*}}_{\infty}}\\
  &\quad+ \min\set{R_{T}^{\text{PF}_{\Theta}}(\theta^{*},\norm{\Gbound_{\theta}}), R_{T}^{\text{CW-PF}_{\Theta}}(\theta^{*},\Gbound_{\theta})}\\
  &\quad+ \min\set{R_{T}^{\text{PF}_{\cY}}(y^{*},\norm{\Gbound_{y}}), R_{T}^{\text{CW-PF}_{\cY}}(y^{*},\Gbound_{y})}\\
              &\le dW_{0} + 2d\avg{G}\brac{D_{\infty} + \norm{z^{*}}_{\infty}}\\
              &\quad+ \min\Big\{R_{T}^{\text{PF}_{\Theta}}(\theta^{*},\norm{\Gbound_{\theta}})+R_{T}^{\text{PF}_{\cY}}(y^{*},\norm{\Gbound_{y}}),\\
                &\qquad\qquad\ \ \  R_{T}^{\text{CW-PF}_{\Theta}}(\theta^{*},\Gbound_{\theta})+R_{T}^{\text{CW-PF}_{\cY}}(y^{*},\Gbound_{y})\Big\}\\
              &= dW_{0} + 2d\avg{G}\brac{D_{\infty} + \norm{z^{*}}_{\infty}}\\
  &\quad+\min\set{R_{T}^{\text{PF}}(z^{*}), R_{T}^{\text{CW-PF}}(z^{*})},
\end{align*}
where in the second line we use $\avg{G}=\frac{\avg{G}_{\theta}+\avg{G}_{y}}{2}$. Lastly, again
use $\avg{G}\le G_{\infty}=\max_{t}\norm{\hat g_{t}}_{\infty}$.
\hfill$\blacksquare$

\subsection{Discussion of Parameters}\label{app:parameters}

Similar to prior works, our proposed algorithms avoid requiring knowledge of problem-dependent parameters such as
the Saddle-point $\norm{z^{*}}$ and Lipschitz constant $G$, yet still retain a dependence on the user-specified initial wealth $W_{0}$
and initial hint $\gbound$ \citep{cutkosky2019artificial}. We emphasize that these should \textit{not} be thought of as tunable hyperparameters ---
indeed, inspection of \Cref{prop:pfgtd,prop:cwpfgtd,prop:pfgtdplus} reveal that tuning of either parameter can at best improve
constant and log factors. Rather, these parameters should \textit{at most} be considered as ways to incorporate prior information,
\textit{if} it happens to be available. For instance, we observe from the regret guarantees that the initial wealth $W_{0}$ turns up only
in the constant factors and log terms of the form $\Log{\frac{\norm{z^{*}}GT}{W_{0}}+1}$.
If one knows that both the horizon $T$ and the Lipschitz bound $G$ will be large, for example, it may be favorable to set $W_{0}$
to be large, inducing a smaller logarithmic penalty on the horizon-dependent terms at the expense of a larger
constant penalty (which has only a transient effect).
However, we note that no such considerations were used in our
experiments --- all of our results are attained by naively setting $W_{0} = \gbound = 1$.

The parameter $\gbound$ has a similarly limited effect. If $\gbound<G$, we end up with the constant penalty $O(G(D+\norm{z^{*}}))$
induced by the gradient clipping reduction (see \Cref{app:reductions}, \Cref{thm:imperfect-hinting}). Otherwise $\gbound$ shows up only in the logarithmic terms as $\Log{\frac{\norm{z^{*}}\gbound T}{W_{0}}+1}$,
and thus $\gbound$ needs to be \textit{exponentially larger} than $G$ in order to increase the coefficient of these terms by any meaningful amount. Such an
extreme over-estimate is typically easy to avoid in practice.

It goes without saying that there are surely pathogenic cases for which our proposed default parameters can cause issues. As a simple example,
it is easy to see from the preceeding discussion that using $\gbound=1$ in an MDP for which $R_{\max}=e^{-T}$ would induce linear regret.
However, such extreme conditions rarely arise in practice, and it is unlikely that the practitioner would be completely unaware of such
extreme conditions when they are present. Finally, it is important to note that neither of these input
parameters can effect asymptotic convergence; the average regret approaches zero as $T\rightarrow\infty$
no matter how poorly one manages to set these parameters.

\section{Experiments}

This section contains supplementary details and results for the experiments in \Cref{section:experiments}.
Details and procedures of each experiment are presented in \Cref{app:experiment-details}, and
additional experimental results can be found in \Cref{app:experiment-results}.

\subsection{Experimental Details}\label{app:experiment-details}

\textbf{Baselines.}
Our parameter-free algorithms are based off of the Saddle-point formulation of the MSPBE
used by the GTD2 algorithm. Thus, GTD2 is our primary baseline;
success in these experiments means performing reasonably well relative to a well-tuned GTD2
baseline, suggesting that our algorithms can be used as a drop-in replacement for GTD2, with
similar guarantees and similar performance in practice, yet requiring no parameter tuning.
Additional baselines are included to further contextualize the performance.
We include TDC, another member of the
gradient TD family of algorithms, which has no Saddle-point interpretation but
can be formulated as a two-timescale stochastic approximation algorithm \citep{liu2018proximal,sutton2009fast}.
We also include for reference regular TD. It is well known that in practice, the performance of
vanilla TD generally remains unmatched by the gradient algorithms on \textit{most} problems, despite its
potential instability issues.
The trade-off is that on some particular problem instances, TD will diverge for any fixed step-size \citep{baird1995residual}.
There has yet to be a stable gradient TD algorithm which achieves robust guarantees
while also significantly improving over the performance of semi-gradient TD, so we omit this baseline
in the main text, instead focusing on how our method measures up against other gradient TD algorithms.
A recent improvement of the TDC
algorithm, called TDRC, uses regularized corrections to enable more ``TD-like'' behavior while
still maintaining the guarantees of the gradient TD methods \citep{ghiassian_gradient_2020}, and currently represents
the state-of-the-art method for the off-policy policy evaluation with linear function approximation problem setting.
TDRC has an additional hyperparameter $\beta$ controlling the strength of regularization. We use the
default setting $\beta=1$ suggested by \citet{ghiassian_gradient_2020}, which was used in their experimental evaluation as well.

We note that the guarantees of the gradient TD algorithms (including our own) hold only under the assumption of
iterate averaging (or under the typical Robbins-Monro step-size schedule). It is common to instead just use the last iterate
with a constant step-size, because it tends to work better in practice. Our experiments conform to this practice in order to
make the conditions as favorable as possible for the baselines.
Indeed, the baselines were tested both with and without
iterate averaging, and their performance was strictly worse on all problems when using iterate averaging; we omit
these results for brevity.
In contrast, we enforce that our methods use iterate averaging to show the performance of
the proposed algorithm, implemented exactly as prescribed by the theory.
This puts our methods at a marked disadvantage in some problems (particularly
in Baird's counterexample, as noted in \Cref{app:experiment-results}), but it is interesting to note that
the performance of our methods is still relatively competitive \textit{while still achieving the promised guarantees},
suggesting a smaller gap between theory and practice for our methods.

\subsubsection{Classic RL Problems}\label{app:classic-rl}

\begin{figure}[t!]
  \centering
  \includegraphics[width=\columnwidth]{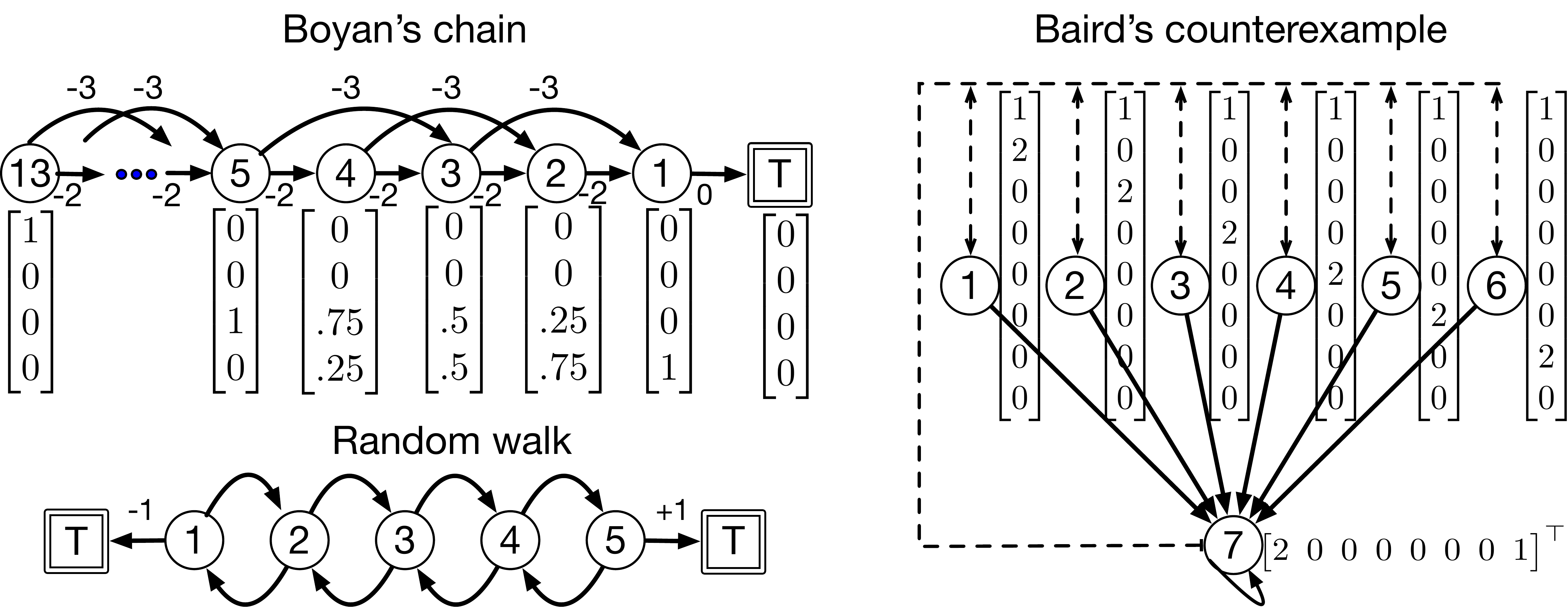}
  \caption{
    An illustration of the three classic RL MDPs used in our experiments,
    reproduced from \citet{ghiassian_gradient_2020}.
    The illustrations of Boyan's chain and Baird's star counterexample
    include a depiction of their respective feature representations.
    The three feature representations used in the random walk experiments
    are described \Cref{app:classic-rl}
  }
  \label{fig:classic-rl}
\end{figure}

We first evaluate performance on a number of classic RL problems. We consider
 three domains, each depicted in \Cref{fig:classic-rl}: a 5-state random walk with a number of different feature representations \citep{sutton2009fast},
Boyan's chain \citep{boyan2002technical}, and Baird's counterexample \citep{baird1995residual}.
In each of these simple problems, the RMSPBE can be computed exactly using
\Cref{eq:objective}, and is measured after each step of interaction with the environment.

\textbf{In all experiments in this section,} learning curves are averaged over 200 independent runs.
We tuned the step-size of the baselines over the geometric range $\alpha\in\set{2^{-10},\ldots,2^{0}}$, and
step-sizes were chosen to minimize the area under the RMSPBE curve.
For the CDF plots, each algorithm was given a budget of $5,000$ runs, and on each run
a step-size was sampled uniformly between $2^{-1}$ and $2^{-10}$, as described in \Cref{section:experiments}.
The final RMSPBE was measured at the end of each run, and the data was aggregated as described in
\Cref{section:experiments} to create the CDF plots. The procedure was additionally repeated with
hyperparameters sampled uniformly from a discrete grid of values; the results were qualitatively similar
so we omit them for brevity.

\textbf{Random Walk.} The random walk problem is a simple episodic undiscounted MDP, consisting
of a chain of five states with terminal states on the left and right ends of the chain.
The rewards are zero  everywhere except when transitioning to the left or right terminal states,
in which the reward is $-1$ and $+1$ respectively.
Each episode begins in the middle state, and
the agent follows a behavior policy which transitions left and right with equal probability.
The objective is to estimate the value function of a target policy which
transitions left/right with $40\%$/$60\%$  probability respectively.
Following \citet{sutton2009fast} and \citet{ghiassian_gradient_2020}, we test three
different feature representations: \textit{tabular}, \textit{dependent}, and \textit{inverted}
features. The tabular features are given by a one-hot encoding of the states (\ie $\phi(s_{1})=(1, 0,0,0,0)^{\top}$). The inverted features are the opposite encoding scheme, a ``one-cold'' encoding
in which $[\phi(s_{i})]_{j}=1-\delta_{ij}$ for $\delta_{ij}$ being the dirac delta function.
Finally, the dependent feature representation uses features $\phi(s_{1})=(1, 0, 0)^{\top}$,
$\phi(s_{2})=(\frac{1}{\sqrt{2}},\frac{1}{\sqrt{2}},0)^{\top}$, $\phi(s_{3})=(\frac{1}{\sqrt{3}},\frac{1}{\sqrt{3}},\frac{1}{\sqrt{3}})^{\top}$,
$\phi(s_{2})=(0,\frac{1}{\sqrt{2}},\frac{1}{\sqrt{2}})^{\top}$, and $\phi(s_{5})=(0,0,1)^{\top}$.
Performance was measured over $3,000$ steps of interaction with the environment.
Value estimates of all algorithms were initialized $\hat v^{\pi}(s)=0,\ \forall s\in\cS$.

\textbf{Boyan's Chain.} Boyan's chain is a classic on-policy, episodic, undiscounted MDP.
Each episode starts in the left-most state, and in
state $s_{i}$ the agent transitions with equal probability to either to state $s_{i+1}$ with reward $-2$ or to state $s_{t+2}$ with reward $-3$. The exception
to this rule is state $s_{1}$, in which the agent transitions deterministically
to an absorbsing state with reward of $0$.
This problem is frequently used to investigate the ability of an algorithm to
handle representations with harmful aliasing properties --- the structure of
the feature representation leads
to inappropriate generalization between the value estimates of adjacent states.
Performance was measured over $10,000$ interactions with the environment, and the value estimates of
each all algorithms were initialized to $\hat v^{\pi}(s)=0,\ \forall s\in\cS$.

\textbf{Baird's Counterexample.} Baird's ``star'' counterexample is a well-known MDP which
causes off-policy TD with linear function approximation to diverge for any scalar non-zero step-size \citep{baird1995residual}.
The MDP consists of $7$ states and two actions. The first action leads deterministically to state $7$, while the second action leads
to one of the first $6$ states with equal probability. A reward of zero is received on all transitions, so the value function under \textit{any} policy is
$v^{\pi}(s)=0,\ \forall s\in\cS$. The task is to estimate the value function with discount factor $\gamma=0.99$
under a deterministic target policy which chooses $\pi(s)=1$ for all states (\ie policy which always chooses to go to state $7$), while following
an equiprobable random behavior policy. The features are linearly independent and can perfectly realize the value funtion.

Performance was measured over $5,000$ steps of interaction with the environment.
Value function estimates were initialized with parameters $\theta_{1}=(1, 1, 1, 1, 1, 1, 1, 10)^{\top}$, which is the standard initialization for this problem.
Note that there is no clear way make such an initialization for our coin-betting based algorithms. Typically
coin-betting algorithms bet $\theta_{1}=\beta_{1}W_{0}$ with $\beta_{1}=0$ before having observed any feedback, but this is would lead to a trivial solution
in Baird's counterexample. For the CW-PF component, we choose instead $\beta_{1i}=\half$ and $W_{1i}=2\theta_{1,i}$ to get $\beta_{1i}W_{0i}=\theta_{1,i}$,
and similarly, for the PF component we initialize $\beta_{1}=\half$, $W_{0}=2\norm{\theta_{1}}$, and $u_{1}=\theta_{1}/\norm{\theta_{1}}$ to give $v_{1}u_{1}=(\beta_{1}W_{0})u_{1}=\theta_{1}$.

\subsubsection{Large-scale Prediction}\label{app:nexting}

Our final experiment tests the ability of our algorithms to scale to real-world prediction problems.
The task contains many of the difficulties associated with real-world problems:
the data is generated from the raw sensor readings
of a mobile robot as it interacts with its environment; the data is high-dimensional, noisy, contains
unpredictable changes and non-stationarities, and the prediction magnitudes can be extremely large --- some reaching
values in the millions --- posing difficulties to learning stability.

We recreate the robot prediction task of \citep{modayil2014multi}, in which the learner predicts the future sensor readings of a real mobile robot as it interacts with its environment according to a fixed behavior policy.
The data was generated from each of the robot's 53 sensors, recorded at a time interval of 100 milliseconds for a total of $120,000$ time-steps, corresponding to approximately $3$ hours of runtime of the mobile robot.
The predictions in this task are formulated as $\gamma$-discounted returns with $\gamma=0.9875$,
corresponding to an approximate horizon of $80$ time-steps into the future. The objective is to
accurately predict these discounted returns for each of the robot's $53$ sensors.
We used the same sparse representation used in the original work, consisting of
a coarse-coding of
6065 binary features. A complete description of how these features are constructed from the raw sensor readings can be found in \citet{modayil2014multi}.
In short, a number of mappings are defined by considering a subset of the sensors and defining a discretization over the joint space of their readings;
a binary feature is set depending on which cell of the grid the joint sensor reading falls into at time $t$. In particular, 457 of these mappings
are defined in this experiment, leading to 457 active binary features on each time step.

For each algorithm we processed the dataset incrementally, constructing the feature vectors from the raw sensor readings and making a prediction for
each sensor. At the end of the experiment, the returns $G_{t}^{(s)}$ were computed for each sensor $s$, and performance was measured in terms
of the Symmetric Mean Absolute Percentage Error (SMAPE), defined by
\begin{align*}
  \text{SMAPE}(t, s) \defeq \frac{1}{T} \sum_{\tau = 1}^t  \frac{| \hat{v}^{(s)}(S_\tau) - G_\tau^{(s)}|}{|G_\tau^{(s)}| + |\hat{v}^{(s)}(S_\tau)|}.
\end{align*}
The SMAPE is a relative error, and has the advantage of always being in the range $[0,1]$, making it
possible to compare and aggregate the performance across different sensors.
Overall performance is
measured by aggregating the resulting learning curves via the median.
The median is generally a better measure of aggregate performance than the mean
in this problem, due to the presence of a small number of high-magnitude noisy
sensors \citep{jacobsen2019meta,modayil2014multi}.

In this problem, we tuned the step-size of the baseline algorithms over the range
$\alpha\in\set{\frac{1}{1.785}2^{-i} : i=-16,\ldots,-2}$. The range was chosen to span a
wide range of possible step-sizes --- from $8.5\mathrm{e}{-6}$ to $0.14$ ---
while also maintaining a moderate tuning budget. Step-sizes were then chosen
to minimize the area under the median SMAPE curve, corresponding to the step-size
which generally performs best across all sensors.
Value estimates were initialized $\hat v^{(s)}(\cdot)=0$ for all sensors $s$.

\subsection{Additional Experimental Results}\label{app:experiment-results}

\begin{figure}[!t]
  \centering
  \includegraphics[width=0.85\columnwidth]{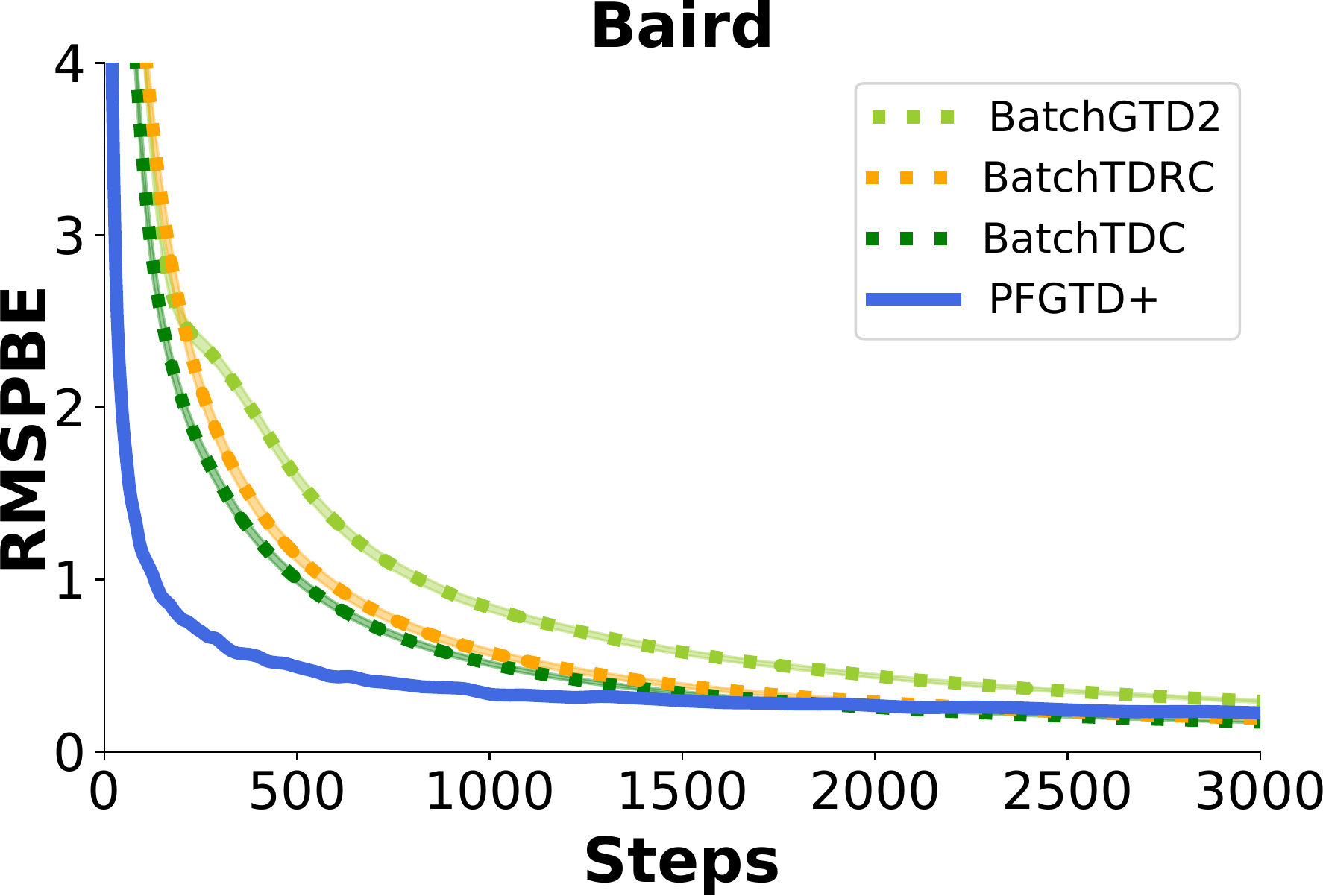}
  \caption{Performance in Baird's counterexample when using iterate averaging.
    Learning curves show the accuracy of $\hat v^{\pi}_{t}(s) = \inner{\phi(s),\thetabar_{t}}$,
    where $\thetabar_{t}=\frac{1}{t}\sum_{\tau=1}^{t}\hat \theta_{\tau}$.
    Results are averaged over 200 runs, and $\pm 1$
    standard error is depicted with shading.
  }
  \label{fig:batch-baird}
\end{figure}
In this section we include additional experimental results and discussion.
All plots are best viewed in color.

\Cref{fig:classic-rl-all} (included on the last page) shows the results on the
classic RL problems, with all baselines included.
The plots reinforce the result of \Cref{prop:pfgtdplus} as well as the conclusion made from the in-line barplot of \Cref{section:experiments}:
in all problems, PFGTD+ performs similarly to either PFGTD or CW-PFGTD --- whichever happens to perform better in a given problem.
Another result which seems to stand out is Baird's counterexample,
being the
only problem in which the parameter-free methods fail to reach the same final performance as
the baselines. Interestingly, this seems to represent a weakness of \textit{iterate averaging}
rather than some underlying weakness of parameter-free OLO methods. In particular, one of the key difficulties
in Baird's counterexample is that $\hat v(7)$ is initialized to a high value and there is
only a $1/7$ change in making a transition that results in lowering $\hat v(7)$ towards the correct value of zero.
Because of this, methods using iterate averaging
end up converging to zero MSPBE only asymptotically.
This can be seen in \Cref{fig:batch-baird}, which shows the performance of all baselines when iterate averaging is used.
We observe that when using iterate averaging, all baselines see similarly slow convergence. We recall again that
the guarantees that these baselines achieve hold only under the assumption that they're implemented with some form of
iterate averaging --- in a sense, this result is a more faithful representation of the performance of the baselines on this problem.

\Cref{fig:cb-medians-all} shows a direct comparison of the median SMAPE over time for each of the baselines.
A few trends stick out immediately. First, we observe that PFGTD performs quite poorly in comparison with all
other baselines. This is unsurprising as PFGTD makes use of gradient descent with step-sizes $\eta_{t} = \sqrt{2}/2\sqrt{\sum_{\tau=1}^{t}\norm{\widetilde{g}_{t}}^{2}}$
as a component in its hierarchy of reductions; the combination of high-dimensional features and high-magnitude sensor readings
quickly drives the step-sizes towards zero. Yet, the features are quite sparse, so few (if any) updates get made to each of the parameters
managed by this sub-procedure before the step-size is small enough to impede learning, leading to poor performance we see in these figures.
Second, we observe a slight upward trend in the SMAPE of PFGTD+ after it reaches its lowest point. This illustrates another drawback associated
with iterate averaging: the data is non-stationary, so converging to a fixed solution inevitably leads to some unwanted drift in performance
thereafter. In practice this could likely be avoided by using an exponential moving average, for example, though this would require choosing a
decay rate. Doing so without tuning hyperparameters is non-trivial and is an ongoing direction of future work.
Finally, we observe  that none of the gradient-based methods are able to compete with TD in this problem. Though this is the case in general,
the performance gap appears to be particularly pronounced in this problem. We conjecture that this is related to non-stationarity
of the data; the gradient-based algorithms are likely less able to track non-stationary data in general,
due the interactions between the primary and secondary parameters $\theta_{t}$ and $y_{t}$.

\begin{figure}[t!]
  \centering
  \includegraphics[width=\columnwidth]{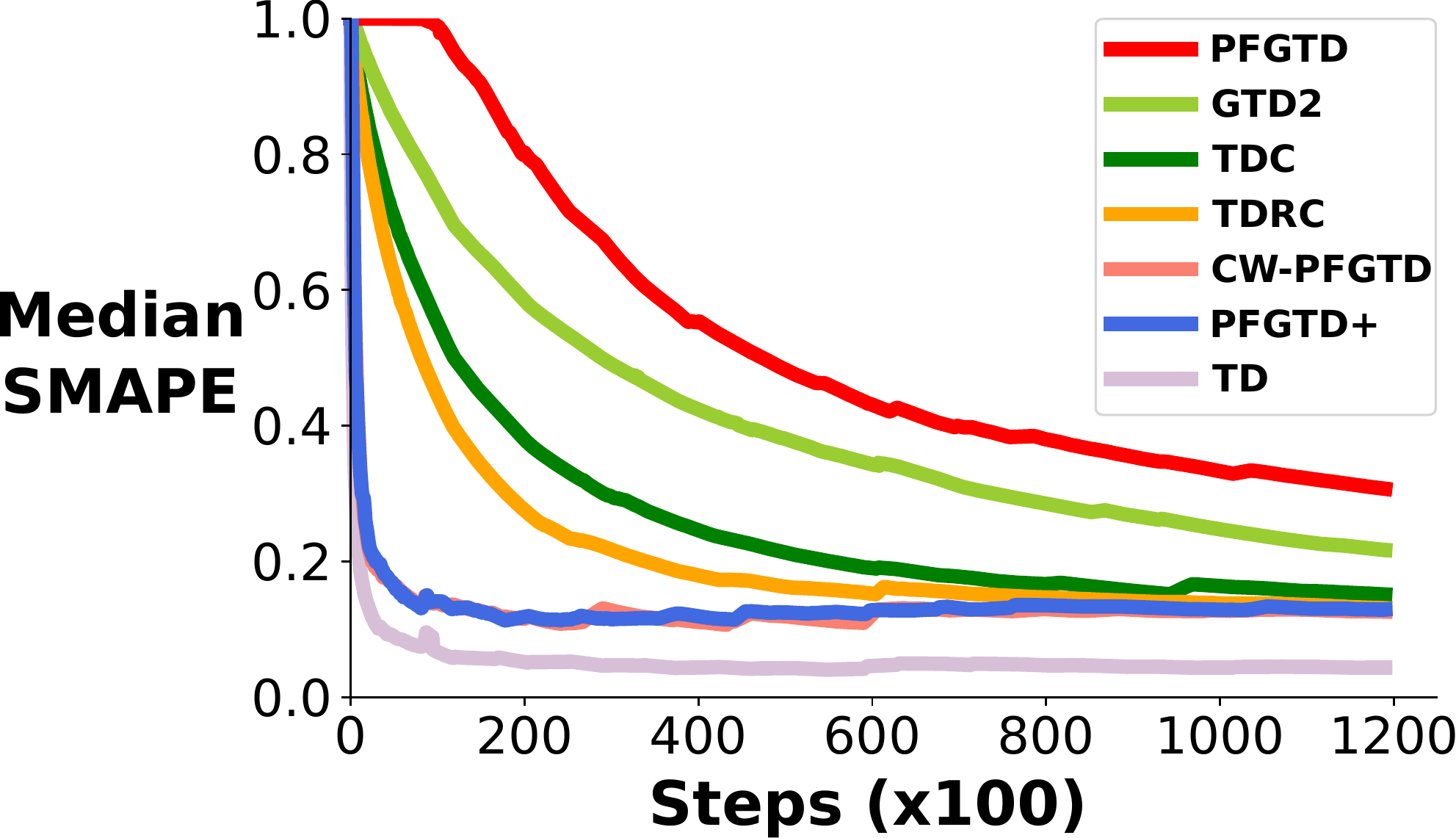}
  \caption{
    Median SMAPE across all sensors in the large-scale prediction task. Baseline step-sizes
    were tuned over the values $\alpha\in\set{\frac{1}{1.785}2^{-i} : i=-16,\ldots,-2}$, and
    chosen to minimize the area under the median SMAPE curve.
  }
  \label{fig:cb-medians-all}
\end{figure}

The unaggregated performance of each algorithm in the large-scale prediction task is depicted in \Cref{fig:cb-all}.
Each plot in the grid shows a given algorithm's performance on each of the individual sensors, with the median
performance shown in black. The plot for PFGTD+ is repeated in each row to enable easy comparison to the baselines.
The plots show that PFGTD+ achieves favorable performance over TDC and TDRC not only in the median, but in general across
sensors in this problem. The second row of plots demonstrates that PFGTD+ performs nearly identically to CW-PFGTD, despite
the fact that it also uses iterates from the PF subroutine which performs poorly in this specific problem, further demonstrating
\Cref{prop:pfgtdplus}.

\begin{figure}[t]
  \centering
  \includegraphics[width=\columnwidth]{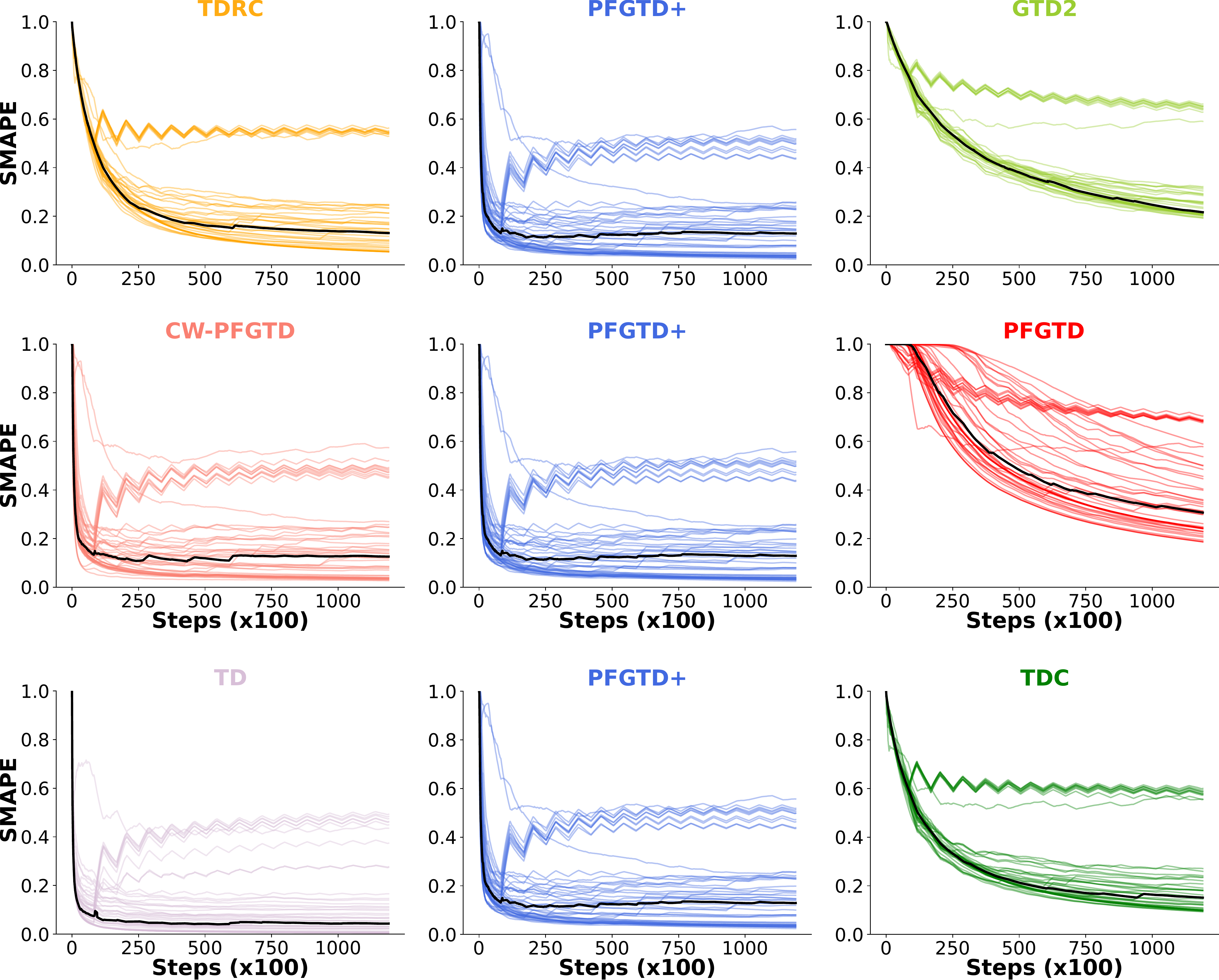}
  \caption{
    SMAPE over time for each individual sensor in the large-scale prediction task. Each colored line represents
    the SMAPE between the learner's predictions and the true discounted sum of future sensor readings for a given sensor,
    with the median shown in black. The plots are arranged in a grid to facilitate direct visual comparison between
    the proposed algorithm, PFGTD+, and each of the baselines.
  }
  \label{fig:cb-all}
\end{figure}


\begin{figure*}
  \centering
  \includegraphics[width=0.7\textwidth]{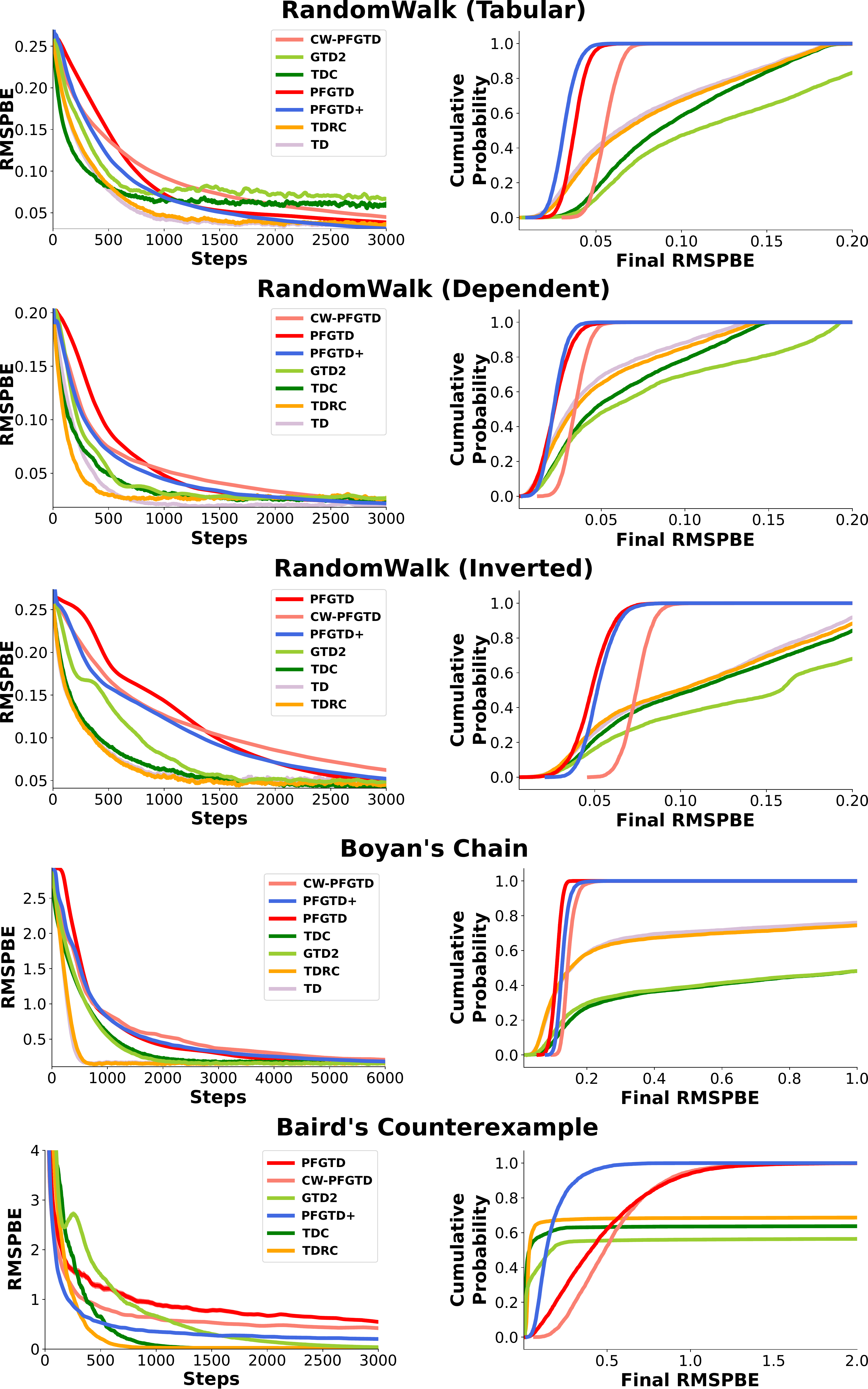}
  \caption{Learning curves and CDF plots for each of the classic RL tasks (view in color).
    \textbf{Left column:} Learning curves for each of the classic RL tasks. Results are averaged over 200 independent runs. $\pm 1$ standard error
    is shaded around each of the lines, but it is generally vanishingly small.
    Baseline hyperparameters were tuned over the range $\alpha\in\set{2^{-i}:i=0,\ldots,10}$, and were chosen to minimize the area under the learning curve.
    \textbf{Right column:} CDF plots for each of the classic RL tasks. Each plot depicts the distribution of final RMSPBE over a sample of 5,000 independent runs.
    Hyperparameters of baseline algorithms were sampled uniformly between $[2^{0},2^{-10}]$.
     The experiment was also repeated with hyperparameters sampled from a
     discrete grid of values; the results were qualitatively similar and have been omitted for brevity.
  }
  \label{fig:classic-rl-all}
\end{figure*}

}

\end{document}